\documentclass[runningheads]{llncs}
\usepackage{graphicx}
\usepackage{amsmath,amssymb} 
\usepackage{color}
\usepackage{xcolor}
\usepackage{colortbl}
\usepackage{wrapfig}
\usepackage{array}
\newcommand{\mypar}[1]{\vspace{2mm}\noindent\textbf{#1}}
\newcommand{\cfbox}[2]{%
    \colorlet{currentcolor}{.}%
    {\color{#1}%
    \fbox{\color{currentcolor}#2}}%
}
\begin{document}

\newcommand{\point}{
    \raise0.7ex\hbox{.}
    }


\pagestyle{headings}

\mainmatter

\title{Spatio-Temporal Attention Models for Grounded Video Captioning} 

\titlerunning{Spatio-Temporal Attention Models for Grounded Video Captioning} 

 \authorrunning{Mihai Zanfir, Elisabeta Marinoiu, Cristian Sminchisescu}
\newcommand*\samethanks[1][\value{footnote}]{\footnotemark[#1]}
\author{Mihai Zanfir$^2$\thanks{Authors contributed equally.},  Elisabeta Marinoiu$^2$\samethanks, Cristian Sminchisescu$^{1, 2}$\\
{\tt\small mihai.zanfir@imar.ro,  elisabeta.marinoiu@imar.ro, cristian.sminchisescu@math.lth.se}} 
\institute{$^1$Department of Mathematics, Faculty of Engineering, Lund University\\ $^2$Institute of Mathematics of the Romanian Academy} 

\maketitle

\begin{abstract}
 Automatic video captioning is challenging due to the complex interactions in dynamic real scenes. A comprehensive system would ultimately localize and track the objects, actions and interactions present in a video and generate a description that relies on temporal localization in order to ground the visual concepts. However, most existing automatic video captioning systems map from raw video data to high level textual description, bypassing localization and recognition, thus discarding potentially valuable information for content localization and generalization. In this work we present an automatic video captioning model that combines spatio-temporal attention and image classification by means of deep neural network structures based on long short-term memory. The resulting system is demonstrated to produce state-of-the-art results in the standard YouTube captioning benchmark while also offering the advantage of localizing the visual concepts (subjects, verbs, objects), with no grounding supervision, over space and time.

\end{abstract}

\section{Introduction}
In this work, we consider the problem of automatic video captioning, where given an input video, a learned model should describe its content with one or more sentences. This is important considering the increasing rate at which multimedia content is uploaded on the Internet, which in turn requires automatic understanding and description for the retrieval of meaningful content. Automatic video captioning would also be beneficial to human-computer interaction, surveillance and monitoring, and as an aid to the blind and visually-impaired. 

However, the problem of translating from the visual domain to a textual one is challenging, as it ideally involves understanding the key actors, objects and their interaction in the scene, followed by the construction of a both semantically and grammatically correct natural language description. The first part is made difficult by the large number of semantic categories, which exhibit a high inter-class variability. Objects may be of different sizes, shapes and colors, or only partially visible. The lack of available, rich annotated data, and the absence of localization information for the key elements present in a video makes the problem even harder. Data is difficult to collect due to the tedious and time-consuming process of annotating individual video frames.  Progress has been made in the related problem of automatic image captioning, where annotations are plentiful \cite{coco_eval,krishnavisualgenome}. Even with a complete understanding of the video content, there still remains the problem of delivering a sufficiently relevant digest, at different levels of abstraction, required by specific tasks. A bird enthusiast may require a specific video to be described as "Sudan golden sparrows are bathing in water", whereas a regular person may be satisfied with a simpler description like "Birds playing in water". 

Very recent work has focused on attention mechanisms that ground textual elements into the video timeline \cite{hierarchical_rnn15,temporal_att15}. However, the visual domain is usually represented only at a coarse frame level without explicitly revealing the spatio-temporal structure pertaining to a textual element. We believe that a video captioning model can benefit from the work in spatio-temporal segmentation in video \cite{taralova2014motion,oneata_eccv2014,Fragkiadaki_2015_CVPR}, that could offer plausible proposals for localization of the textual elements. Our contributions can be listed as follows: 1) an attention mechanism that links textual elements to spatio-temporal object proposals and is able to provide localized visual support of the words in the generated sentence, with no grounding supervision, 2) integration of  high-level semantic representations obtained both from classifiers learned on YouTube dataset \cite{chen_acl11} (subjects, verbs, objects) and from pre-trained models with state-of-the-art recurrent neural networks, and 3) competitive or better than state-of-the-art results on three different metrics on the challenging YouTube video description dataset. An overview of our model is given in figure \ref{fig:overview}. Illustrations of the detailed textual and visual output produced by our model appear in figure \ref{fig:visual_results}.
\section{Related Work}

Previous approaches to video and image captioning follow broadly two main lines of work: 1) intermediate concept prediction in the form of subject, verb, object or place (S,V,O,P) followed by a template sentence generation step, or 2) full sentence generation using recurrent neural networks, mainly long short-term memory units (LSTMs).

\mypar{Concept Discovery}. Earlier work on video captioning has attempted to first detect a subject, a verb and an object for each video and then form a sentence using a template model and a learned language model. In \cite{youtube_to_text} the authors first mine (S,V,O) triplets from the video descriptions. Then, they learn a  semantic hierarchy over subjects, verbs and objects and use a multi-channel SVM to predict an (S,V,O) triplet over the learned hierarchies by trading-off specificity and semantic similarity. Once an (S,V,O) triplet is obtained for each video, a sentence is formed using a template-based approach. In \cite{fgm} a factor graph model is proposed to combine visual detections with language statistics in order to learn an (S,V,O,P) tuple for each video. The sentence generation step is similar to that of \cite{youtube_to_text}. In \cite{aaai15} the authors propose a framework to jointly model language and vision. The language model is a compositional one, learned over (S,V,O) triplets while the vision model is a two-layer neural network built on top of deep features. Treating concept discovery and sentence generation in separate steps has the advantage of solving two potentially easier and better specified problems instead of a harder, less constrained one. However, the downside is that the resulting sentences can be rigid and may fail to capture the richness of human descriptions.

\mypar{Recurrent Networks for Image and Video Captioning}. Inspired by the recent success of recurrent neural networks in automatic language translation \cite{sutskever2014sequence,bengio_iclr15}, a series of papers made use of similar models in "translating" from a visual input to a textual output where the visual information is usually encoded using convolutional neural networks (CNN). In the case of image captioning, significant progress has been made in recent work \cite{showandtell2015,Xu2015show,karpathy2015deep,johnson2015densecap}. The authors in \cite{Xu2015show} use an attention mechanism on top of a CNN and extract features
from a lower convolutional layer in order to obtain correspondences between the feature
vectors and regions of the 2D image.\cite{karpathy2015deep} use external region proposals and learn an alignment model between image regions and sentences. In \cite{johnson2015densecap}, using a rich annotated dataset \cite{krishnavisualgenome} of image regions and corresponding textual descriptions, the problem of localization and description is addressed jointly. They propose a fully convolutional localization network (FCLN) for dense captioning.

In video captioning, the authors of \cite{lrcn2015} use a stack of two recurrent sequence models (LSTMs) to tackle the problems of activity recognition, image and video description. For image description, features extracted from a pre-trained CNN are fed directly into the LSTMs, while in the case of video description, first, a CRF model is used to obtain a distribution over subjects, verbs and objects. Then, the CRF responses are fed to the LSTMs to form a full sentence. Similarly, \cite{venugopalan:naacl15} uses a two stack LSTMs model for video description, but the visual information is encoded as mean-pooled CNN feature over the video frames. They also show improvement by transfer learning from the image domain (where more training data is available) to video. Approaches have  also been pursued using semantic classifiers responses for subjects, verbs and places in combination with LSTMs\cite{svp_lstm}.  Despite the fact that such approaches have achieved a significant improvement (under BLEU and METEOR metrics)  compared to previous template based approaches, they do not fully exploit the underlying structure of the video, nor do they attempt to explicitly identify (localize) the main actors or objects that correspond to the textual descriptions produced. 
\cite{temporal_att15} incorporate a spatio-temporal 3D CNN trained on video action recognition and use a temporal attention mechanism to select the most relevant temporal segments. \cite{hierarchical_rnn15} are interested in generating multiple sentences and discuss temporal and spatial attention mechanisms. The spatial elements are obtained by sampling image patches around a central actor, on datasets where this assumption holds. They use deep convolutional features like VGG \cite{Simonyan14c} and C3D \cite{c3d} to represent an image frame.
The purpose of these frame-level attention approaches is to guide the model towards different frames of the video at each time step (when a word is produced). Unlike these, our attention model focuses on a pool of spatio-temporal proposals and learns to choose the best spatial-temporal support for every word of the sentence.

\section{ Methodology}

\begin{figure*}
\begin{center}
\scalebox{1}{
   \includegraphics[width=\linewidth]{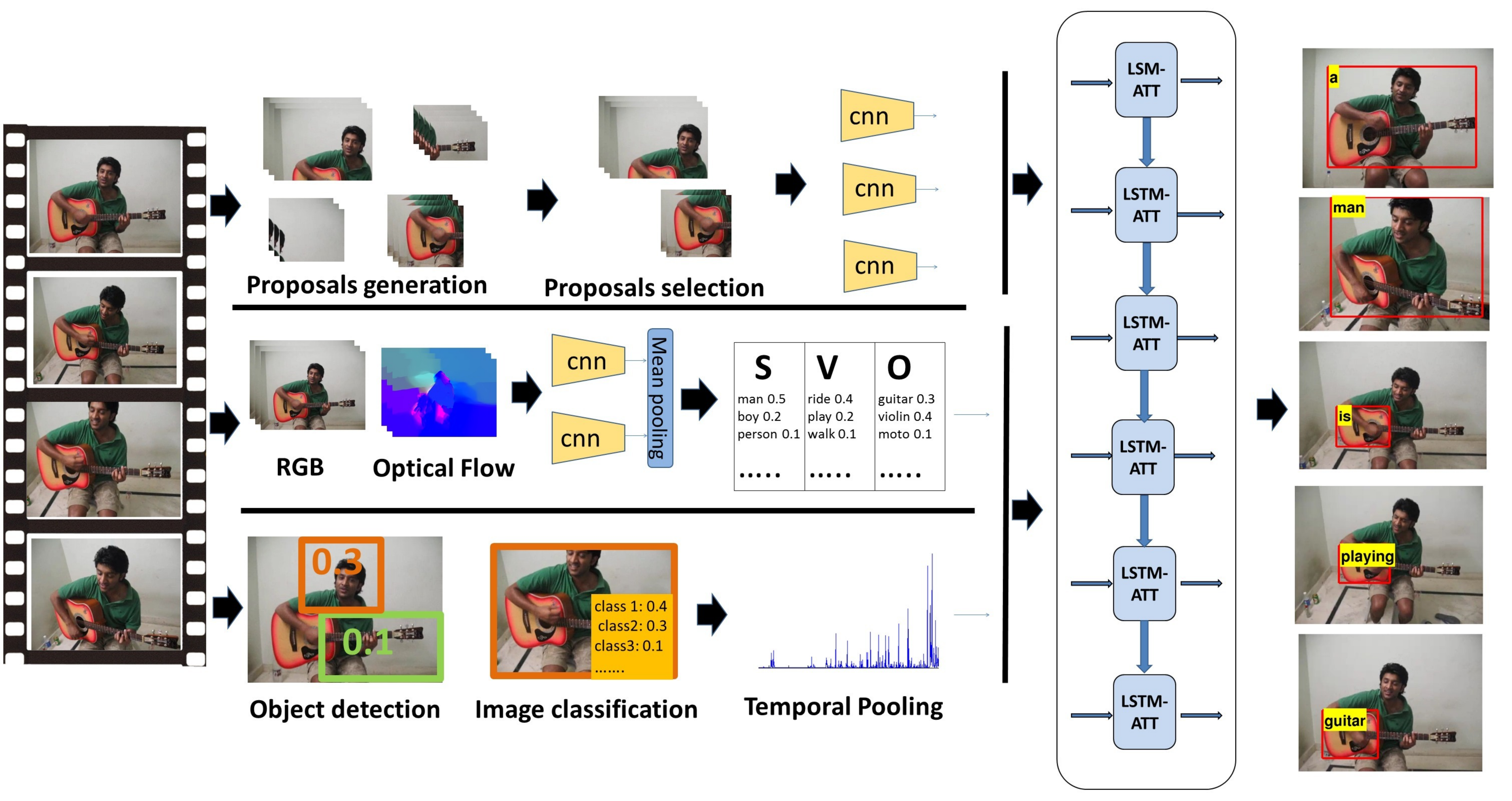}
   }
\end{center}
\caption{Overview of our approach. We build an attention mechanism on top of spatio-temporal object proposals and integrate it with state-of-the-art image classifiers, object detectors and recurrent neural networks (LSTM). The image classifiers together with learned high-level semantic features in the form of (Subject, Verb, Object) are provided as contextual features to the Attention-LSTM. Our model is able to visually ground each of the words from the sentence it generates, spatially and temporally, with no additional supervision.  }
\label{fig:overview}
\end{figure*}

Our approach to video captioning has two main components: first revealing the spatio-temporal visual support of words in video and then guiding the sentence generation process by including semantic information in the learning process. We integrate a soft-attention mechanism, operating over a pool of spatio-temporal proposals, into a state-of-the-art recurrent network. The joint model learns to produce semantically meaningful sentences while attending to different parts of the video. The semantic information is obtained in two ways: (a) by learning to predict subjects, verbs and objects (S,V,O) and (b) by using pre-trained state-of-the-art image classification and object detection models. An overview of our modeling and computational pipeline is shown in figure \ref{fig:overview}. Section \S\ref{sec:lstm} briefly introduces the recurrent model while section \S\ref{sec:attention} describes the attention mechanism. Learning the semantic concepts is explained in section \S\ref{sec:semantic} and the experimental details and results are given in section \S\ref{sec:results}. 

\subsection{Recurrent Networks for Video Captioning}\label{sec:lstm}

\begin{figure}
\begin{center}
\begin{tabular}{cc}

      \includegraphics[height=70pt]{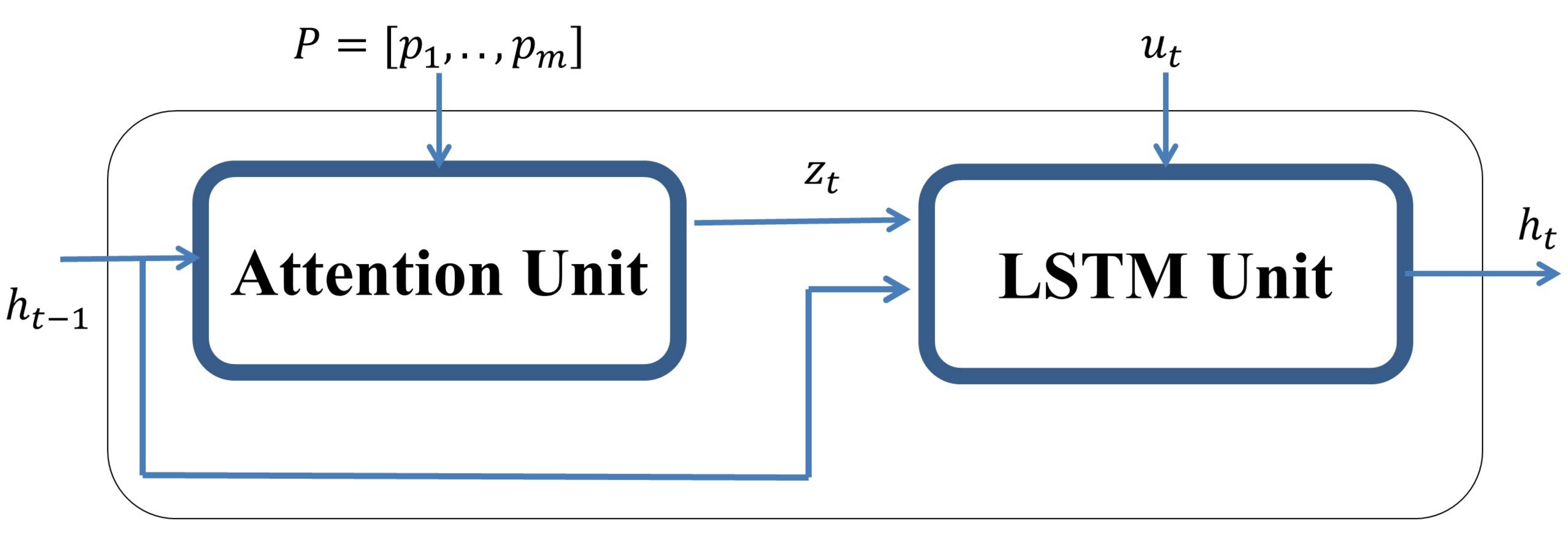} &

   \includegraphics[height=120pt]{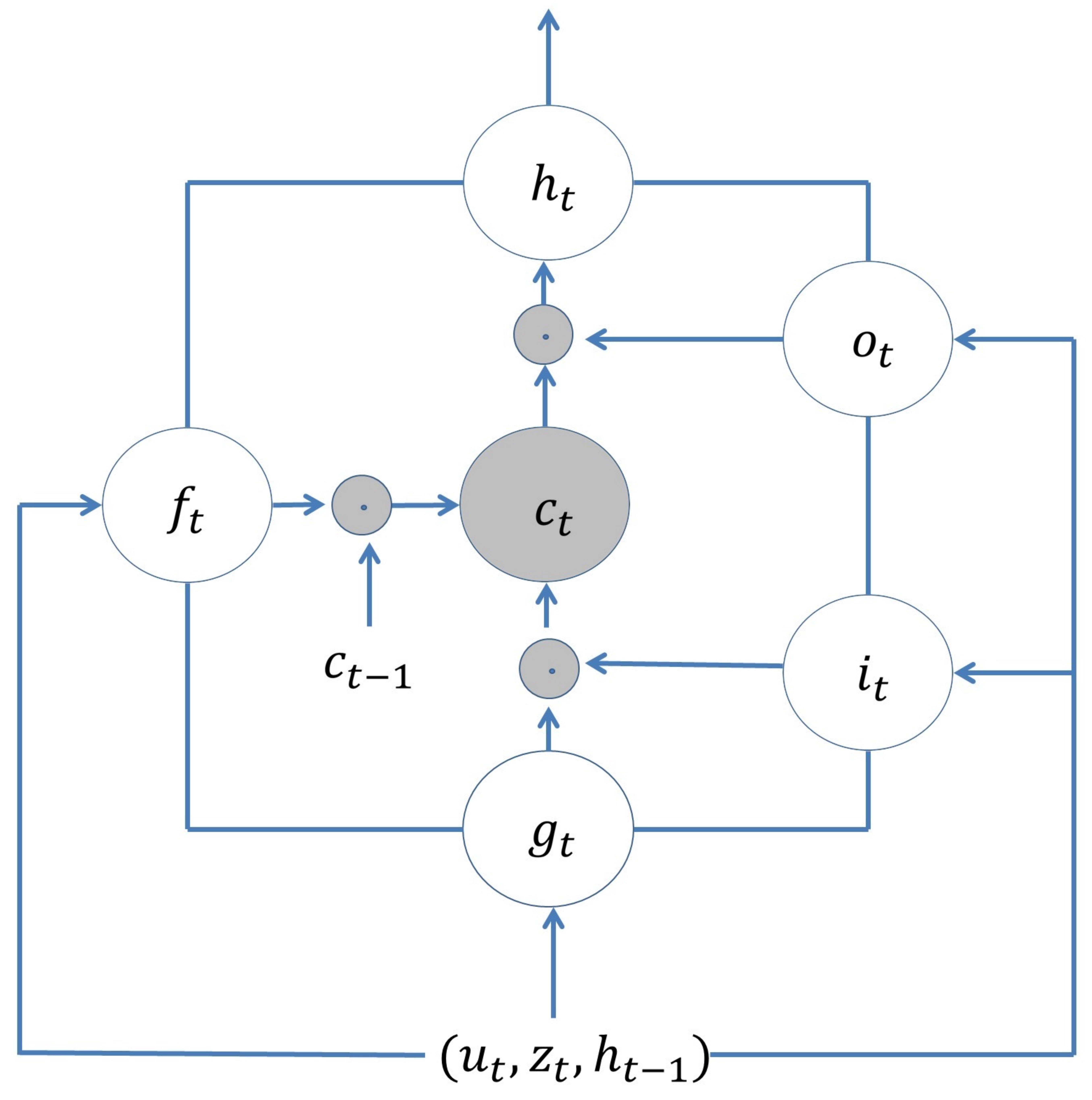} \\

     (a) & (b)
   \end{tabular}
\end{center}
\caption{a) Integration of an attention mechanism with LSTM in our model: at each timestep $t$, given the previous hidden state $h_{t-1}$ and $m$ spatio-temporal proposals, the Attention Unit outputs a weighted mean-pooled visual feature $z_t$. The LSTM Unit receives the visual feature $z_t$, the previous hidden state $h_{t-1}$ and the previously generated word $u_{t}$, and outputs the current hidden state $h_t$. b) Schema of LSTM Unit described in equation \ref{ec:lstm_ec}.}
\label{fig:lstm}
\end{figure}

Recurrent Neural Networks (RNNs) make use of sequential information and learn temporal dynamics by mapping a sequence of inputs to hidden states and then learn to decode the hidden states into a series of outputs. Their major drawback, however, is the \textit{vanishing gradient} which makes it difficult to learn long-range dependencies that exist in the input sequences \cite{Hochreiter01gradientflow}. A solution to this issue is to incorporate explicit unit memories, controlled by gates deciding at each step which information should be passed on and which one should be forgotten. Those units, known as long short-term memory (LSTM) units \cite{lstm}, have proven to perform well for machine translation \cite{sutskever2014sequence} and have recently been used for both image \cite{Xu2015show,lrcn2015} and video captioning\cite{s2vt:iccv15,MM-VDN,venugopalan:naacl15}. A schema of the LSTM unit (introduced in \cite{zaremba2014learning}) used in our experiments is shown in figure~\ref{fig:lstm}. The LSTM unit consists of a memory cell, $c_t$, that encodes the information transmitted from previous units up to current step and \textit{gates}
deciding how the information in the memory cell is updated and what the output should be. The input $i_t$, forget $f_t$, and output $o_t$ gates are sigmoid functions that decide how much to consider from the current input($u_t$), what to retain from the previous cell memory ($c_{t-1}$)  and how much information from the memory cell to be transferred to the hidden state ($h_t$). The updates at time step $t$ given textual input $u_t$, visual input $z_t$, the previous hidden state $h_t$ and the previous memory cell $c_{t-1}$ are given by the following equations:

\begin{equation}
\label{ec:lstm_ec}
\begin{aligned}
i_t &= \sigma(W_{xi}u_{t}+W_{hi}h_{t-1}+W_{zi}z_{t}+b_i) \\
f_t &= \sigma(W_{xf}u_{t}+W_{hf}h_{t-1}+W_{zf}z_{t}+b_f) \\
o_t &= \sigma(W_{xo}u_{t}+W_{ho}h_{t-1}+W_{zo}z_{t}+b_o) \\
g_t &= \phi(W_{xg}u_{t}+W_{hg}h_{t-1}+W_{zg}z_{t}+b_g) \\
c_t &= f_t\odot c_{t-1} +i_t\odot g_t \\
h_t &= o_t \odot \phi(c_{t})
\end{aligned}
\end{equation}

\subsection{Attention-based LSTM} \label{sec:attention}

\mypar{Soft-Attention Mechanism}. We incorporate a soft-attention mechanism into the LSTM in order to allow the model to selectively focus on different parts of the video each time it produces a word. Inspired by the attention mechanism that exploits the spatial layout of an image \cite{Xu2015show}, a few recent methods have attempted to exploit the temporal structure of the video by learning how to assign different weights to frames in a video sequence \cite{temporal_att15,hierarchical_rnn15}. These methods, however, do not localize objects in images, and thus the same frame can offer support to very different words in the output. This approach can work when a video selectively focuses on individual objects at a time and thus in a single frame very few objects of interest are present. However, in most videos, there are multiple actors and objects present in a frame. To address this problem, we allow the sequence model (LSTM) to choose where to focus among a pool of coherent spatio-temporal proposals at each time step $t$. Thus, the model is able to indicate what is the \textit{localized} visual support used to produce a particular word from the video description.

 Considering $P = [p_1,p_2,\ldots,p_m]$ the temporal feature vector  where $m$ is the number of proposals and $p_i$ the descriptor for the $i$-th proposal, the LSTM  learns at each time step  a series of $m$ weights $\beta_{ti}$ such that the final encoding of the visual input is
 \begin{equation}
     z_{t} = \sum_{i=1}^{m}{\beta_{ti}p_i} ,\,\,\,\,\textnormal{with} \,\,\,\, \sum_{i=1}^m{\beta_{ti}}=1
 \end{equation}
 
The weights $\beta_{ti}$ represent the importance of the $i$-th proposal for generating the word at the current time step, given the previously generated words. First, for each proposal we learn a score $\epsilon_{ti}$ based on its visual feature $p_i$ and the previous hidden state $h_{t-1}$, which is given by  

 \begin{equation}
\epsilon_{ti} = W_{ph}\phi(W_pp_i+W_hh_{t-1}+b_{ph}) 
\end{equation}
where $\phi$ is the hyperbolic tangent and $W_{ph}, W_p, W_h, b_{ph}$ are parameters to be learned. Those scores are then normalized to obtain the $\beta_{ti}$ weights: 

\begin{equation}
\beta_{ti} = \frac{e^{\epsilon_{ti}}}{\sum_{i=1}^{m}{e^{\epsilon_{ti}}}}
\label{ec:weights}
\end{equation}
A schematic view of the Attention-LSTM model is shown in figure~\ref{fig:lstm}. At testing time, by inspecting the weights in decreasing order, we can interpret the visual information preferred (selected) by the model when producing a particular word in the textual description of the video.
\subsection{High-Level Semantic Description} \label{sec:semantic}

Generating a textual description of a video requires identifying the actors and their interactions and then constructing a grammatically well-formed sentence. For this purpose, 
in order to generate a human-like textual description of a video, we first represent a video in the form of a Subject(S), a Verb(V) and an Object(O) (similarly to earlier works \cite{youtube_to_text}). We then integrate this representation with state-of-the-art recurrent models, along with spatio-temporal localization processes and object detection and classification information.

\mypar{SVO representation and vocabulary construction}. In order to learn a semantic high-level representation for each video, we represent a sentence in a compact and simplified manner that preserves its main idea  by extracting a (S,V,O) tuple - e.g.  the sentence \textit{A cat plays with a toy} is represented as (cat, play, toy)). We initially used the SVO vocabulary proposed in \cite{youtube_to_text}, but found it to be too small (only 45 subjects, 218 verbs and 241 objects) and too semantically restrictive (e.g. no different words for \textit{man} and \textit{woman}, as it only contains \textit{person}). We mine the intermediate concepts differently from \cite{youtube_to_text}, such that our vocabulary is richer and with fewer constraints. The important changes we made in the way we build the vocabulary are: 1) considering both the indirect and direct objects when parsing the sentences as opposed to only the direct objects, 2) not grouping words into very general classes, and 3) an S, V or O is added to the final vocabulary if it is mentioned in at least two different sentences in any given video. We use the parser available from \cite{cornelp} to extract from each sentence a subject, verb and an object. Our final vocabulary set is a tuple $\mathcal{D} = \{\mathcal{S}, \mathcal{V}, \mathcal{O}\}$ of the corresponding vocabularies for each sentence part and it has a considerably larger size: 246 subjects, 459 verbs and 801 objects.
 
 \mypar{SVO Classification}. We treat the three vocabularies separately and use Least Squares Support Vector Machine (LS-SVM) as a classifier in a one-vs-all approach. Note that an input video can have multiple labels from each vocabulary (e.g. a video can have labels 'cat', 'animal', 'kitten' for the subject class). We use LS-SVM  because it provides a closed form solution both for the leave-one-out prediction and the prediction error via the block inversion lemma \cite{cawley2006leave}.
We represent a video as a classifier response vector for all the classes in the combined vocabulary $\mathcal{D}$. Training videos use the leave-one-out prediction and testing videos use the prediction based on the classifiers learned on the whole training set. This is different from \cite{svp_lstm}, where classifiers scores are learned in the same way for training and testing. The dataset used in \cite{svp_lstm} has around 56k training examples, with roughly the same vocabulary size as ours. Given that our dataset has a much smaller number of training videos ($\approx$ 1,300), we argue that LS-SVM is a better option; we can tune parameters without relying on a separate validation set (further decreasing the amount of label data) and we can better simulate the testing conditions. Our choice is also supported by the increase in classification accuracy when compared to other methods using the same vocabulary (see \S\ref{subsec:SVO_results}).
 


\section{Experimental Details} \label{sec:results}

\mypar{Dataset Description}. We perform our experiments on the YouTube dataset \cite{chen_acl11} which consists of 1,967 short videos (between 10s and 25s length) collected from YouTube that usually depict only one main activity. Each video has approximately 40 human-generated English descriptions collected through Amazon Mechanical Turk. We use the same splitting into train (1,197 videos), validation (100 videos) and test (670 videos) subsets as previous methods, so that our results are directly comparable to them.

\mypar{Evaluation Measures}. We report our results under BLEU \cite{bleu} and METEOR \cite{meteor} metrics which were originally proposed for the evaluation of automatic translation approaches and have also been adopted by previous works in video and image captioning. BLEU@n computes the geometric average of the n-gram precision between generated and reference sentences. METEOR computes an alignment score between sentences by taking into account the exact tokens, the stemmed ones and semantic similarities between them. We use the evaluation software provided by \cite{coco_eval} which we adapt to our dataset.

\subsection{Spatio-Temporal Object Proposals}

We use the method from \cite{oneata_eccv2014} to gather a pool of spatio-temporal object proposals. We split each video into parts using a shot boundary detection method \cite{lienhart1998comparison}. Around 1,000 spatio-temporal proposals are extracted separately for each sub-video and together they form the pool of proposals for the whole video. We filter out the proposals that have a small spatial or temporal extent. To diversify the pool and to eliminate very similar proposals, we keep only those that have low IoU scores with each other. From the pool of proposals, we are interested mainly in those that could be attached a semantic meaning. Thus, we sort the proposals according to a semantic measure based on two scores and retain the top $m$. In our experiments we set $m=20$. The first score is obtained by running the image classification CNN VGG-19 from \cite{Simonyan14c}(trained on 1.3M images from ImageNet Large-Scale Visual Recognition Challenge (ILSVRC) \cite{ILSVRC15}) on every bounding box of each proposal, retaining the maximum activation in each frame among the 1,000 classes and averaging across all frames. The second score is obtained by running the 20-class object detector from \cite{renNIPS15fasterrcnn} on every frame of each video. For each frame of the proposal, we compute the maximum detection score (multiplied by the IoU between the bounding box of the detection and the spatial extent of the proposal) and then we average the scores across all proposal frames. The final proposal score is the average of the image classification and detection scores.

\begin{figure*}
\begin{centering}
\begin{tabular}{ cccc}
\includegraphics[height=65pt]{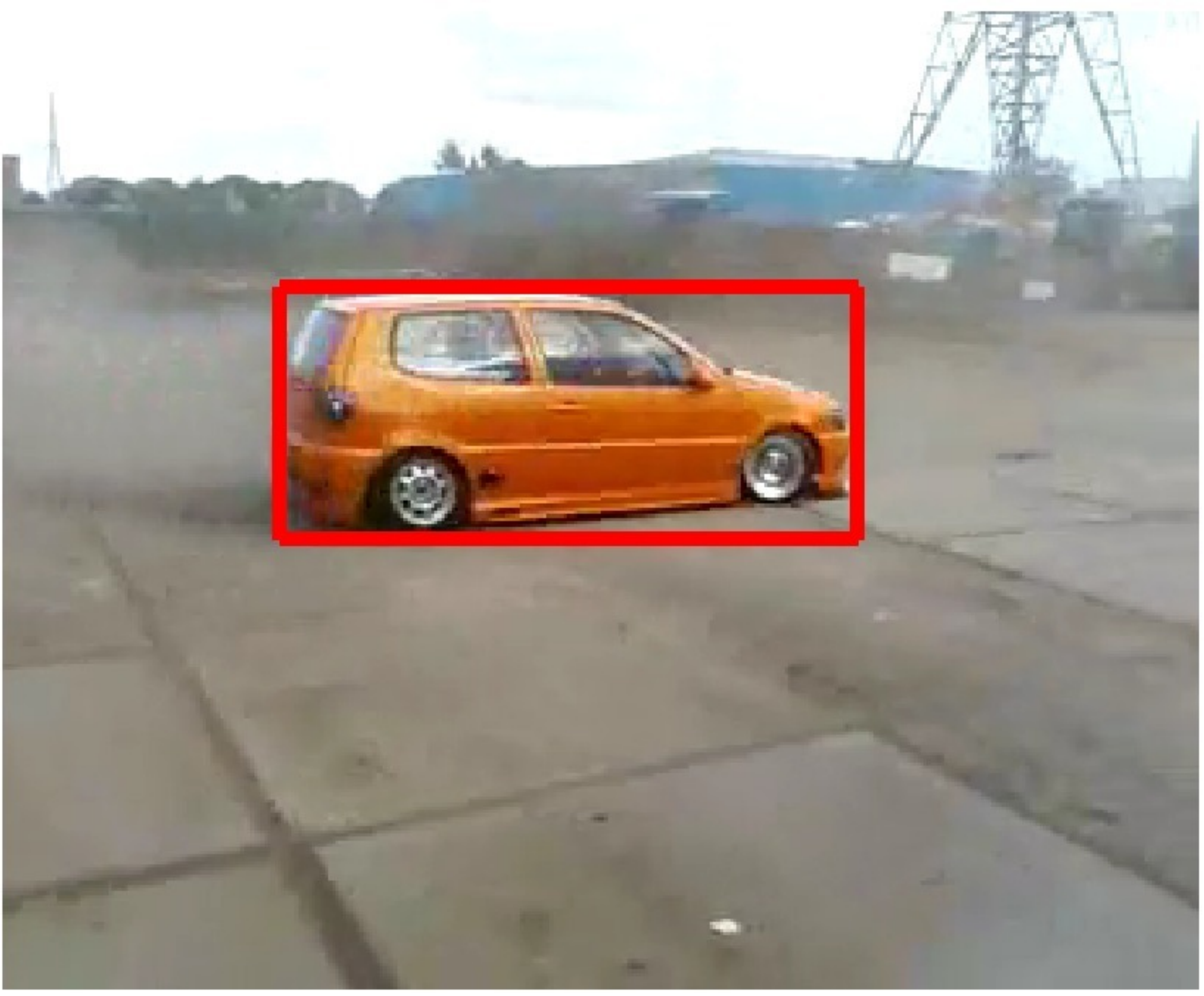}  &
 \includegraphics[height=65pt]{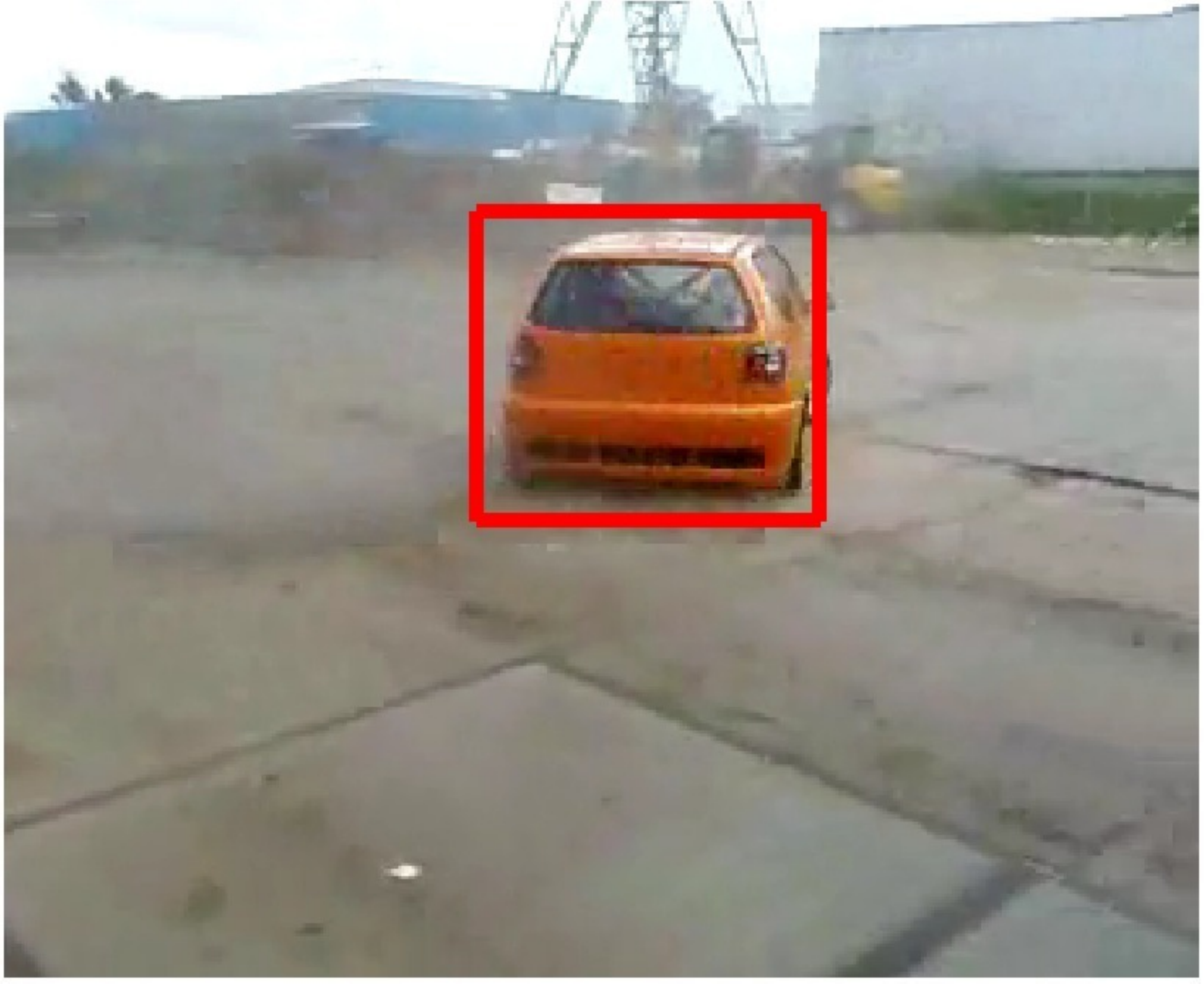} &
 \includegraphics[height=65pt]{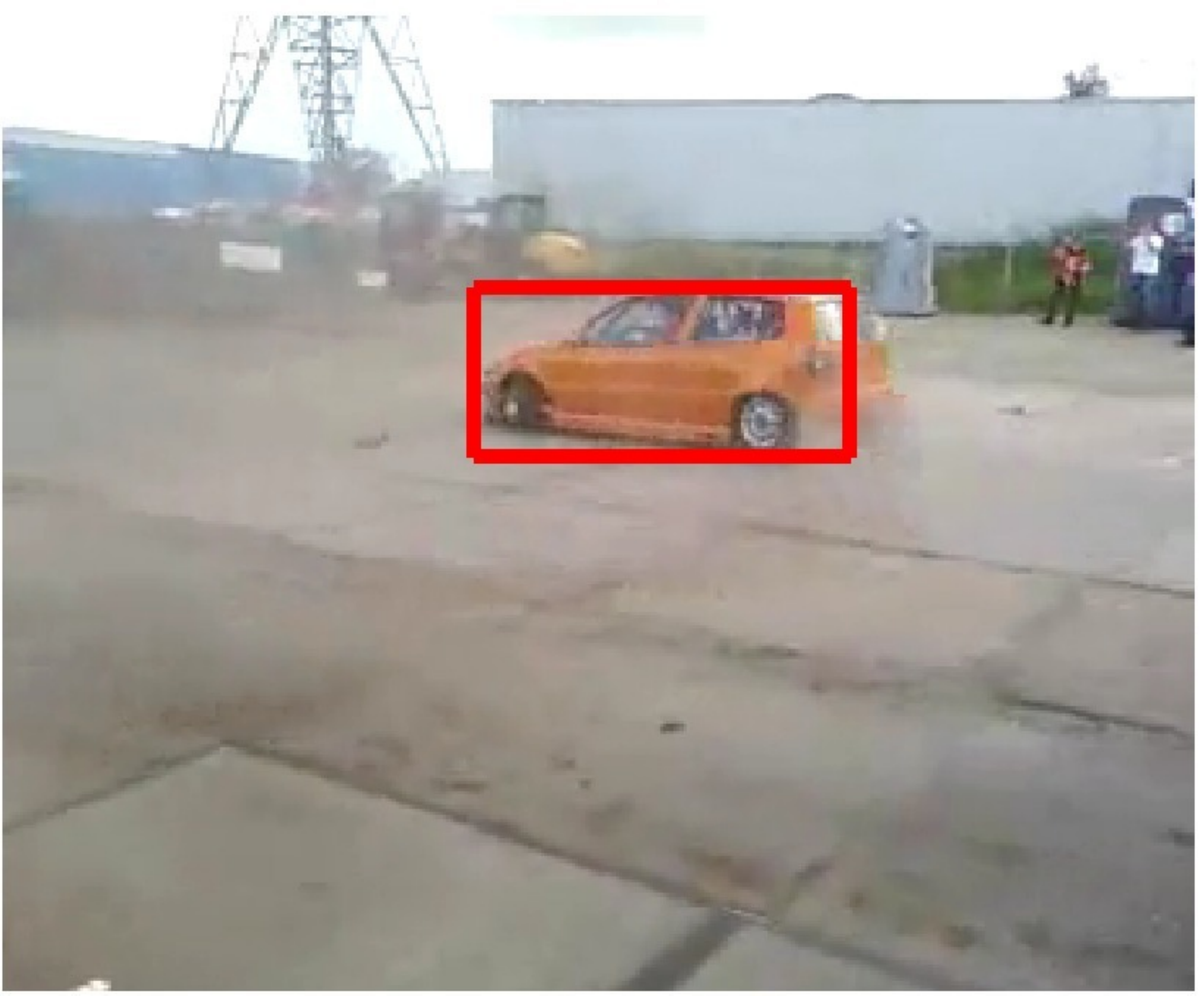} &
 \includegraphics[height=65pt]{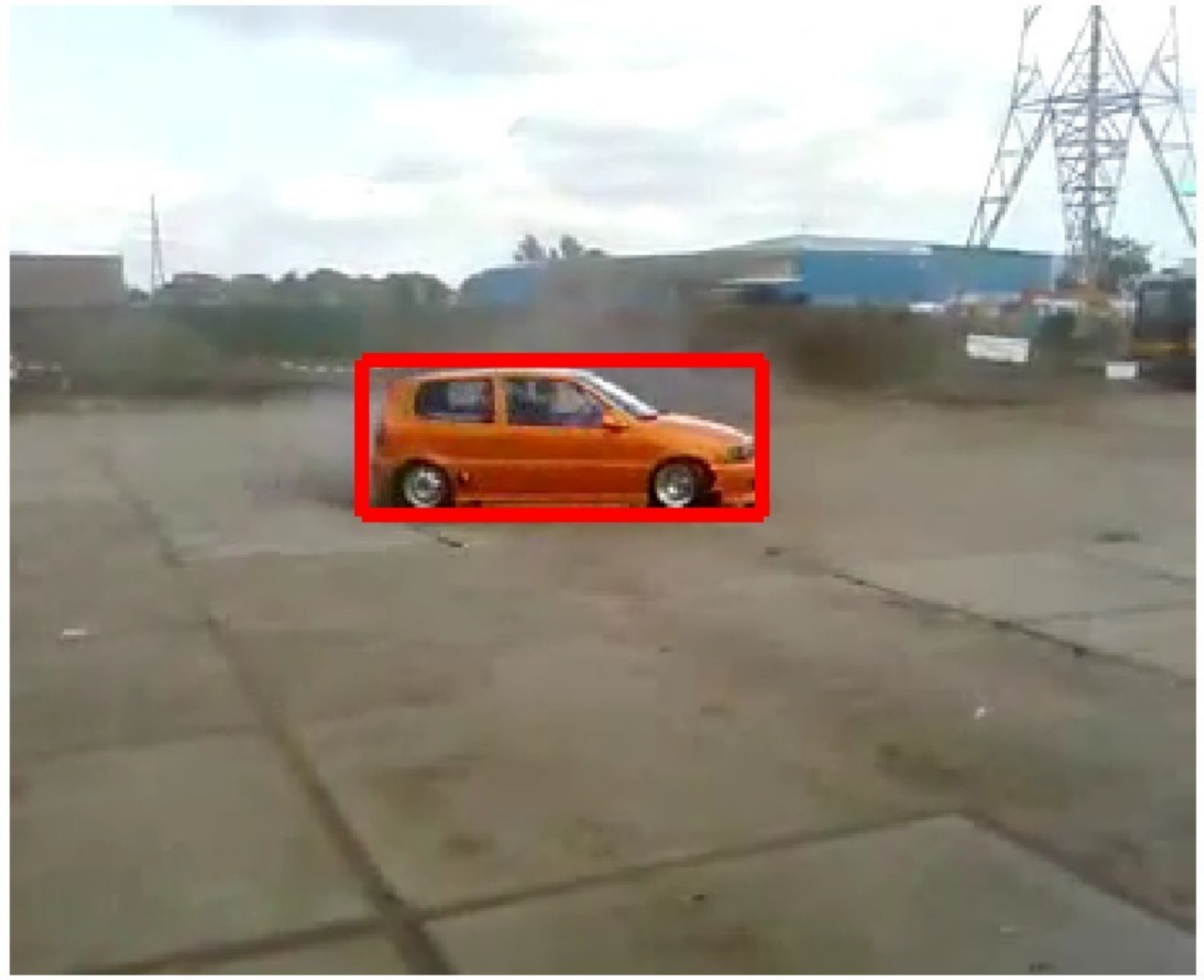} \\
 
 \includegraphics[height=45pt]{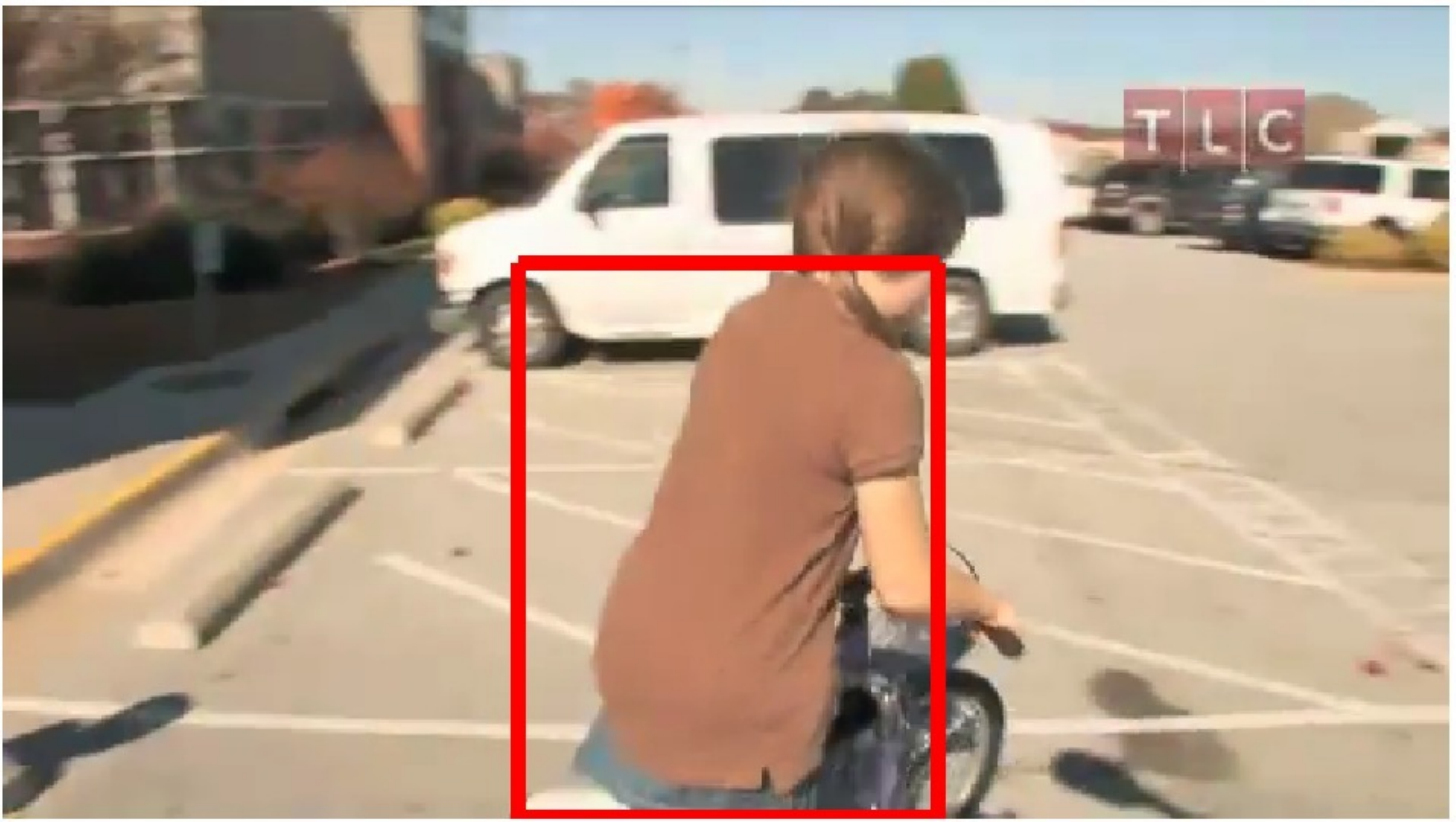}  &
 \includegraphics[height=45pt]{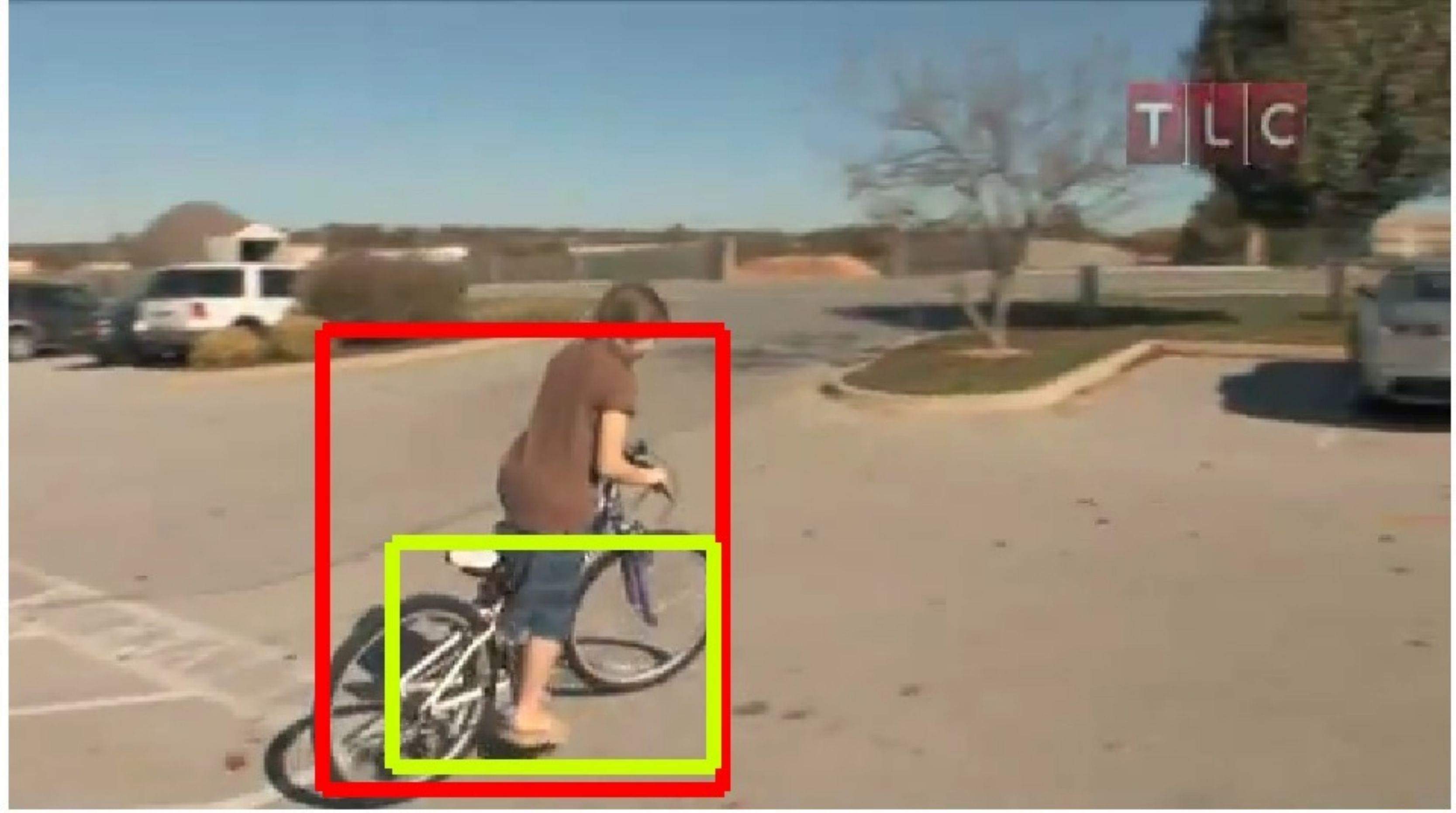} &
 \includegraphics[height=45pt]{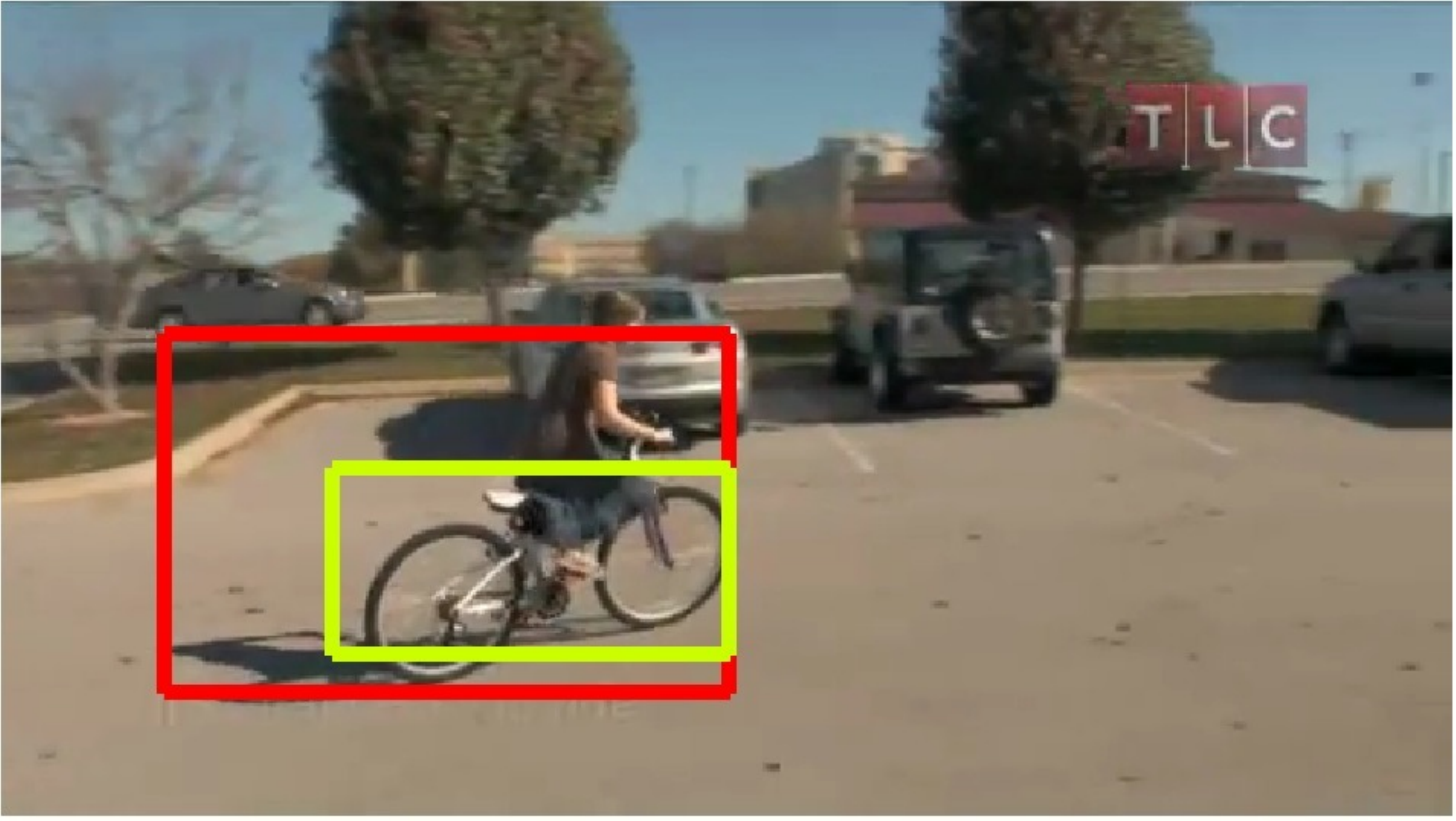} &
 \includegraphics[height=45pt]{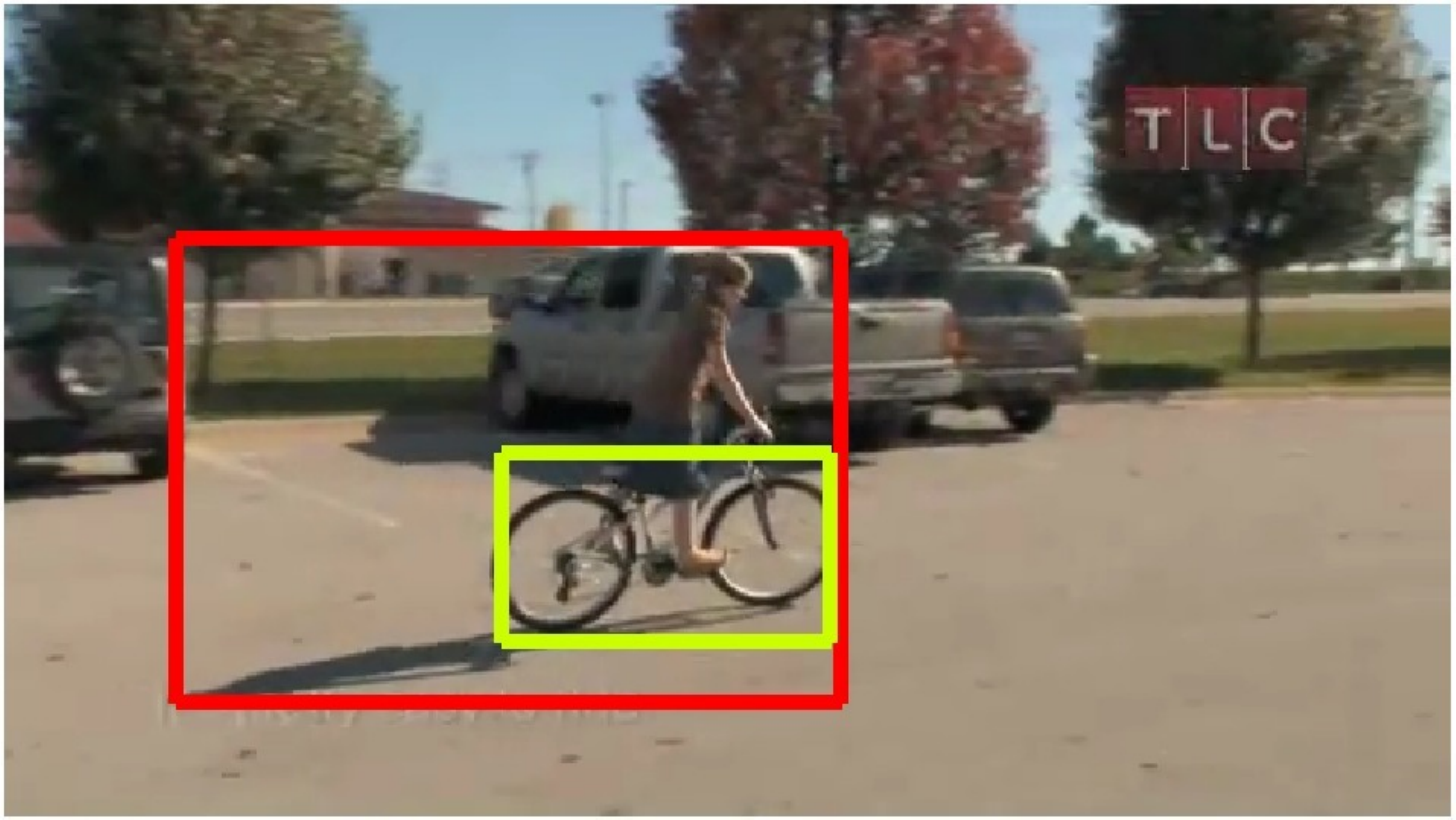} \\
 
  \includegraphics[height=60pt]{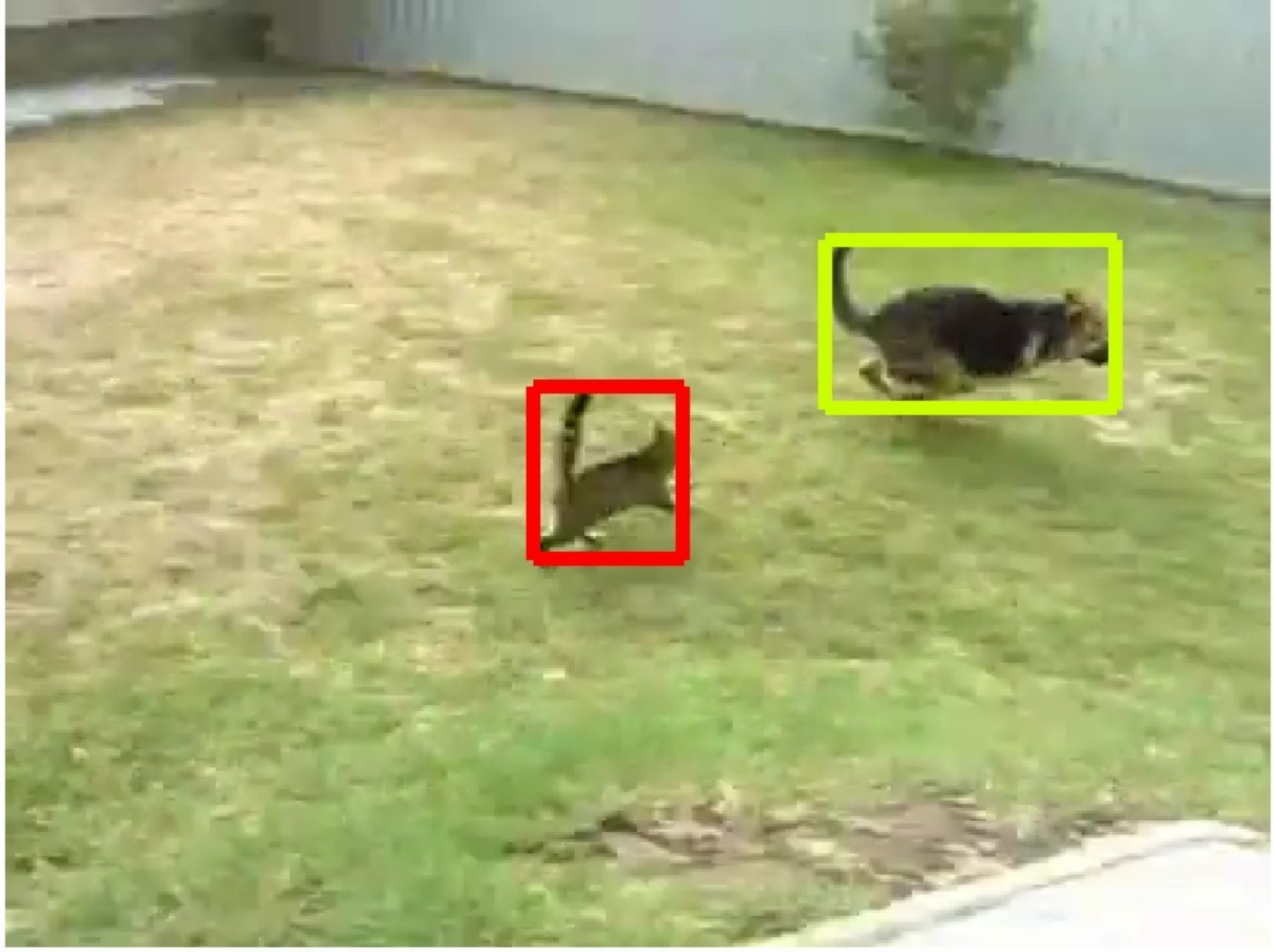}  &
 \includegraphics[height=60pt]{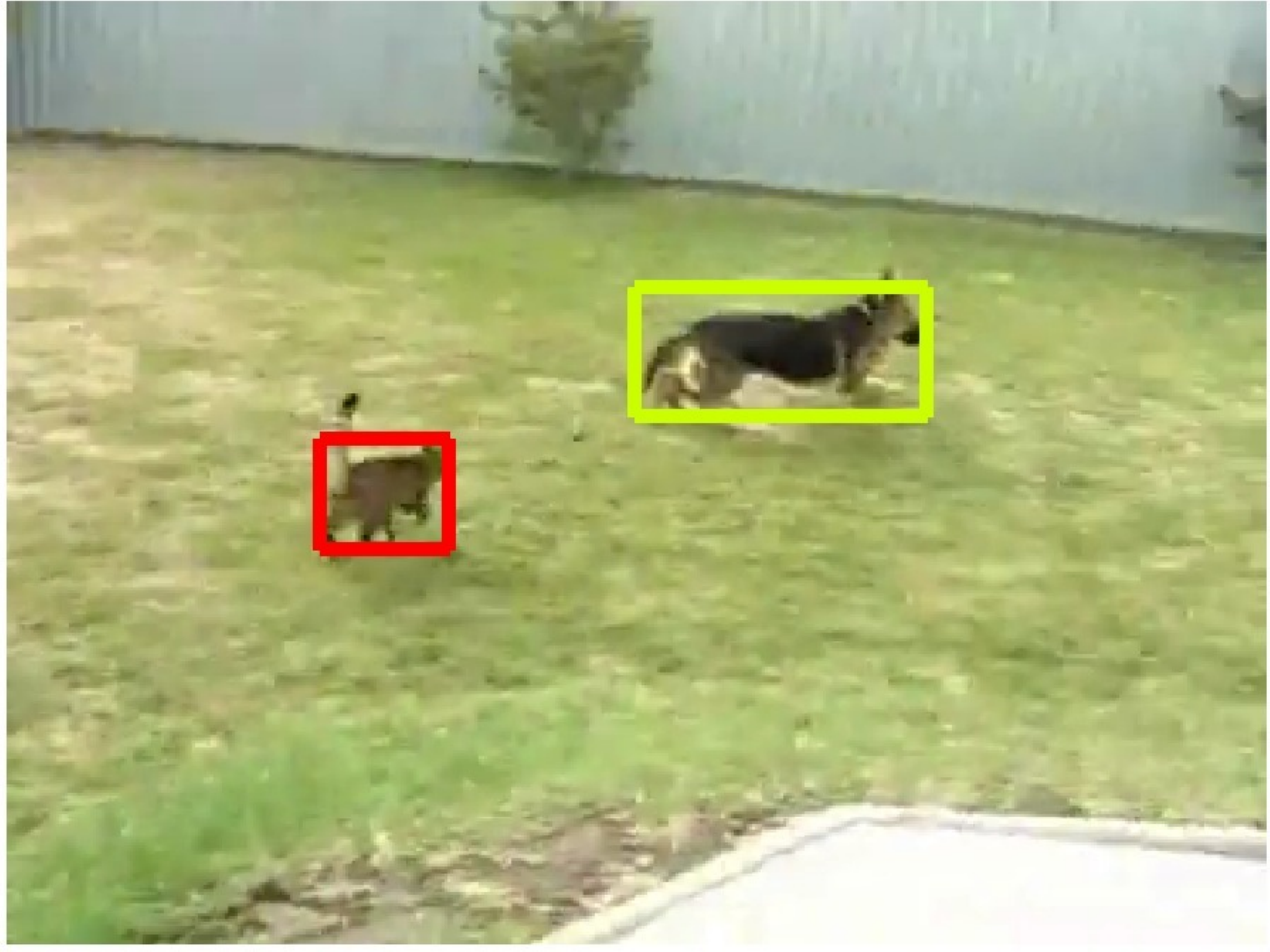} &
 \includegraphics[height=60pt]{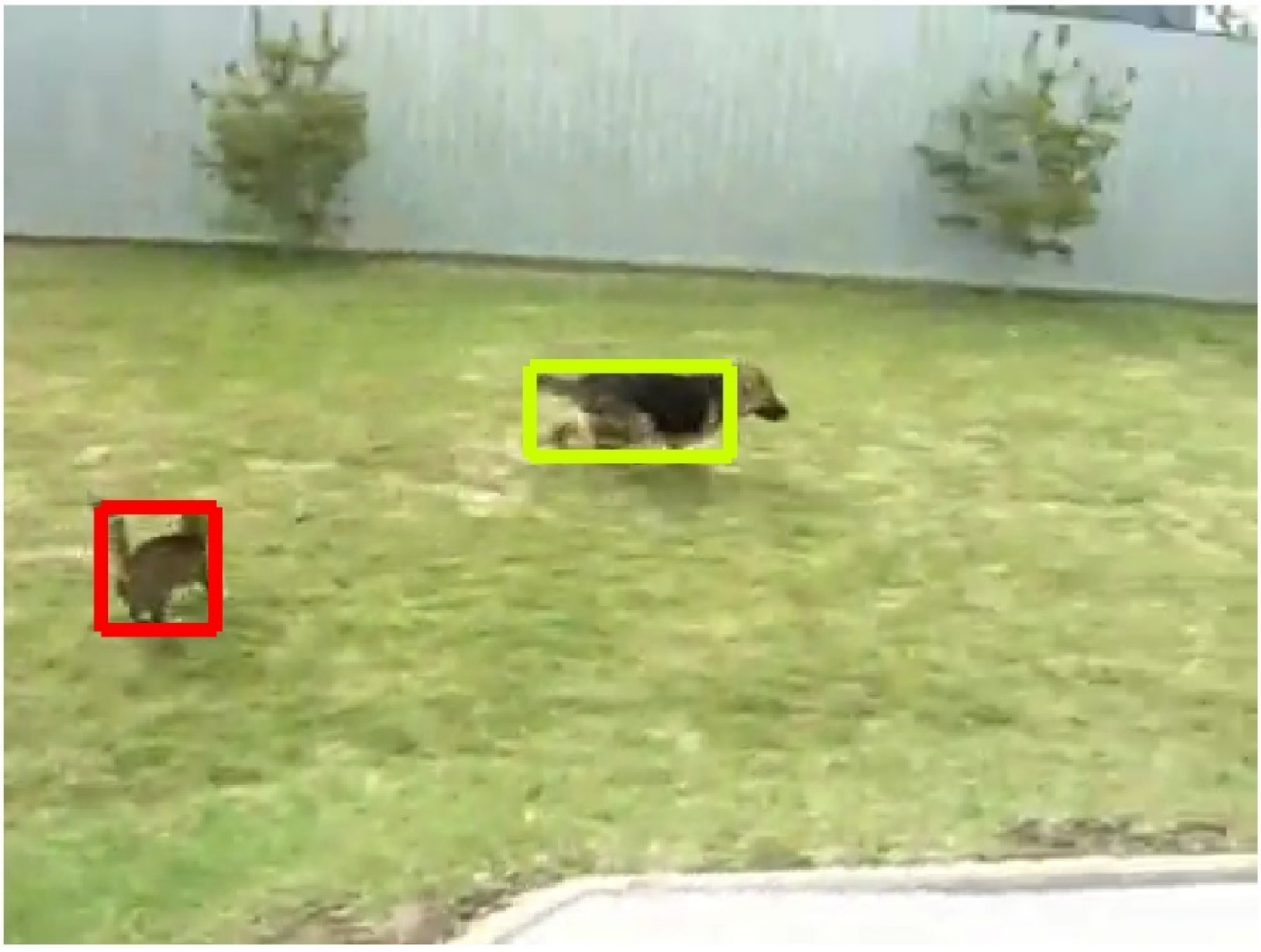} &
 \includegraphics[height=60pt]{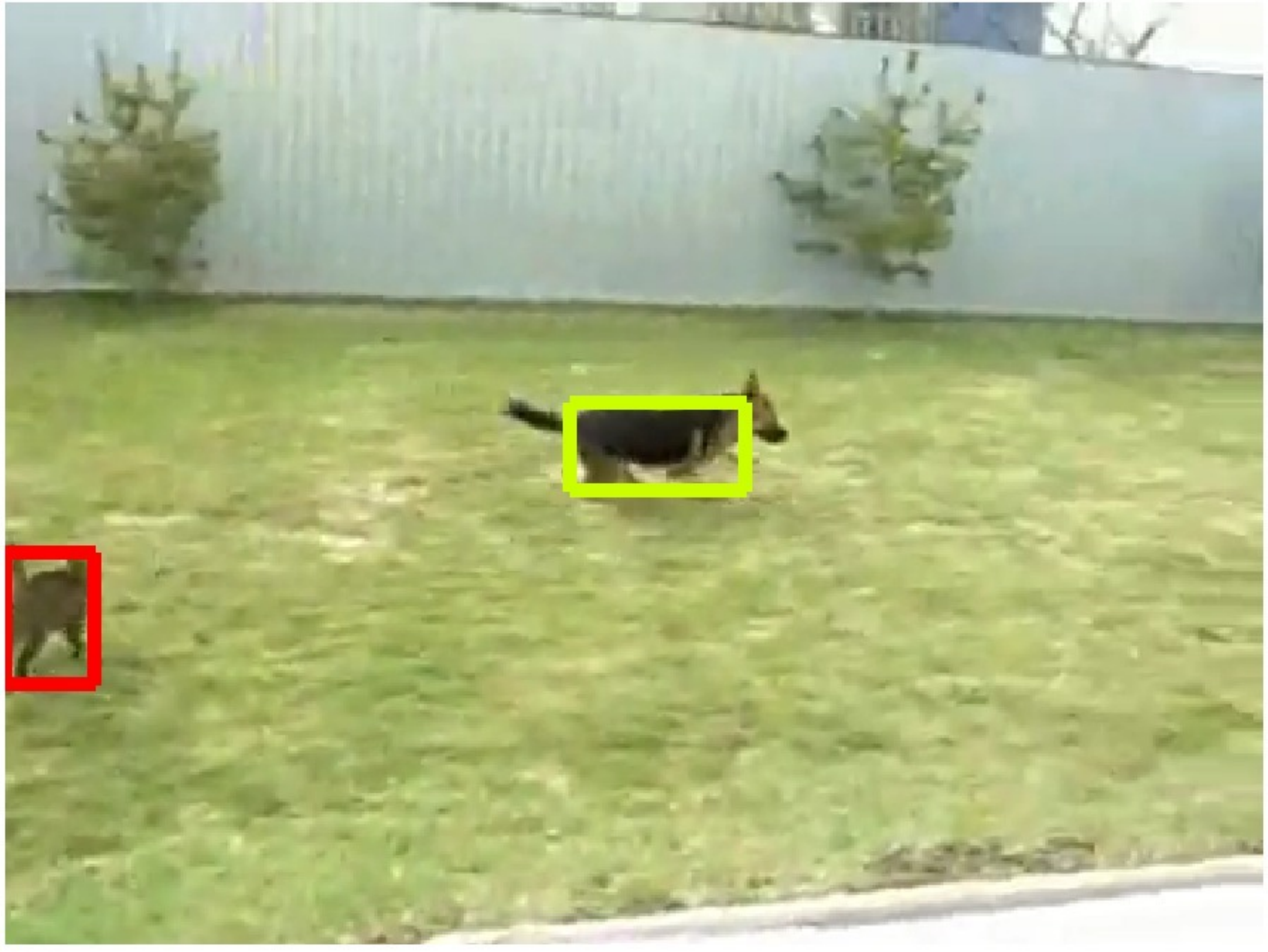}

\end{tabular}
\end{centering}
\caption{Examples of selected object proposals. For each video we generated a large pool of spatio-temporal object proposals and then learned to automatically select those that are most likely to overlap with easily recognisable semantic categories.}
 
\label{fig:example_proposals}
\end{figure*}

Given a proposal, for each of its bounding boxes in the video frames, we extract the output of the $fc_7$ layer of the VGG-19. The feature descriptor for a proposal is obtained by mean-pooling over all the bounding boxes. We represent a video by $m$ such descriptors corresponding to  best scoring $m$ spatio-temporal proposals and we refer to this $m\times D$ descriptor as the \textit{temporal visual feature}. In practice, for some videos, the number of duplicate proposals is very large, and the final number of proposals can be less than $m$. Since the network takes as input a fixed sized array, for the videos that do not have at least $m$ spatio-temporal proposals we pad the feature matrix to obtain a fixed size descriptor. We mark the padding proposals so that they will be ignored in the learning and testing processes. Examples of the selected proposals can be seen in figure~\ref{fig:example_proposals}.


\subsection{Attention LSTM with Spatio-Temporal Object Proposals}

\mypar{Vocabulary}. We use all the words in the training sentences without performing any pre-processing step. This results in a vocabulary of size 9,070. Similarly to previous works, we represent the words as one-hot vectors, set the maximum sentence length to 20 and mark with special characters the beginning and end of a sentence.  

\mypar{Training}. We implemented our attention mechanism using the Caffe \cite{caffe} framework, integrating it on top of the LSTM provided by \cite{lrcn2015}. We refer to our proposed LSTM model which uses an attention mechanism over spatio-temporal object proposals as LSTM-ATT. The training phase in LSTM-ATT is a sequential one: at each time step $t$ the unit is given the temporal feature $P$ (representing the $m$ proposals), the embedded vector $u_{t}$, corresponding to the previous ground-truth word $w_{t-1}$, and the previous hidden state $h_{t-1}$. The output hidden-state $h_t$ (see figure~\ref{fig:lstm}) is then used to predict a distribution $\mbox{P}({w_t})$ over the words in the vocabulary. We use the softmax loss and a dropout of 0.5 to avoid over-fitting. We train our models for a maximum of 128 epochs and use the validation set to choose the best iteration for each of the two metrics, BLEU and METEOR.

\mypar{Inference}. Inference is also performed in a sequential manner: given $m$ proposals and the previous emitted word at time $t-1$, sampled from $\mbox{P}({w_{t-1}})$, the model generates the current word until the special character for end of sentence is met or the maximum length of a sentence is reached. We perform the sampling using beam search with beam size 20. Because we noticed that standard beam search implementations sometimes tends to end sentences too early, we modified it to force longer sentences to be produced (at least 4 words). 

 \subsection{Integrating Contextual Semantic Features} 
 \label{subsec:SVO_results}
\mypar{Semantic SVO Representation}. In order to obtain the SVO responses, we use the LS-SVM described in section \S\ref{sec:semantic} and consider 3 different classification problems, one for S, one for V and one for O. Each video can then be described by concatenating the responses to these three classification problems. We use different features, depending on the part of sentence we want to classify. For the subject and object classes, we use the VGG-19 of \cite{Simonyan14c}, extract feature responses from the fc7 layer for each frame of the video and then perform mean pooling. For the verb class we use two types of motion features: the trajectory features of \cite{wang2011action} and the motion-CNN features of \cite{gkioxari2015finding}, again followed by mean pooling. For S, V and O we obtained the following classification accuracies, respectively: 62.5\%, 40.9\% and 28.30\%. Among the 3 classification problems, the results on  O are the lowest  since the number of classes in the object vocabulary is the largest: 801 compared with 264 for S and 459 for V. Also, the objects have a smaller spatio-temporal extent in video as they usually represent the objects manipulated by a person or animal (e.g \textit{onion}, \textit{ball}, etc). Since we considerably augmented the vocabulary, our classification results on this vocabulary do not compare directly with previous methods. However, we run our method on the initial proposed vocabulary \cite{youtube_to_text} and show results in table~\ref{tab:svo_accuracy_most_common}, against the most common (S,V,O) triple found in human annotations for each video. We show state-of-the-art results for S and O and a slightly lower accuracy than state-of-the-art for V. Notice that the methods marked with (*) generate a full sentence using a recurrent neural network and then extract the S, V and O using a dependency parser. Our aim is to use these intermediate semantic concepts as features to guide and ground our LSTM attention model (LSTM-ATT model) in the sentence generation process.

 \begin{table}
\begin{center}
\scalebox{1}{
\begin{tabular}{lccc}
\hline
 \textbf{Model} & \textbf{S\%} & \textbf{V\%} & \textbf{O\%} \\
\hline
\hline
HVC \cite{hvc-fgm} & 76.5 & 22.2 & 11.9 \\
\hline
FGM \cite{hvc-fgm} & 76.4 & 21.3 & 12.3 \\
\hline
JointEmbed \cite{aaai15} & 78.2 &24.4 & 11.9\\
\hline
(*) LSTM-E (VGG+C3D) \cite{pan2015} & 80.4 & \textbf{29.8} & 13.8\\
\hline
(*) LSTM-YT-coco  \cite{venugopalan:naacl15} & 76.0 & 23.3 & 14.0\\
\hline
(*) LSTM-YT-coco+flicker  \cite{venugopalan:naacl15} & 75.61 & 25.3 & 12.4\\
\hline
LS-SVM(ours) & \textbf{83.6} & 28.1 & \textbf{23.1}  \\
\hline
\end{tabular}
}
\end{center}
\caption{Binary SVO accuracy computed against the most common (S,V,O) triple provided by humans. Entries marked with (*) first obtain a sentence describing the whole video and then mine the (S,V,O), whereas the others perform a classification over the S, V and O vocabularies. }
\label{tab:svo_accuracy_most_common}
\end{table}

\mypar{Semantic Representations}. Apart from the SVO responses obtained by training using only the YouTube dataset, we also extract high level semantic features using state-of-the-art image classification and detection models. More precisely, we run the VGG CNN from \cite{Simonyan14c} in each frame of the video and obtain the 1,000 dimensional score vector representing the classification responses over the 1,000 classes from the ImageNet classification dataset. To obtain a semantic representation of the video, we experimented with both average and max pooling over the frames and noticed that average pooling performs better in our experiments. We also run the 20 class object detector of \cite{renNIPS15fasterrcnn} in each frame of every video and compute a 20-dimensional descriptor. For each class, we retain the detection response scores in every frame then perform temporal pooling across a window of 25 frames. The final score for a class is the maximum  of the  temporal pooled scores for that class. The temporal pooling  ensures that the detection observed is stable and lasts for at least 1 second. The maximum over such detection scores represents the confidence in having seen a particular object in video.
 
\begin{figure}
\begin{center}
\begin{tabular}{cc}
    
      \includegraphics[height=65pt]{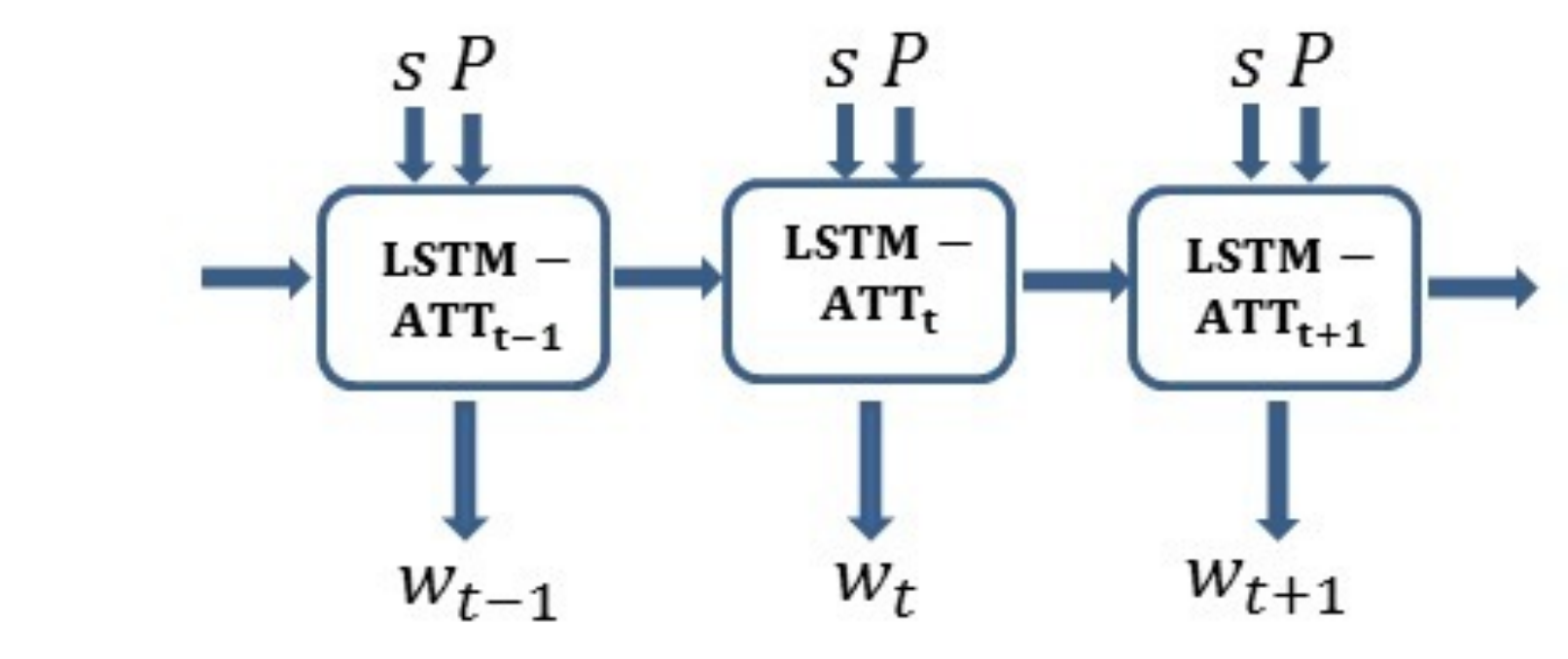} &
      \includegraphics[height=100pt]{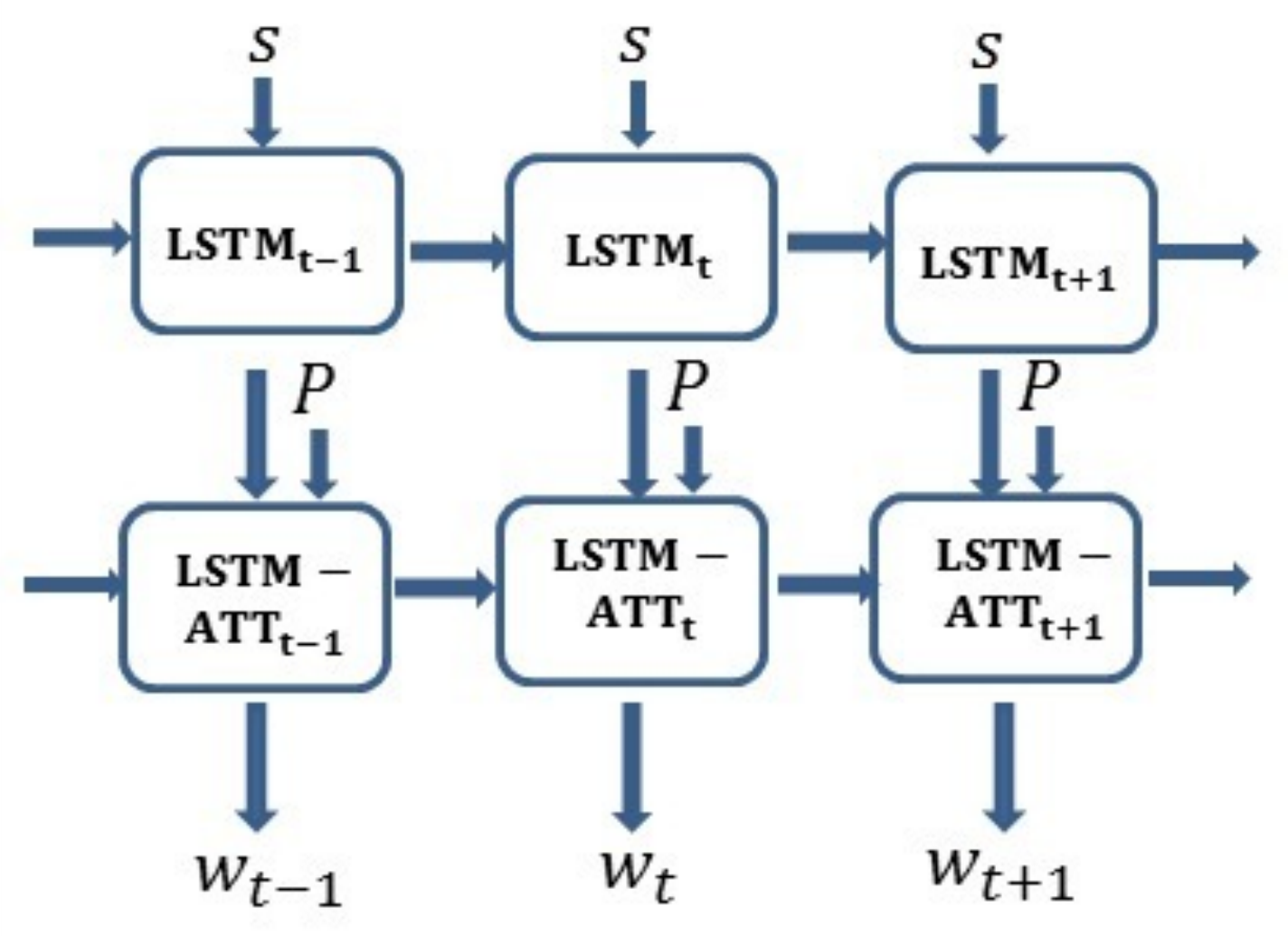} \\

     (a) & \hspace{20px}(b)
   \end{tabular}
\end{center}
\caption{Integration of semantic features with LSTM-ATT.  a) LSTM-ATT(SEM): both semantic features $s$ and temporal features $P$ are processed by the LSTM-ATT unit. b) LSTM2-ATT(SEM): we stack a LSTM unit that processes only the semantic input $s$ on top of the LSTM-ATT.}
\label{fig:lstm_exp}
\end{figure}

 \mypar{Integration with LSTM-ATT}. We have experimented with two methods to integrate the high-level semantic features - SVO classification, object detection (DET) and image classification scores (CLS) - with the LSTM-ATT. In the first method, both the temporal visual features $P$ and semantic features $s$ are provided as input to the LSTM-ATT. In the second one, we stack a LSTM, that processes only the semantic features $s$, on top of LSTM-ATT which receives the temporal visual feature $P$ as input. We refer to these methods as LSTM-ATT(SEM) and LSTM2-ATT(SEM), respectively, where (SEM) stands for different subsets of semantic features.  A schematic view of the two models we use is shown in figure \ref{fig:lstm_exp}.
 
\subsection{Experimental Results} 
 
 \mypar{Quantitative Results}. Results obtained with the proposed models are shown in table \ref{tab:accuracy}. We first check whether the attention mechanism provides an advantage over mean pooling the temporal visual feature, as quantified by the currently most used metrics, BLEU and METEOR. Using a simple LSTM that receives as input the mean pooled temporal visual feature, we obtain a score of 45.4\% on BLEU@4 and 31.2\% on METEOR. With LSTM-ATT, the results are considerably higher: 48.7\% on BLEU@4 and 31.9\% on METEOR, which demonstrates that the attention mechanism not only provides a way to focus selectively on the input video but also improves results. This is also true when adding semantic features both to the standard LSTM (with mean pooled temporal visual feature) and to the LSTM-ATT.
 
Our LSTM-ATT model achieves competitive results compared to other methods. Adding semantic features on top of this model improves the state-of-the-art results on the BLEU@n metric, while also performing well on METEOR. We show results with both LSTM-ATT(SEM) and LSTM2-ATT(SEM). The contributions of the SVO semantic features alone and in conjunction with DET and CLS features are also presented. In the case of SVO features alone, the best results are obtained with LSTM2-ATT(SVO) method for both evaluation metrics (BLEU@4 52.0\%, METEOR 32.3\%), while when using the full semantic features, our best performing method under BLEU is LSTM-ATT(SVO,DET,CLS) (50.6\%) and under METEOR is LSTM2-ATT(SVO,DET,CLS) (32.4\%). 
 
\begin{figure*}
\begin{centering}
\scalebox{1}{
\begin{tabular}{ m{0.18\linewidth} m{0.18\linewidth} m{0.18\linewidth} m{0.46\linewidth} m{0.01\linewidth}}

 \includegraphics[height=35pt]{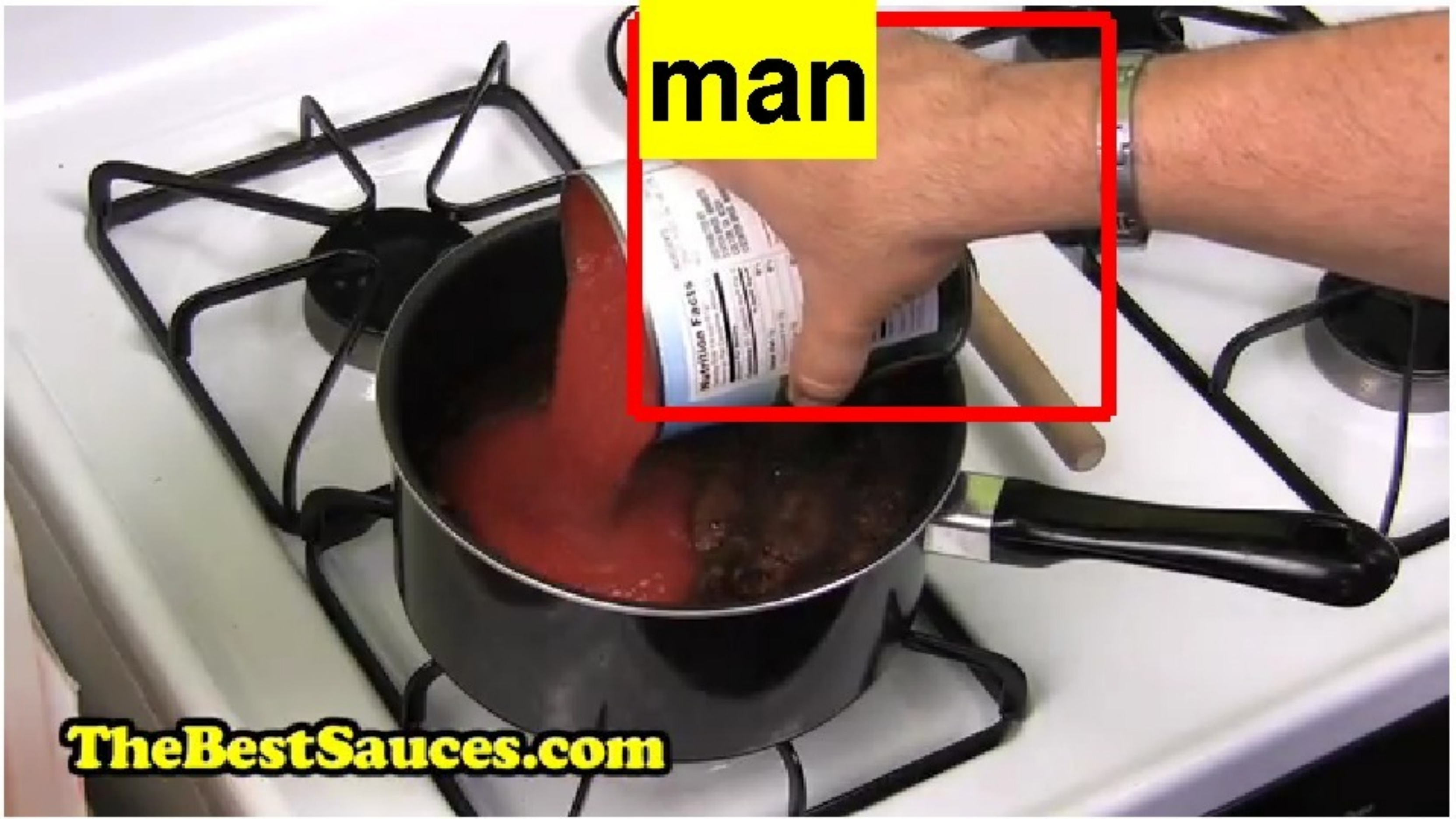}  &
 \includegraphics[height=35pt]{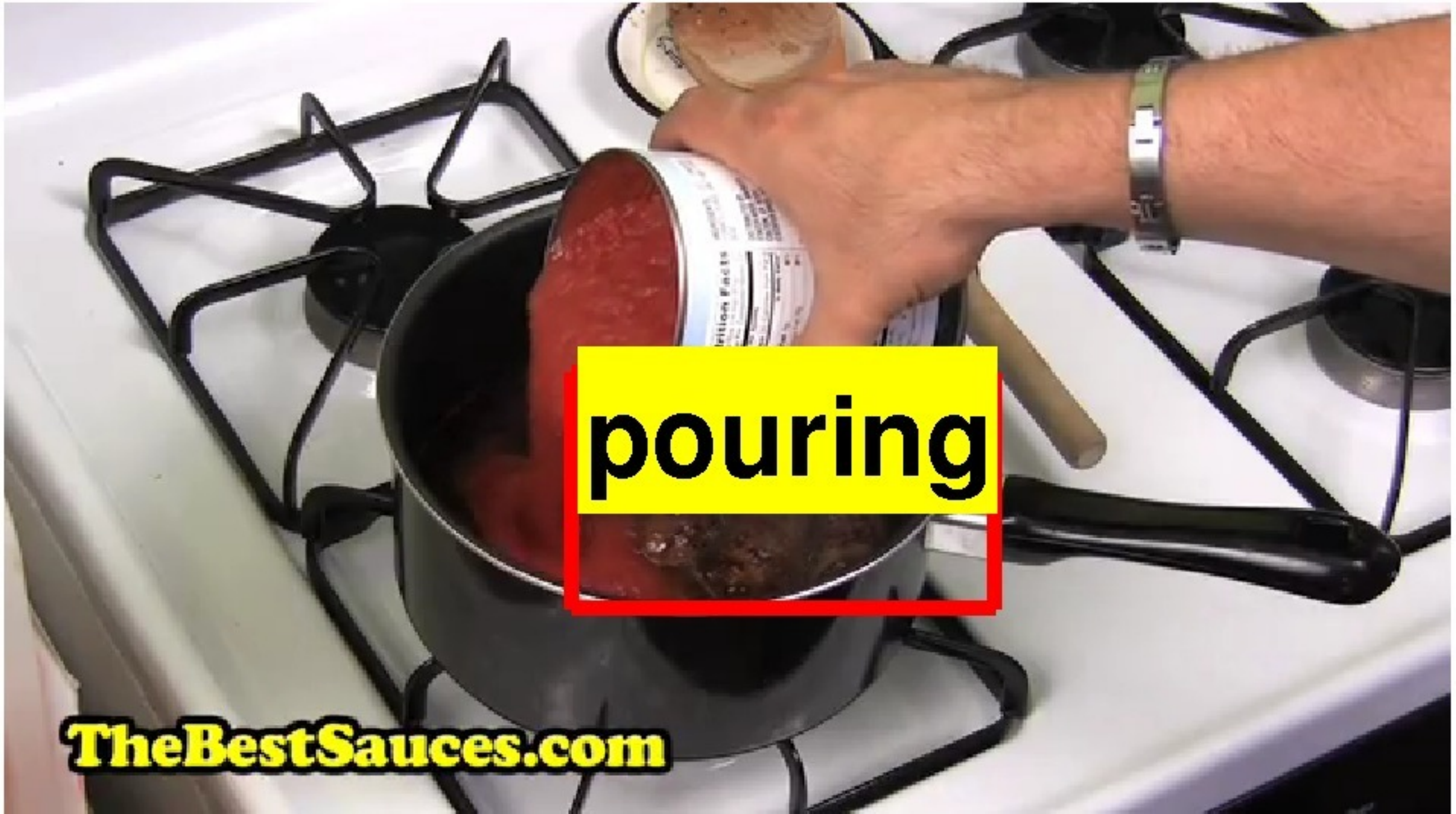} &
 \includegraphics[height=35pt]{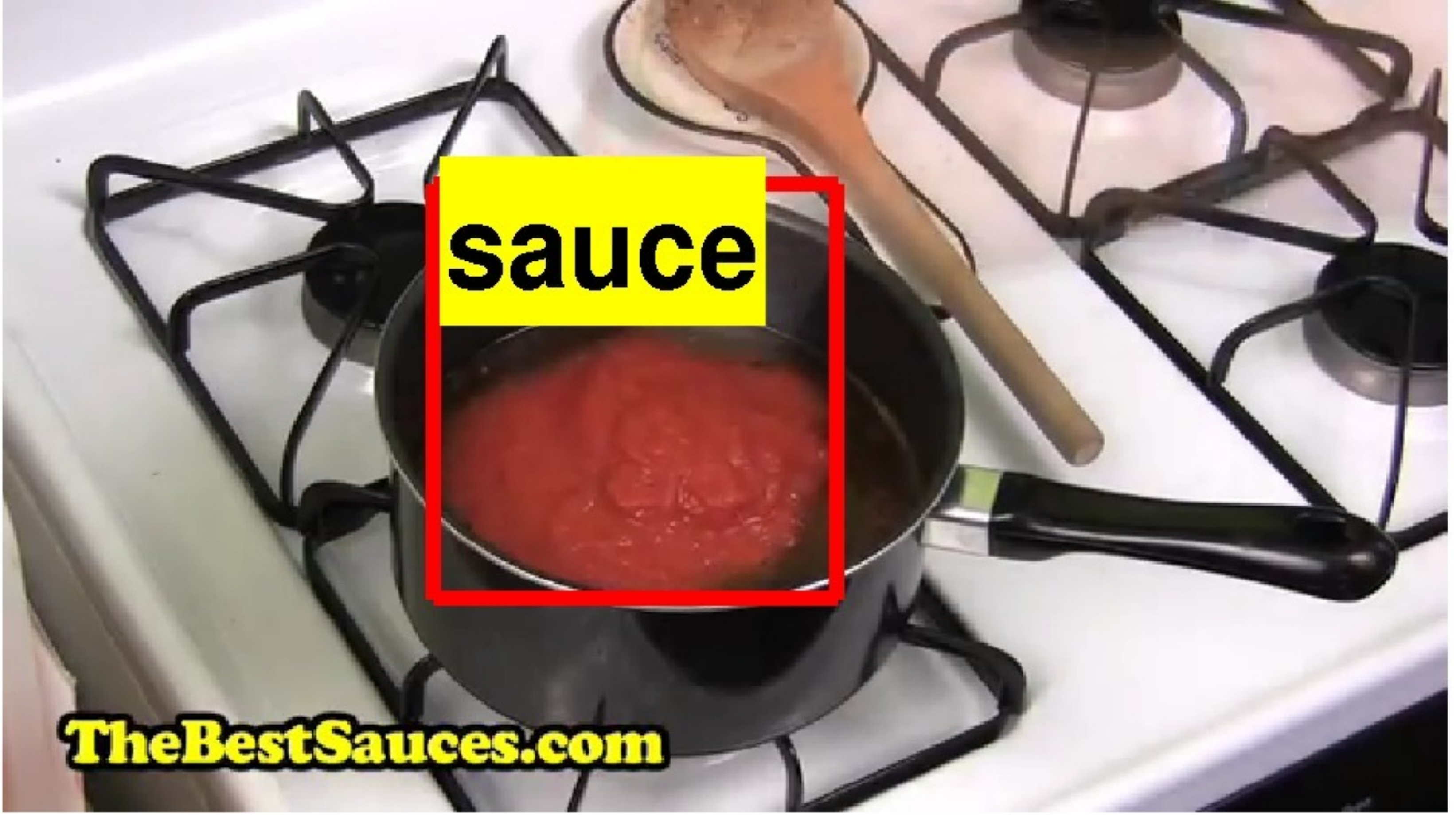} &
 \cfbox{green}{Ours: A \textbf{man} is \textbf{pouring} some \textbf{sauce}.}

 \hspace{0.02cm} Ref: A man is pouring some sauce into a pan. \\

 \includegraphics[height=46pt]{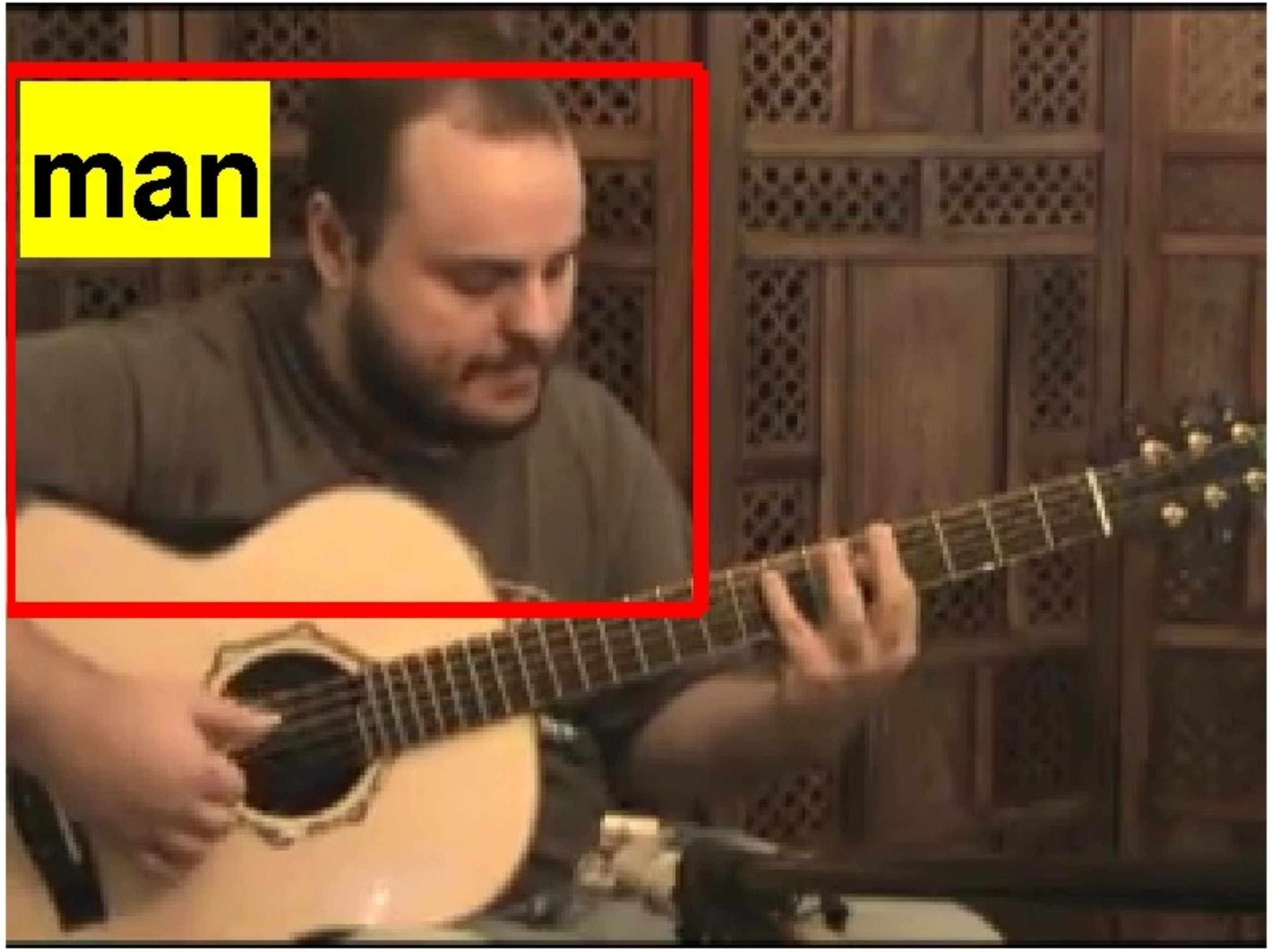}  &
 \includegraphics[height=46pt]{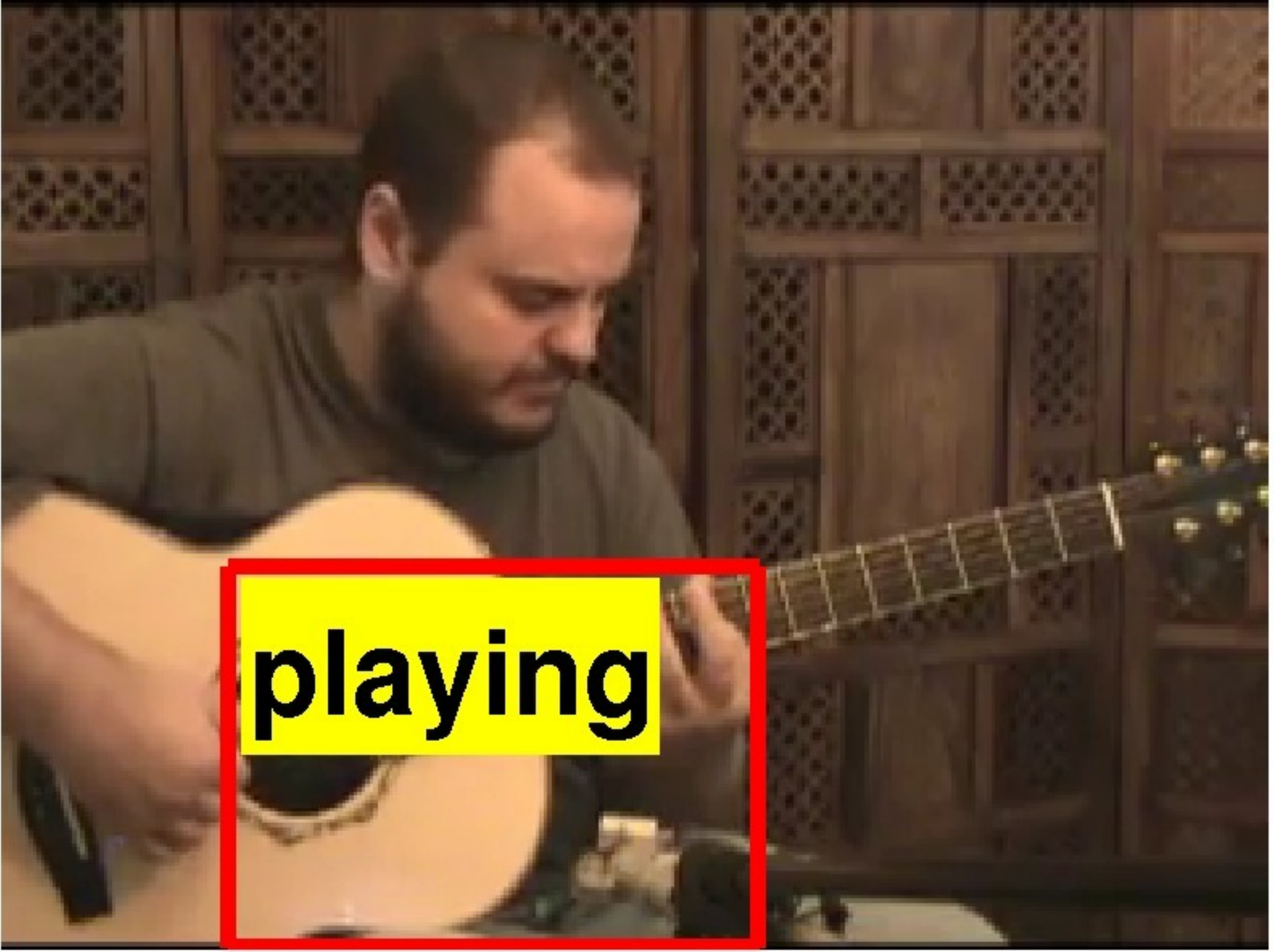} &
 \includegraphics[height=46pt]{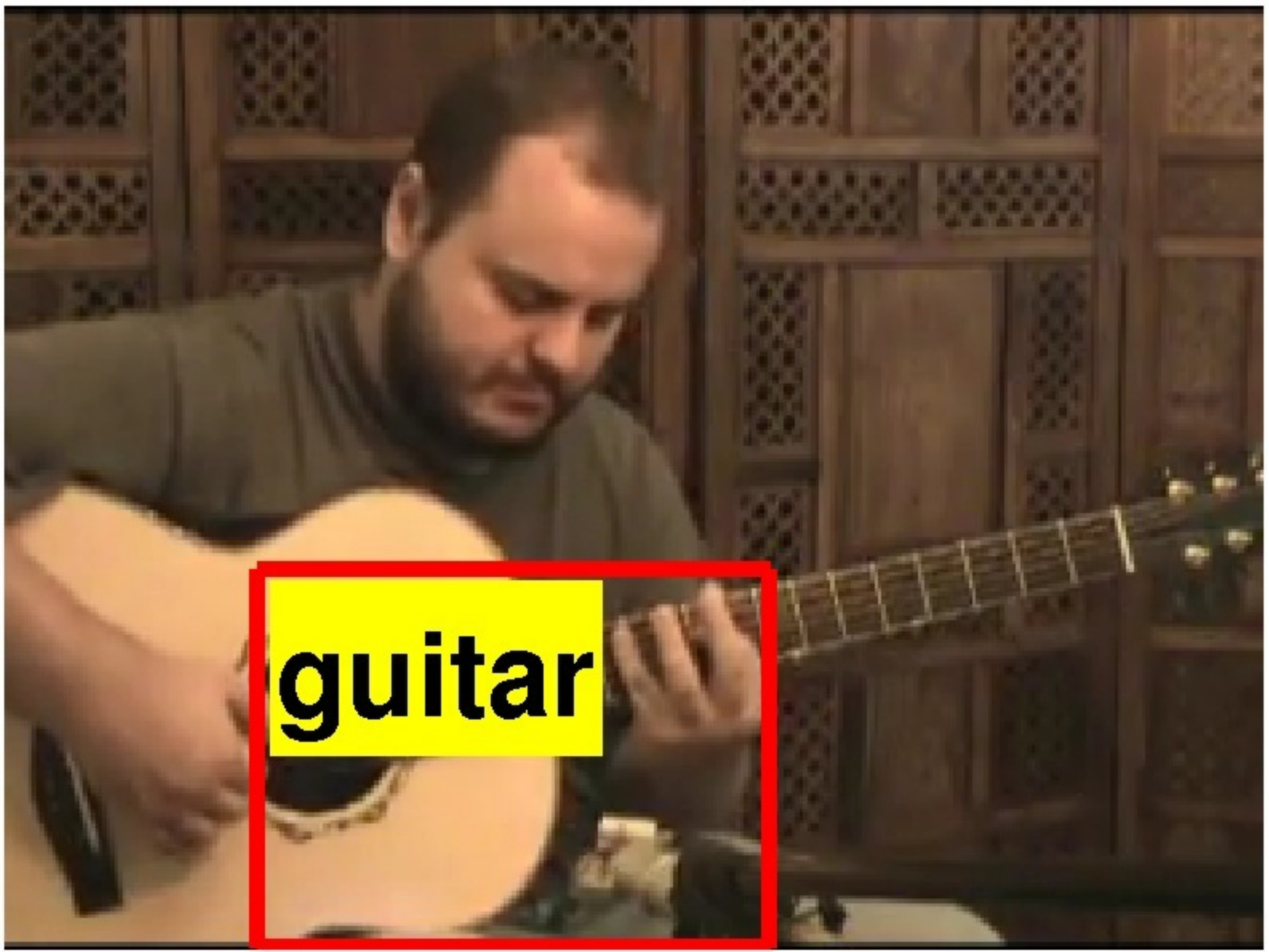} &
 \cfbox{green}{Ours: A \textbf{man} is \textbf{playing} \textbf{guitar}.}
 
 \hspace{0.02cm} Ref: A man is playing a guitar.\\

 \includegraphics[height=46pt]{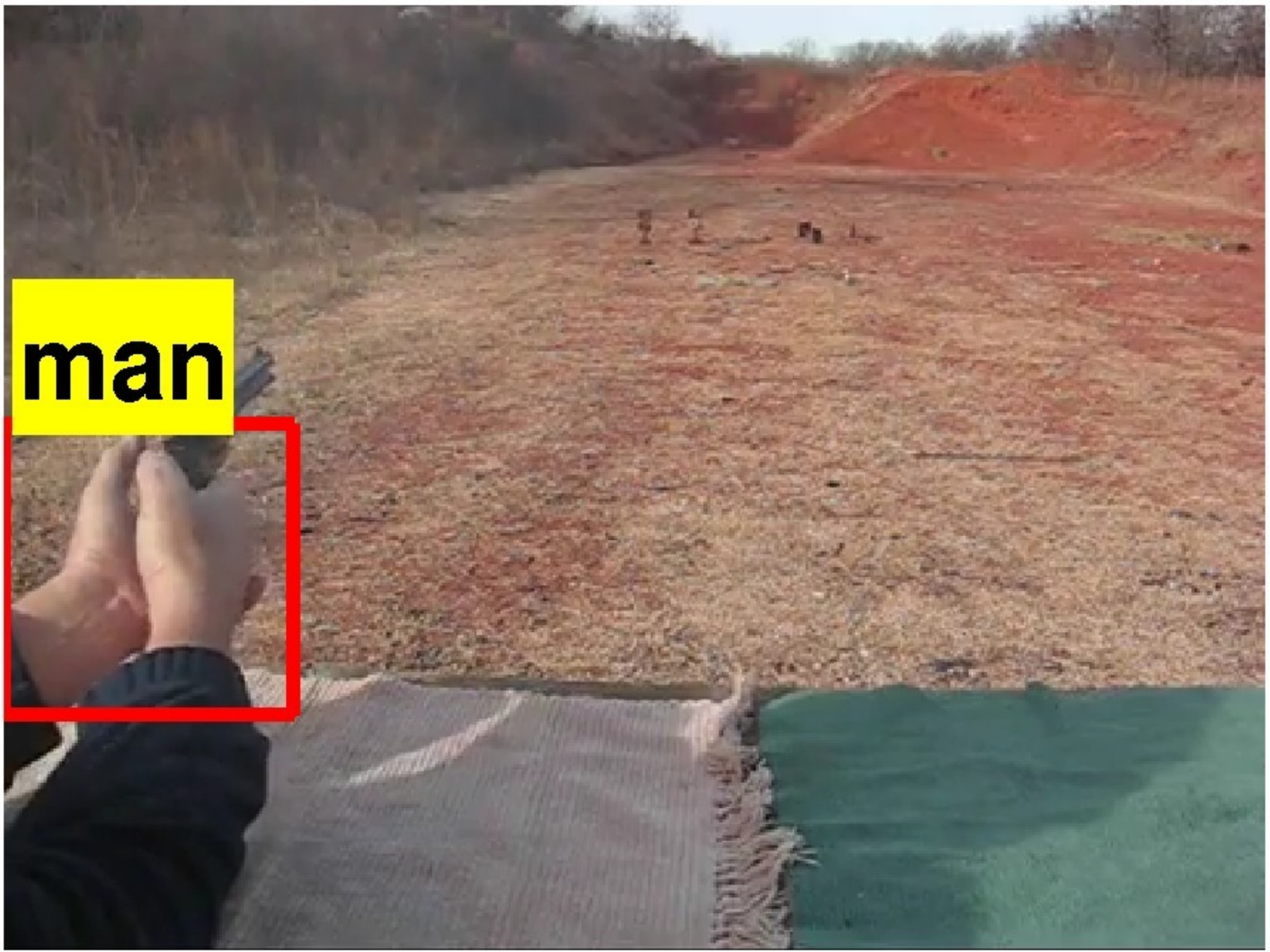}  &
 \includegraphics[height=46pt]{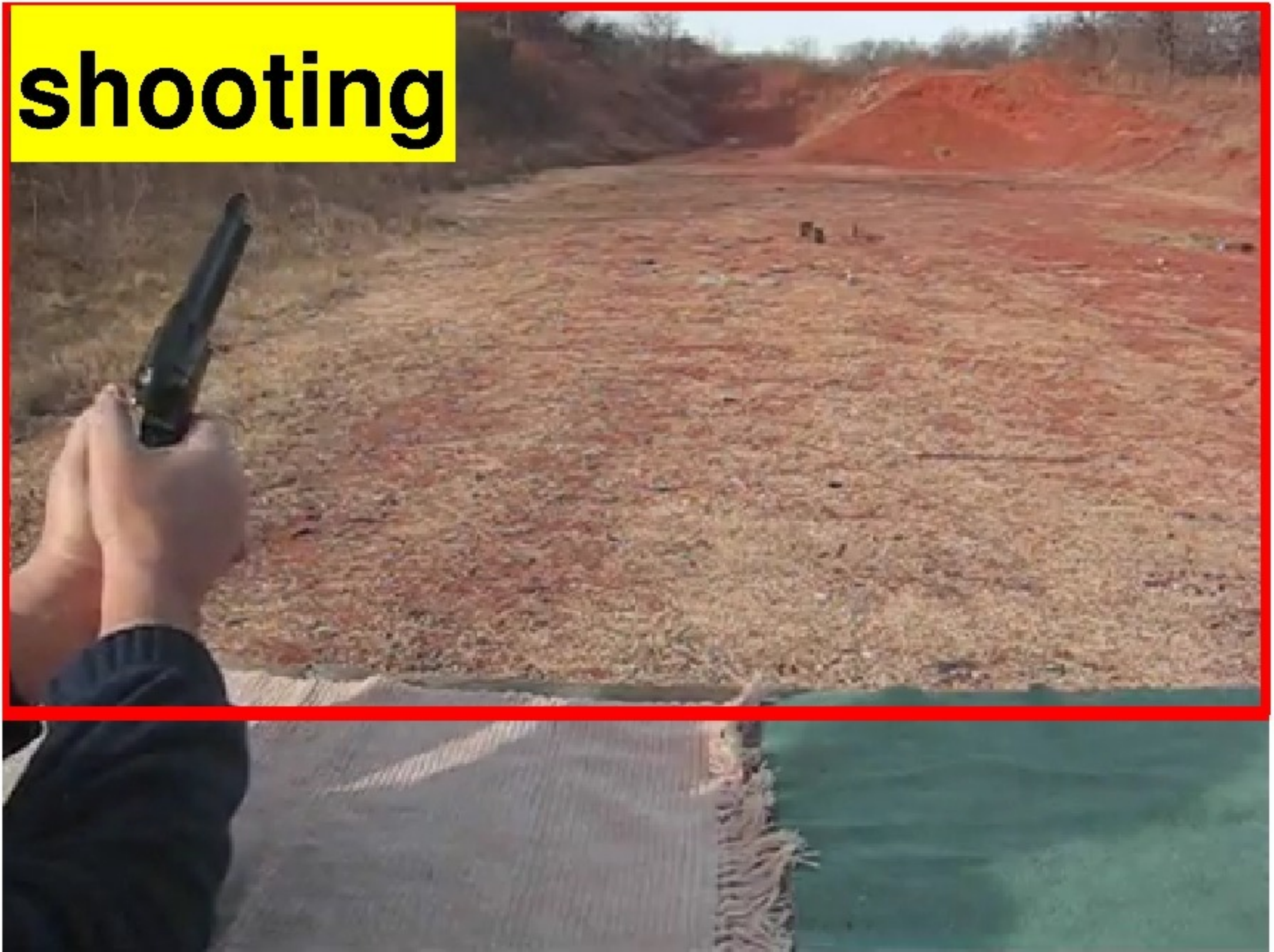} &
 \includegraphics[height=46pt]{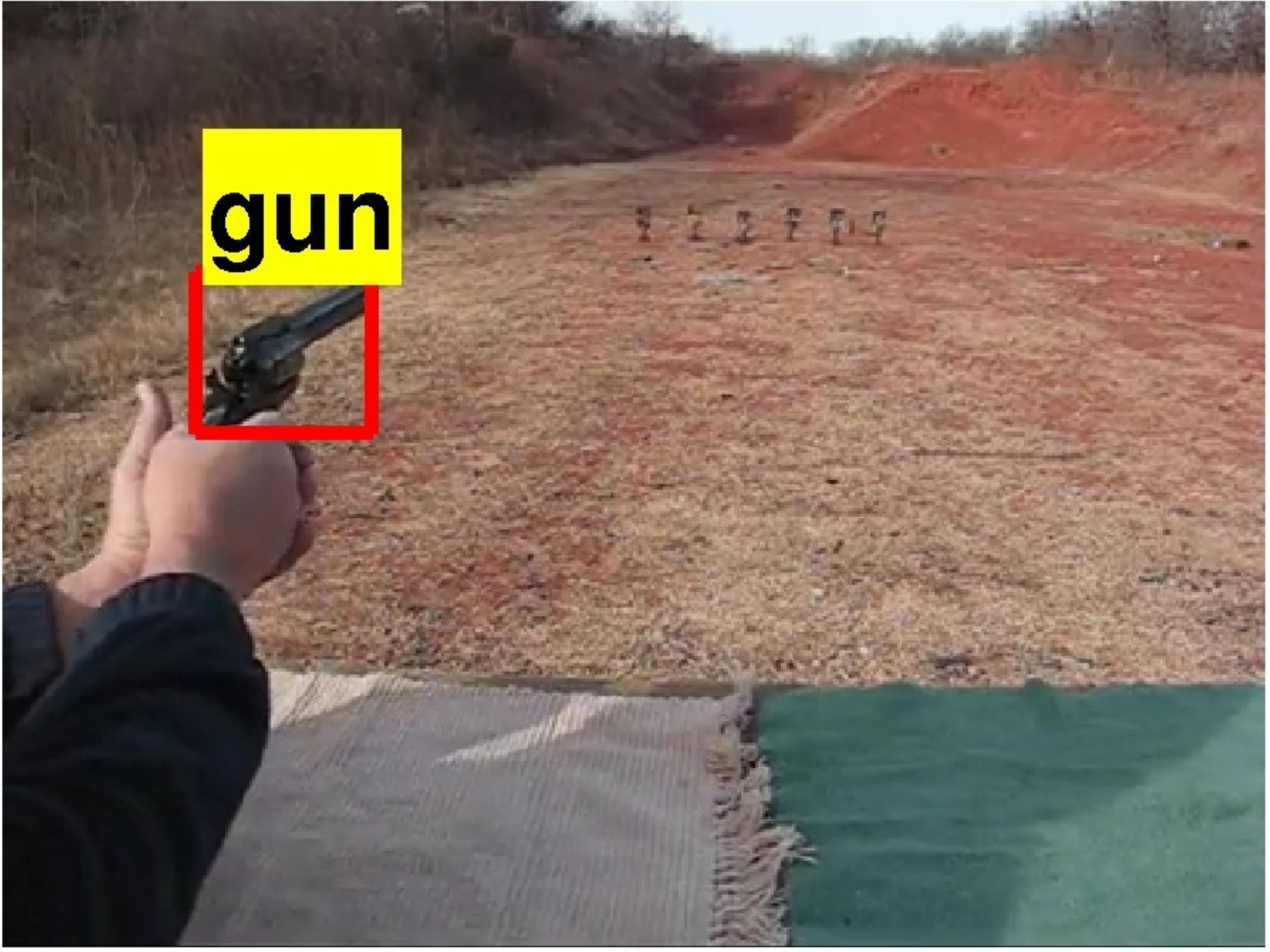} &
  \cfbox{green}{Ours: A \textbf{man} is \textbf{shooting} with a \textbf{gun}.}
  
  \hspace{0.02cm} Ref:  A guy is shooting a gun. \\

 \includegraphics[height=35pt]{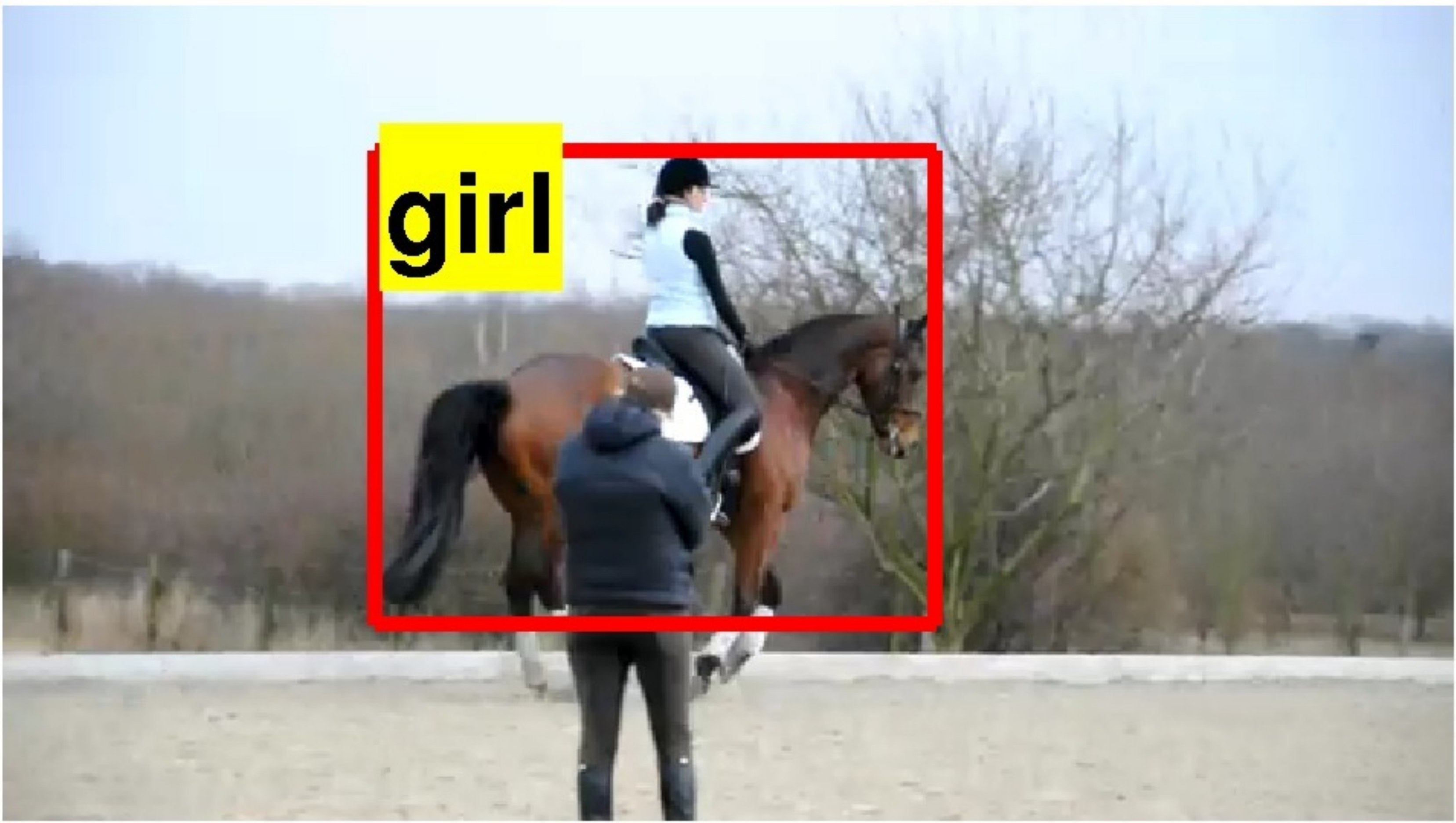}  &
 \includegraphics[height=35pt]{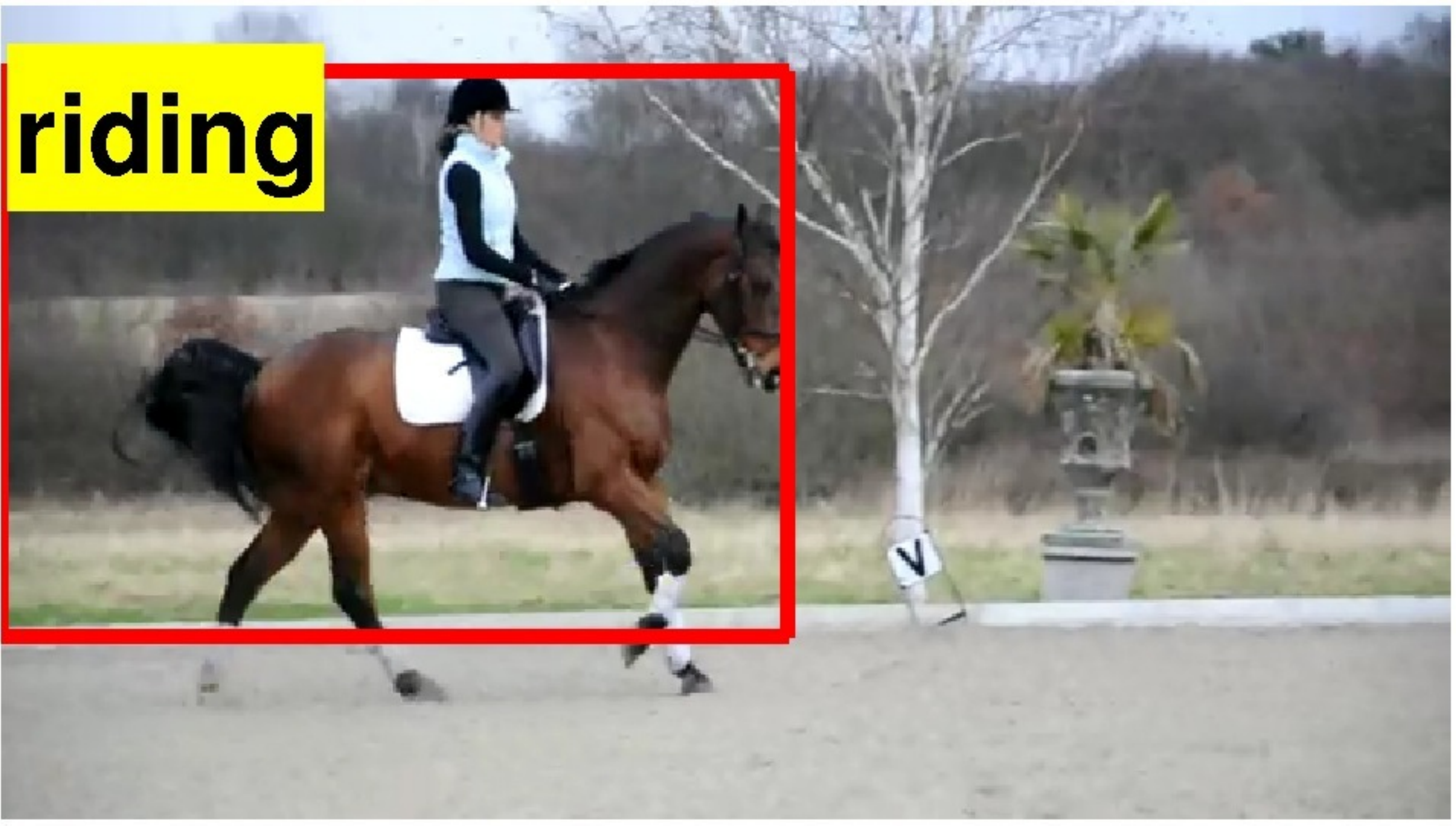} &
 \includegraphics[height=35pt]{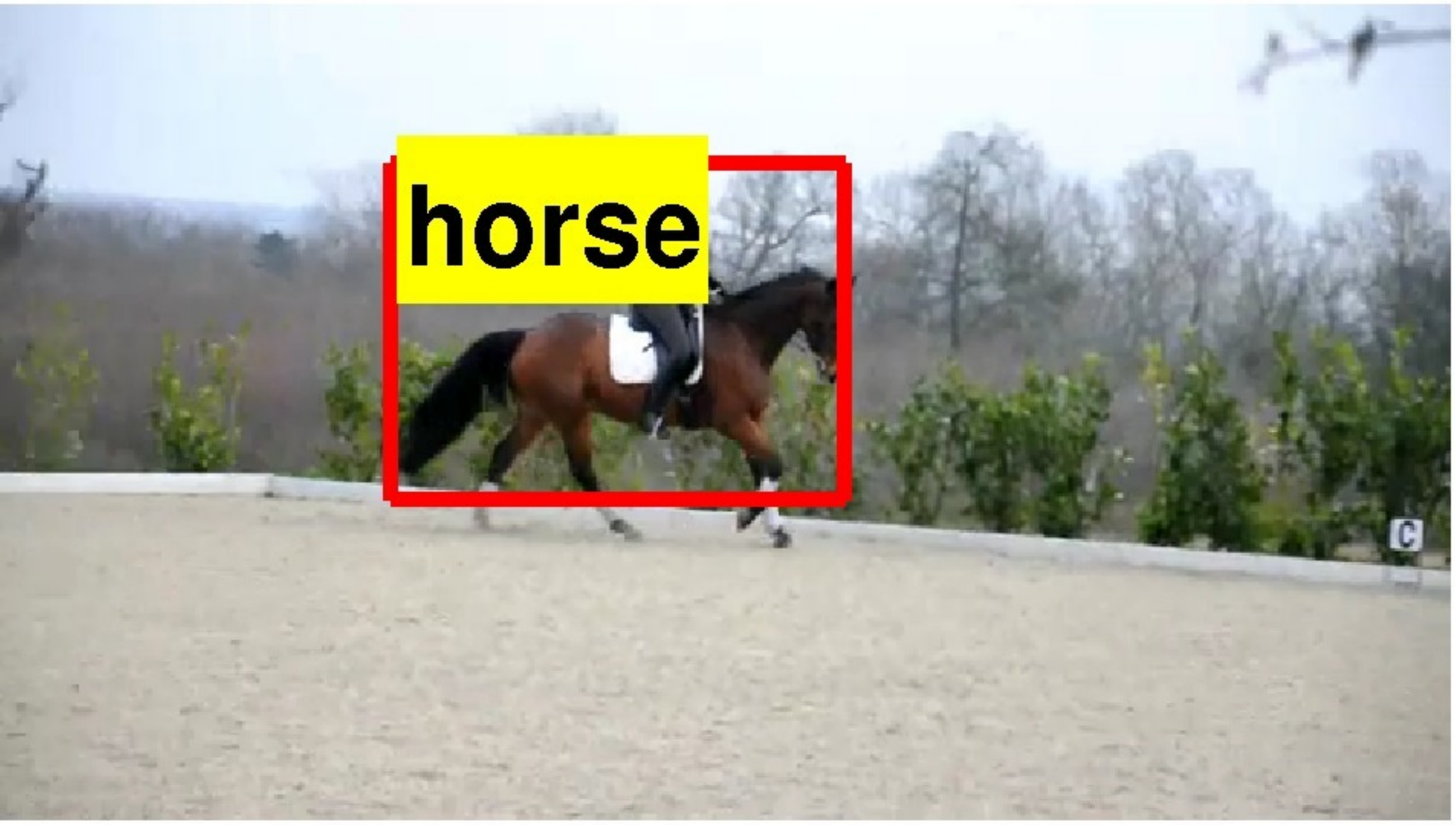} &
  \cfbox{green}{Ours: A \textbf{girl} is \textbf{riding}  a \textbf{horse}.}
  
  \hspace{0.02cm} Ref:  A woman is riding a horse. \\
  
  
  
  
  
    \includegraphics[height=35pt]{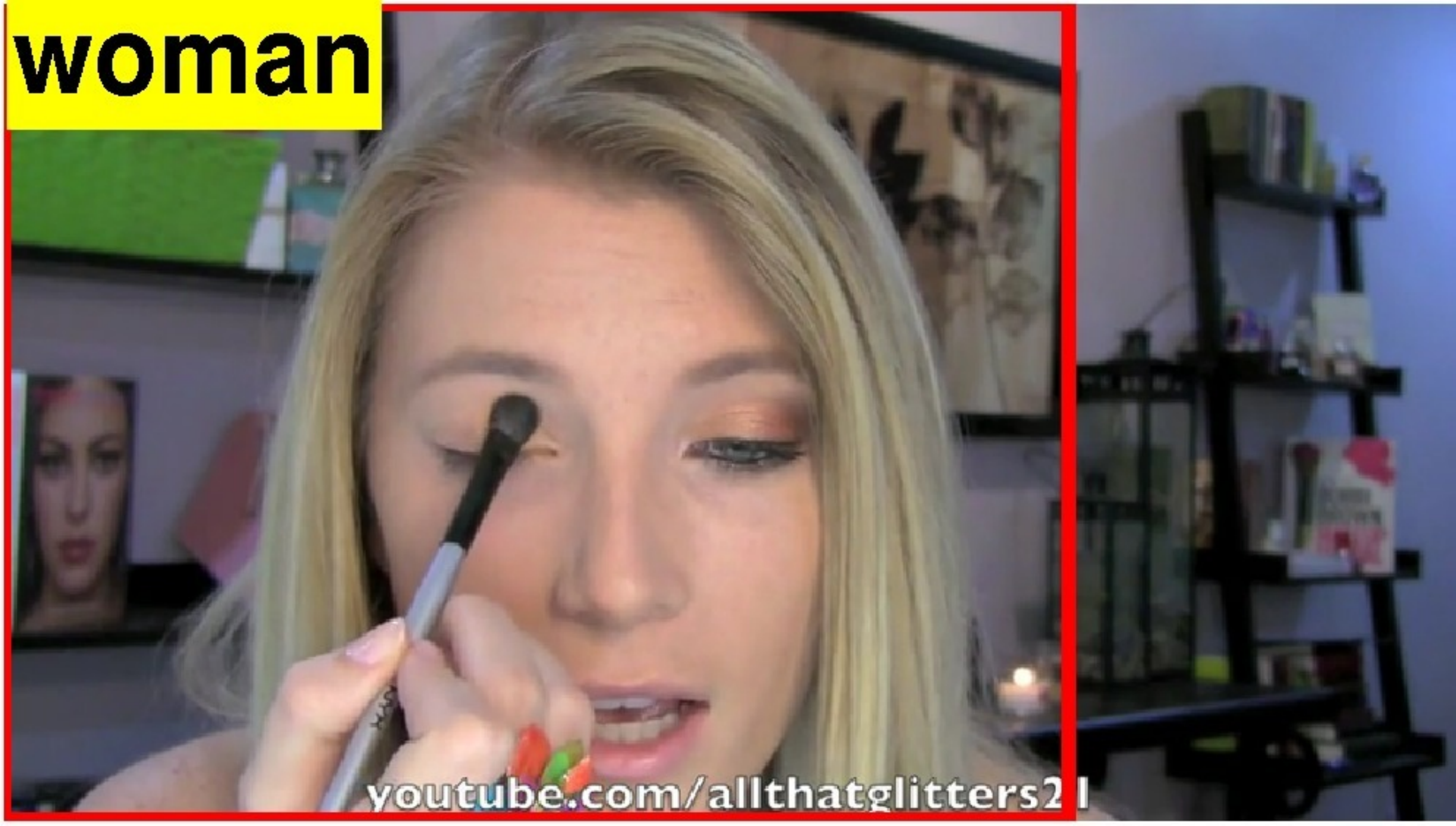}  &
 \includegraphics[height=35pt]{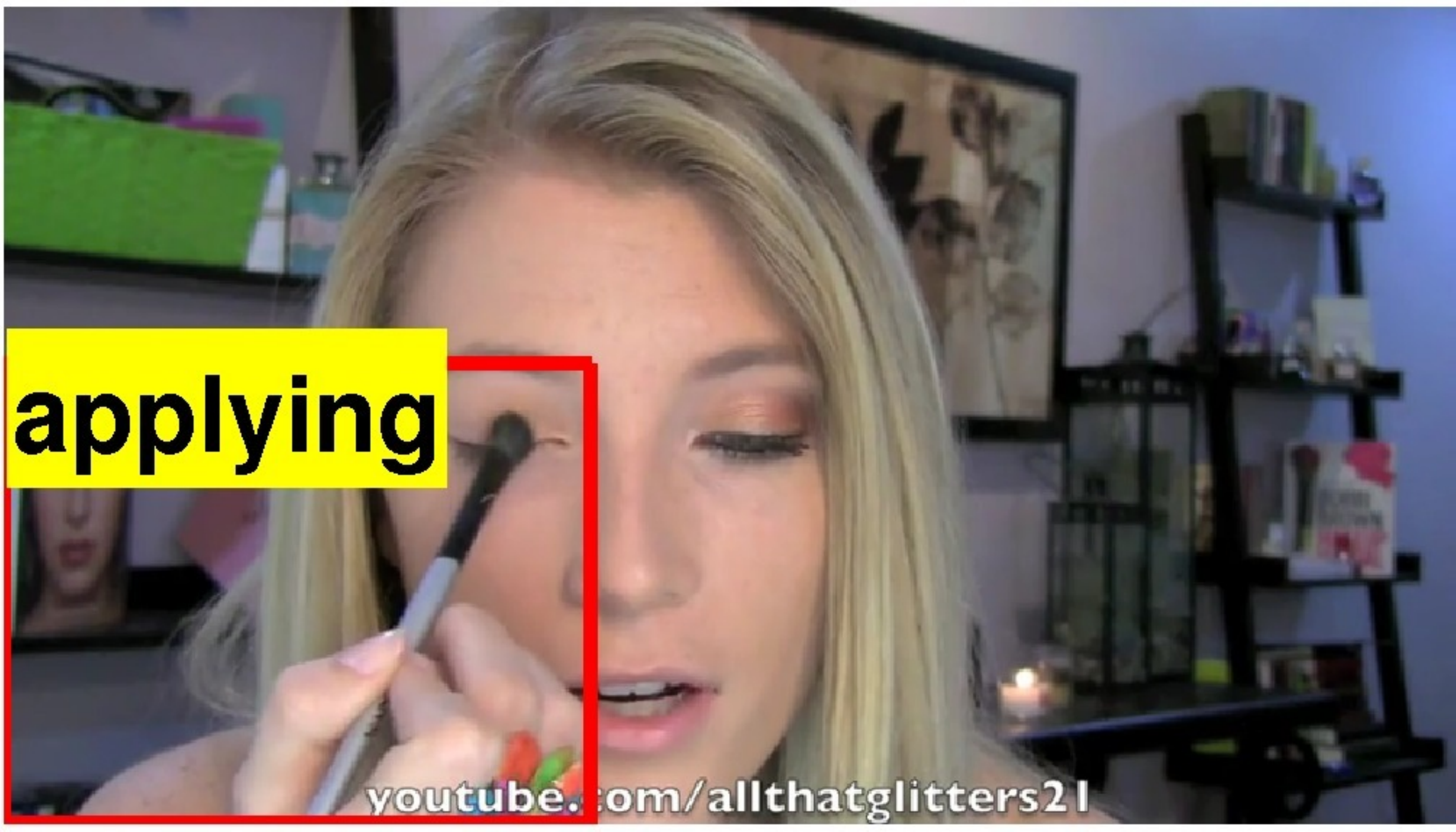} &
 \includegraphics[height=35pt]{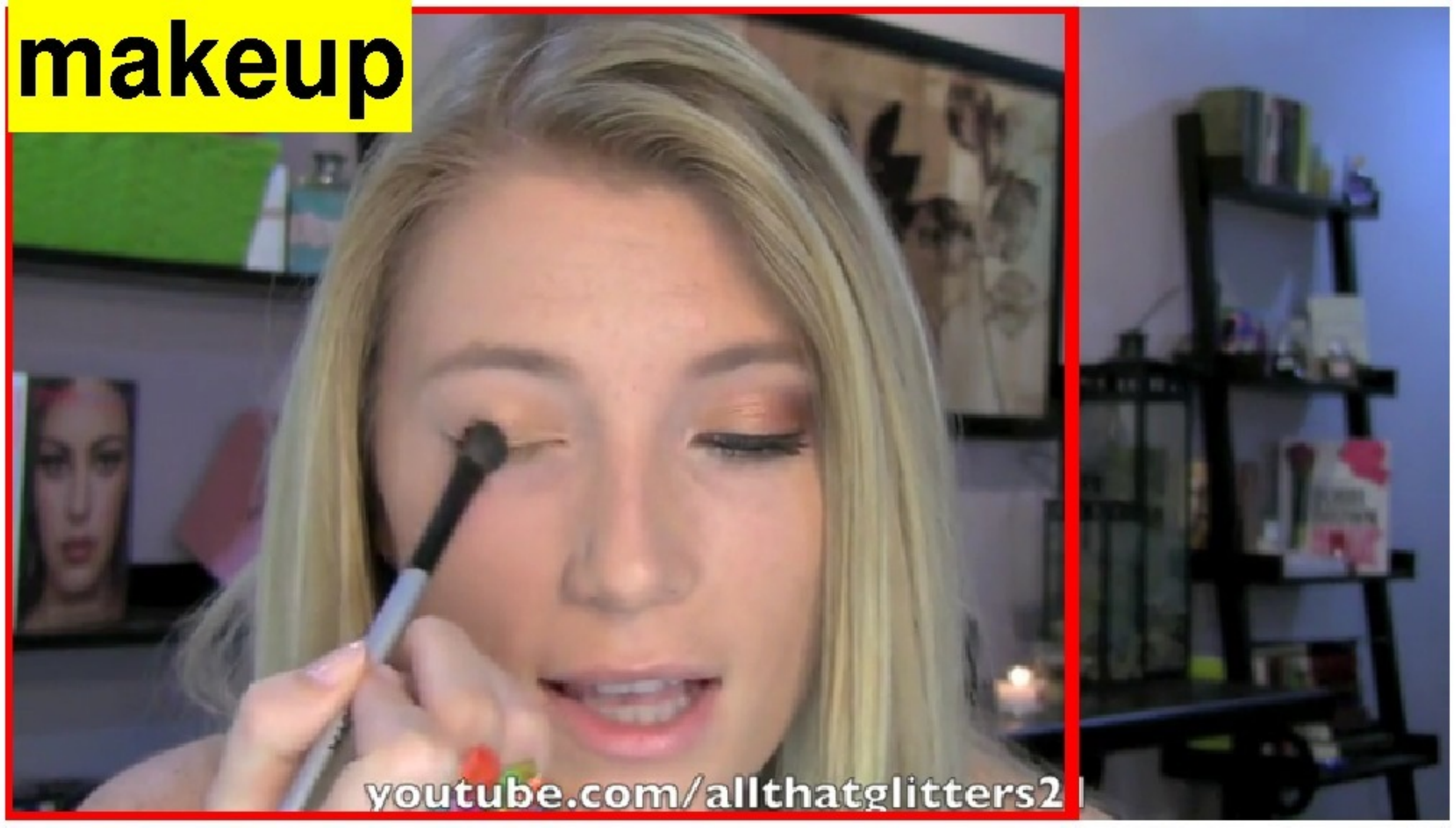} &
  \cfbox{green}{Ours: A \textbf{woman}  is \textbf{applying}  a \textbf{makeup}.}
  
  \hspace{0.02cm} Ref: A woman applies eye makeup. \\

    \includegraphics[height=47pt]{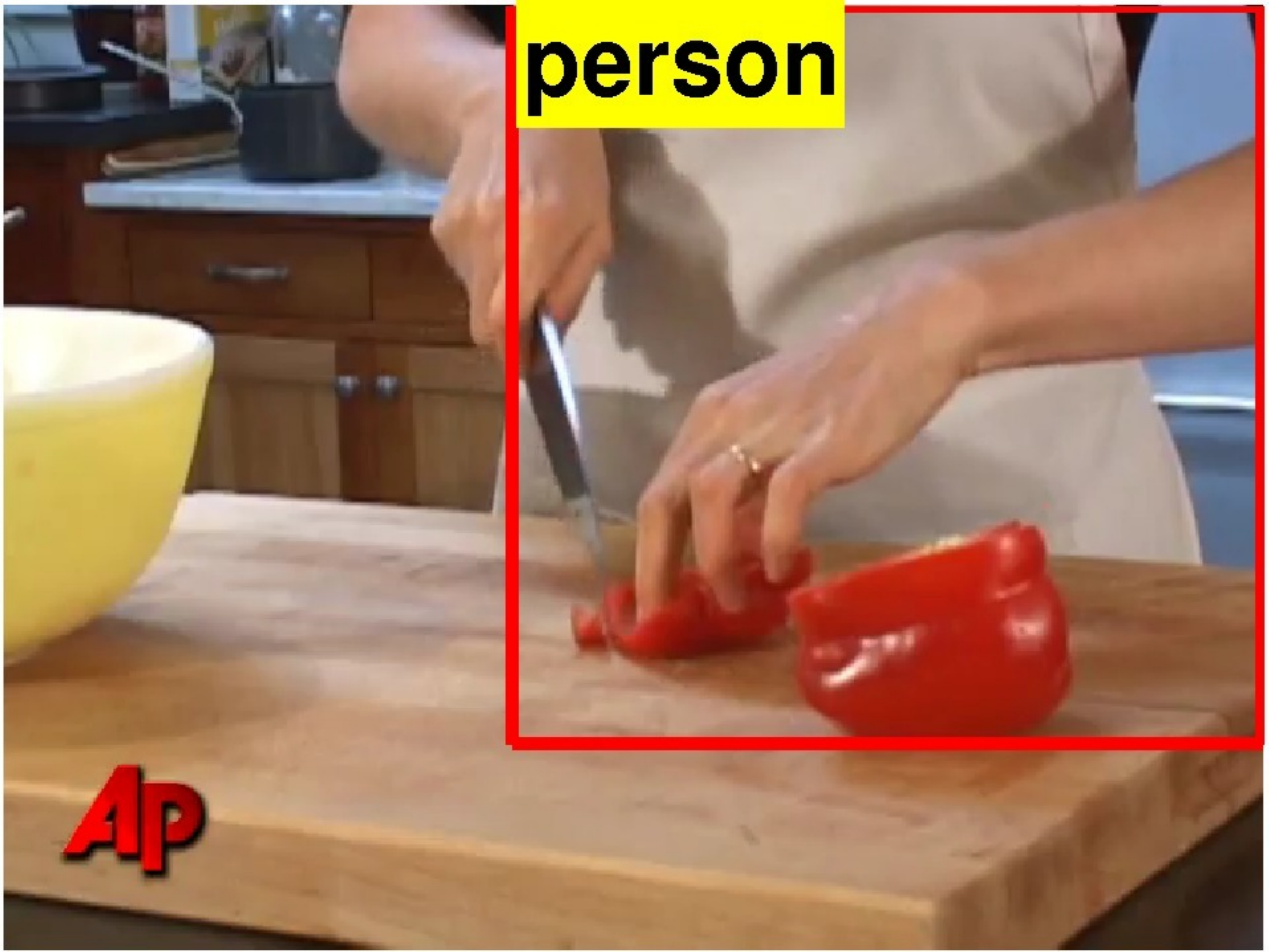}  &
 \includegraphics[height=47pt]{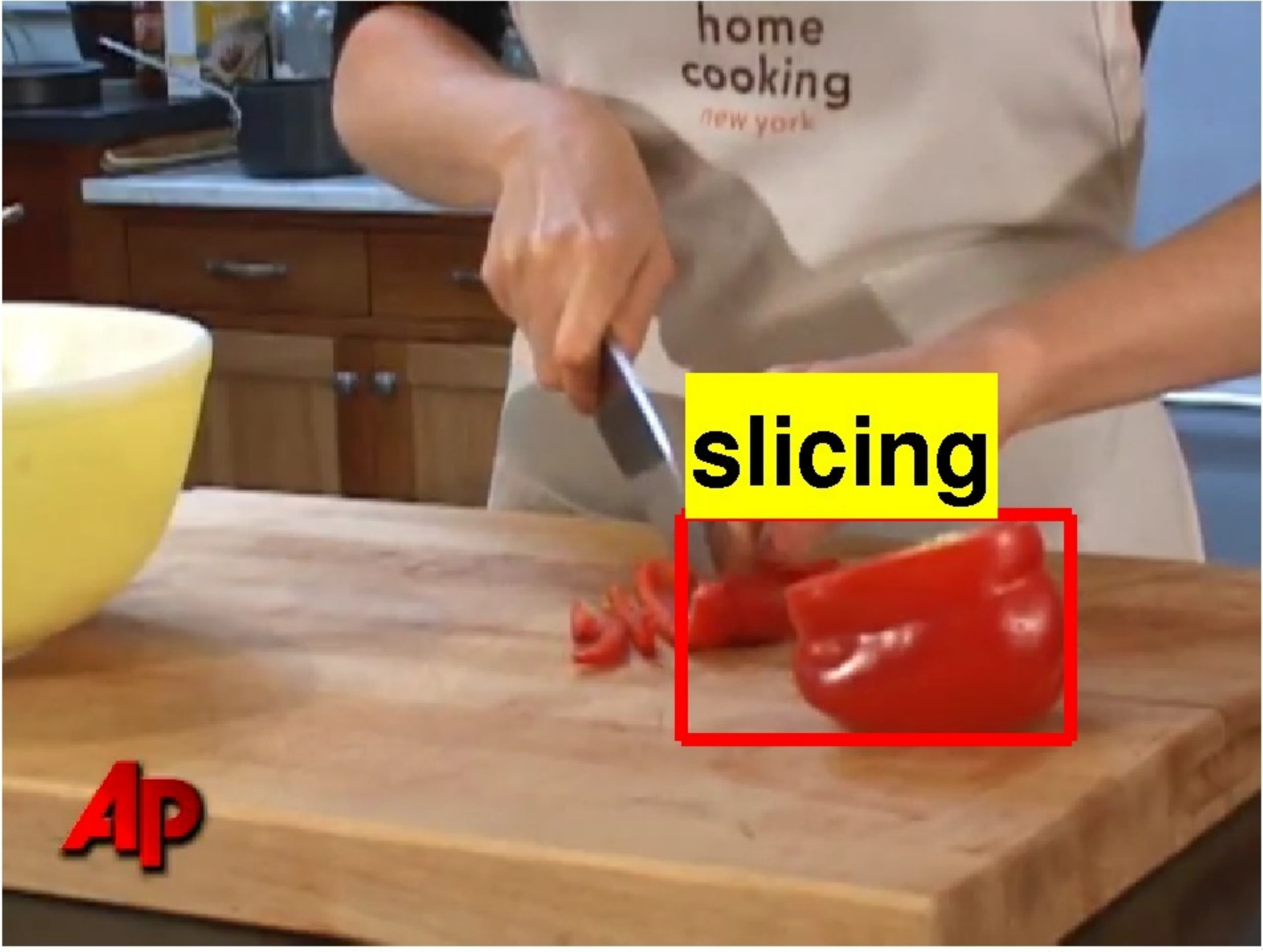} &
 \includegraphics[height=47pt]{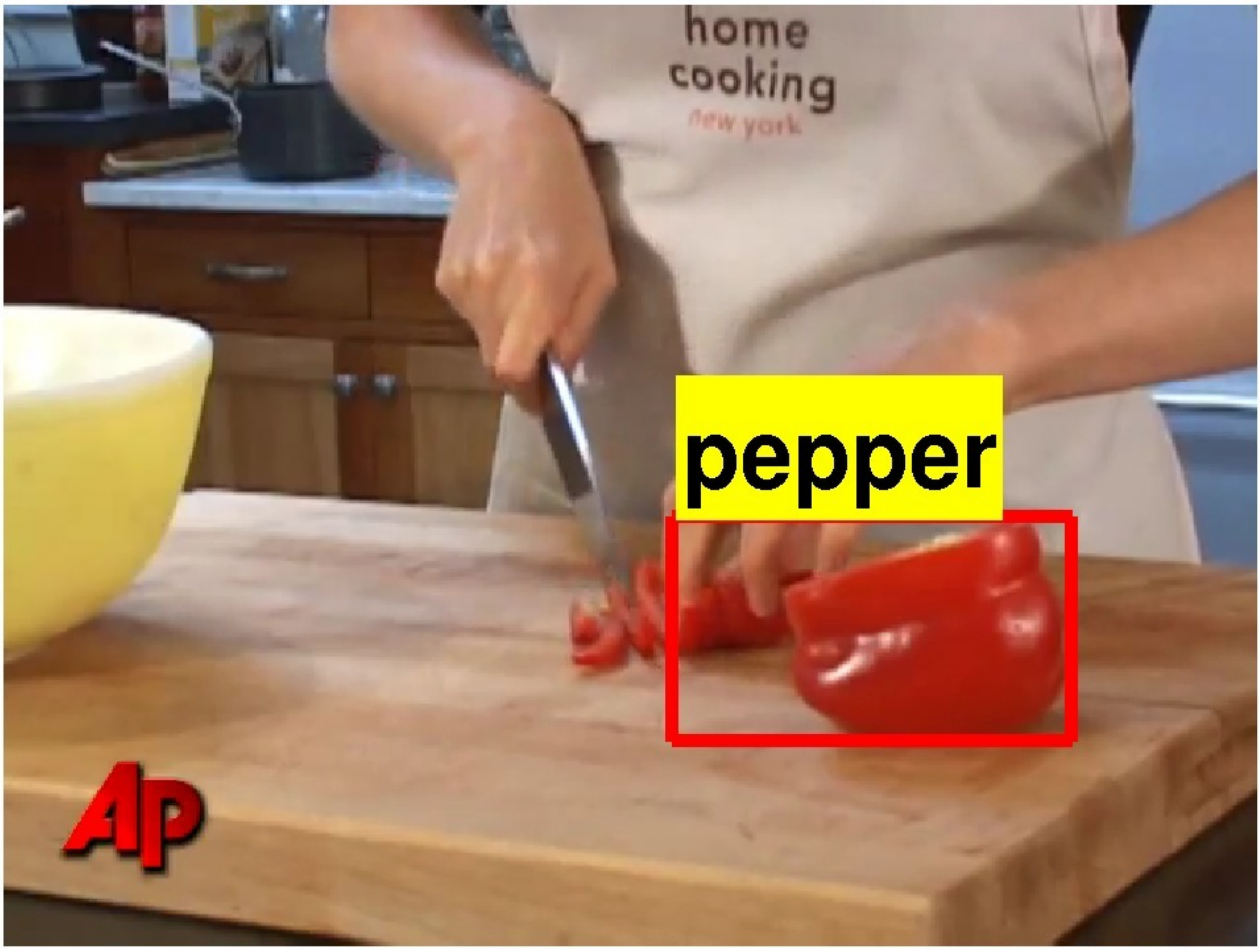} &
  \cfbox{green}{Ours: A \textbf{person}  is \textbf{slicing}  a \textbf{pepper}.}
  
  \hspace{0.02cm} Ref: A woman is slicing a pepper. \\
  
      \includegraphics[height=35pt]{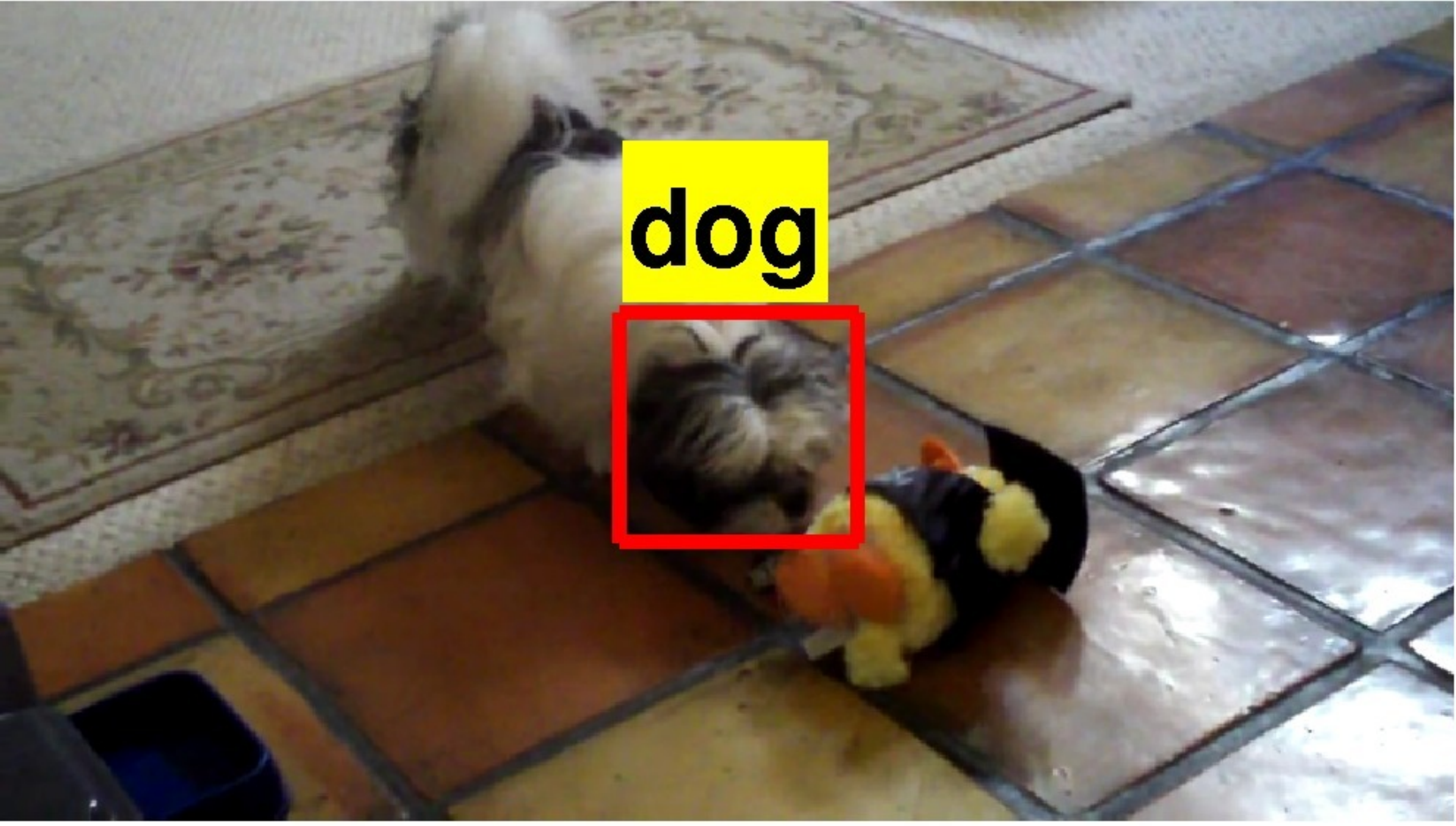}  &
 \includegraphics[height=35pt]{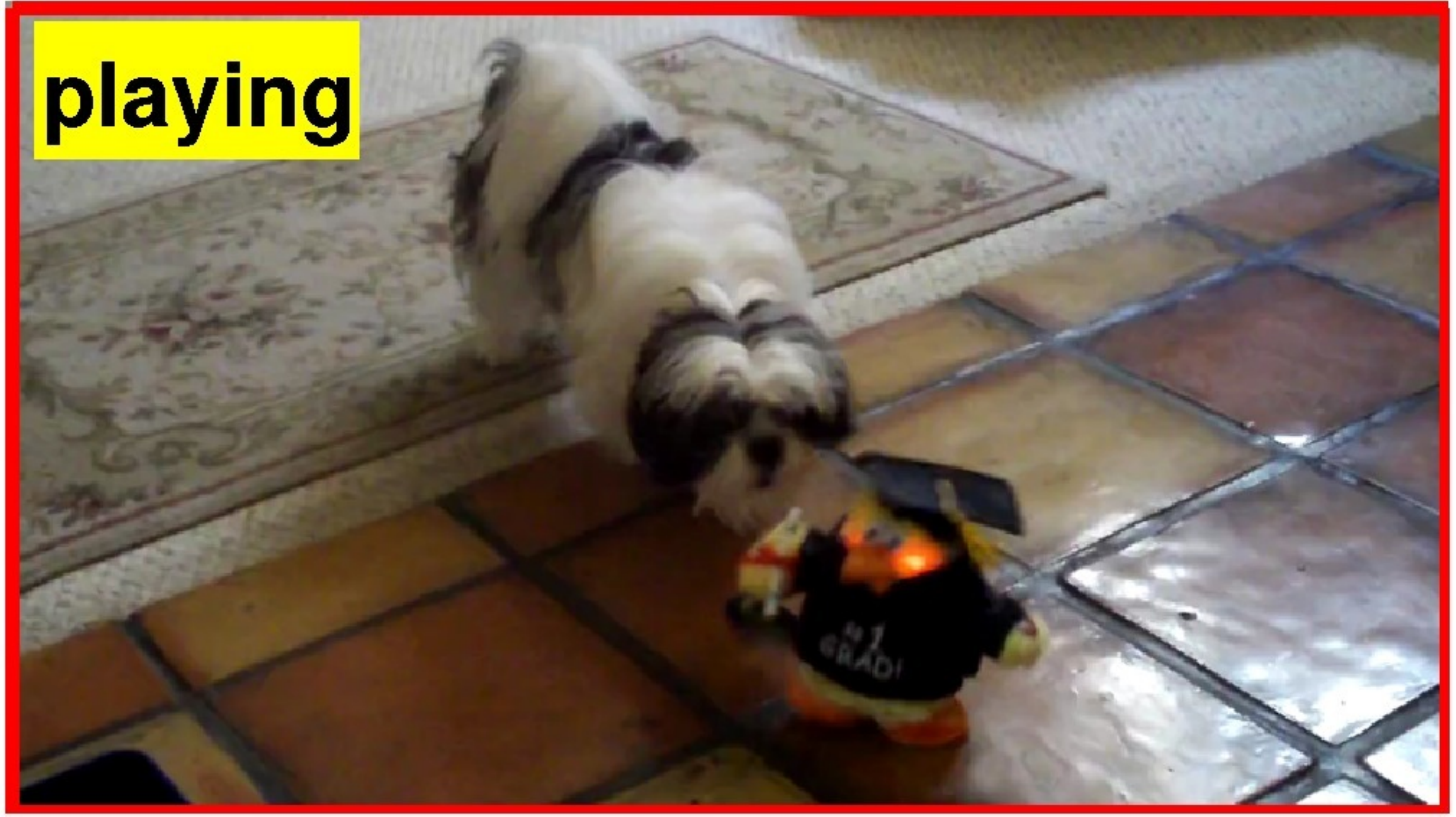} &
 \includegraphics[height=35pt]{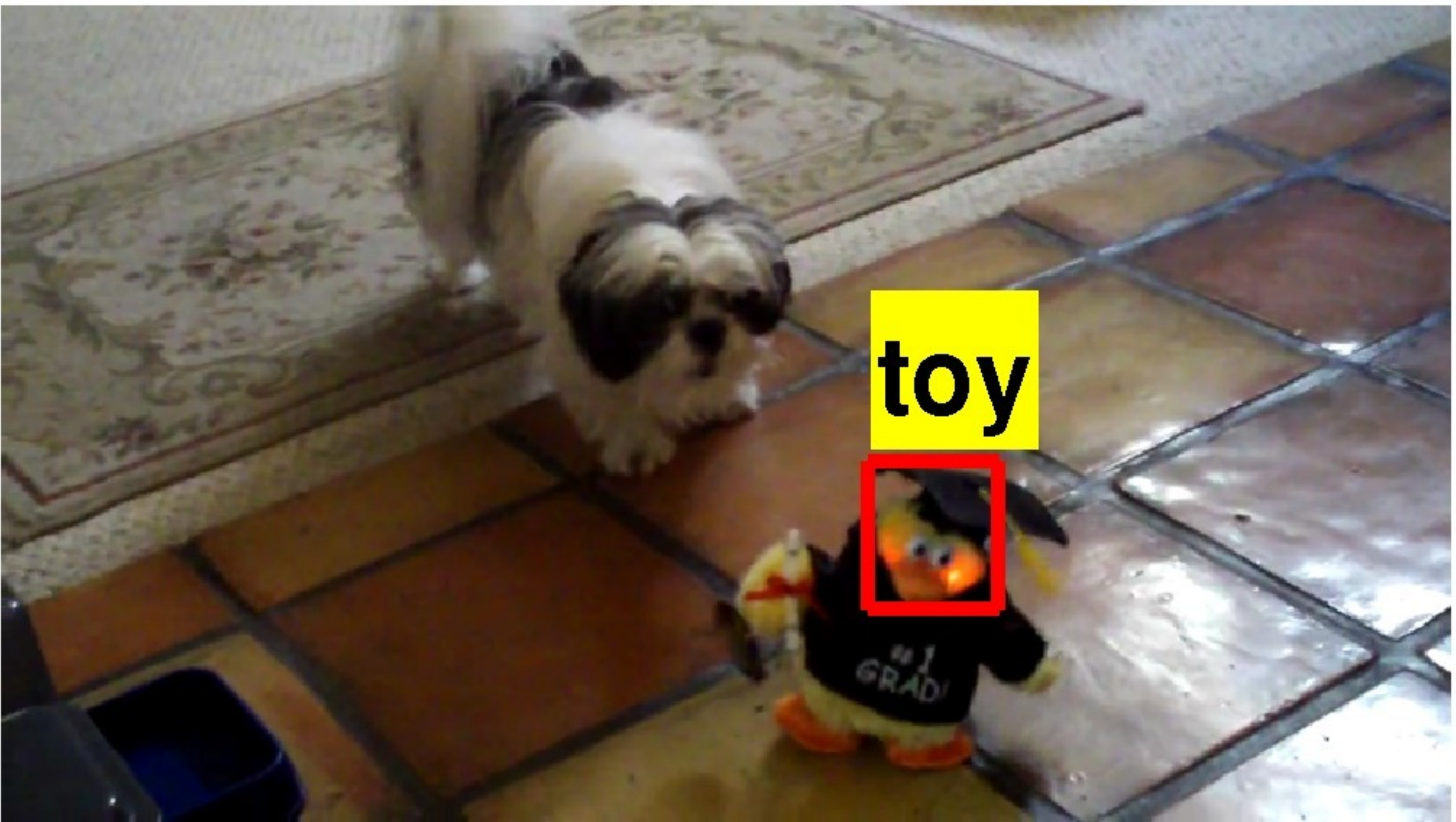} &
  \cfbox{green}{Ours: A \textbf{dog}  is \textbf{playing} with  a \textbf{toy}.}
  
  \hspace{0.02cm} Ref: A dog is playing with a toy. \\

  

      \includegraphics[height=42pt]{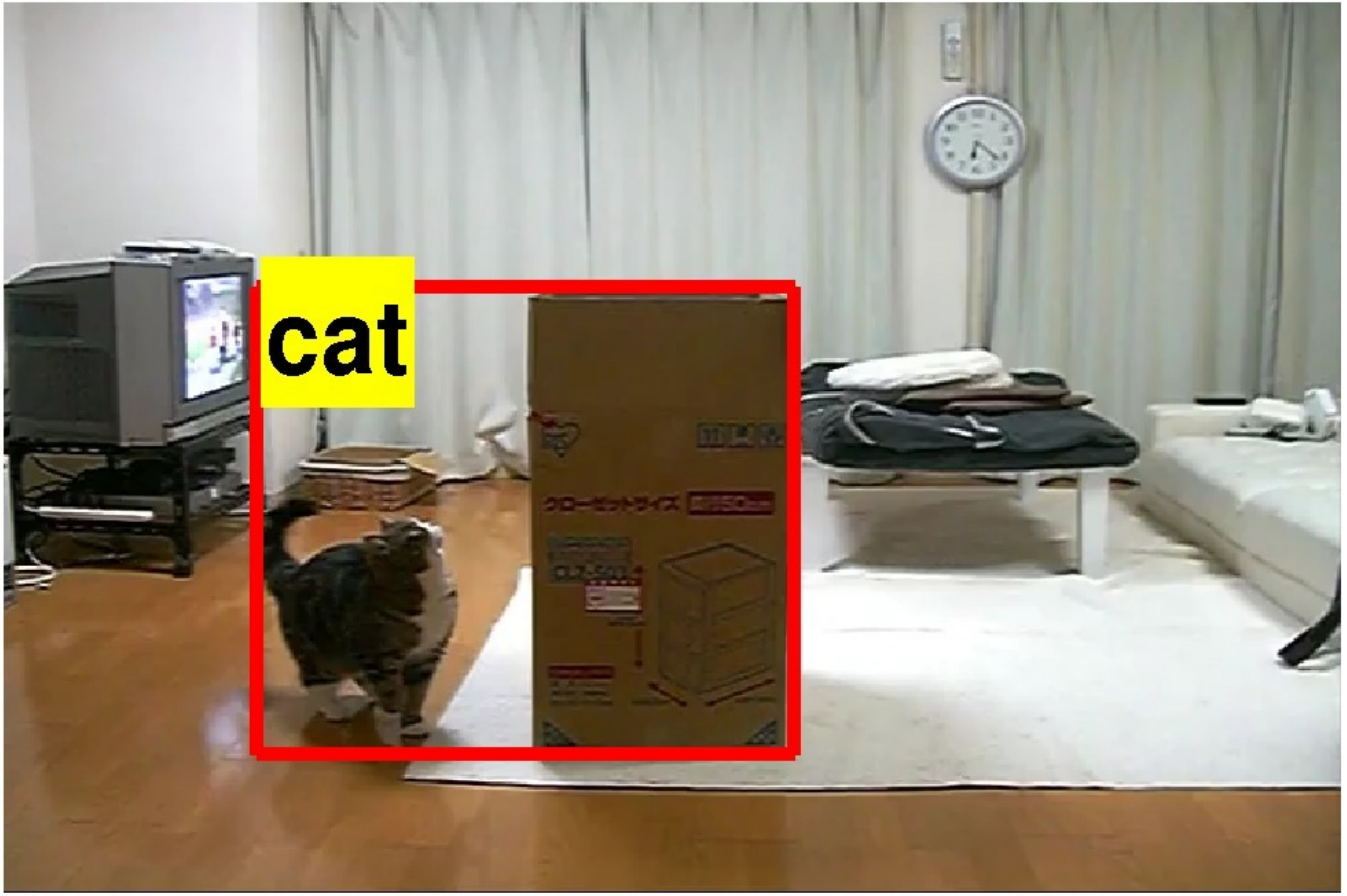}  &
 \includegraphics[height=42pt]{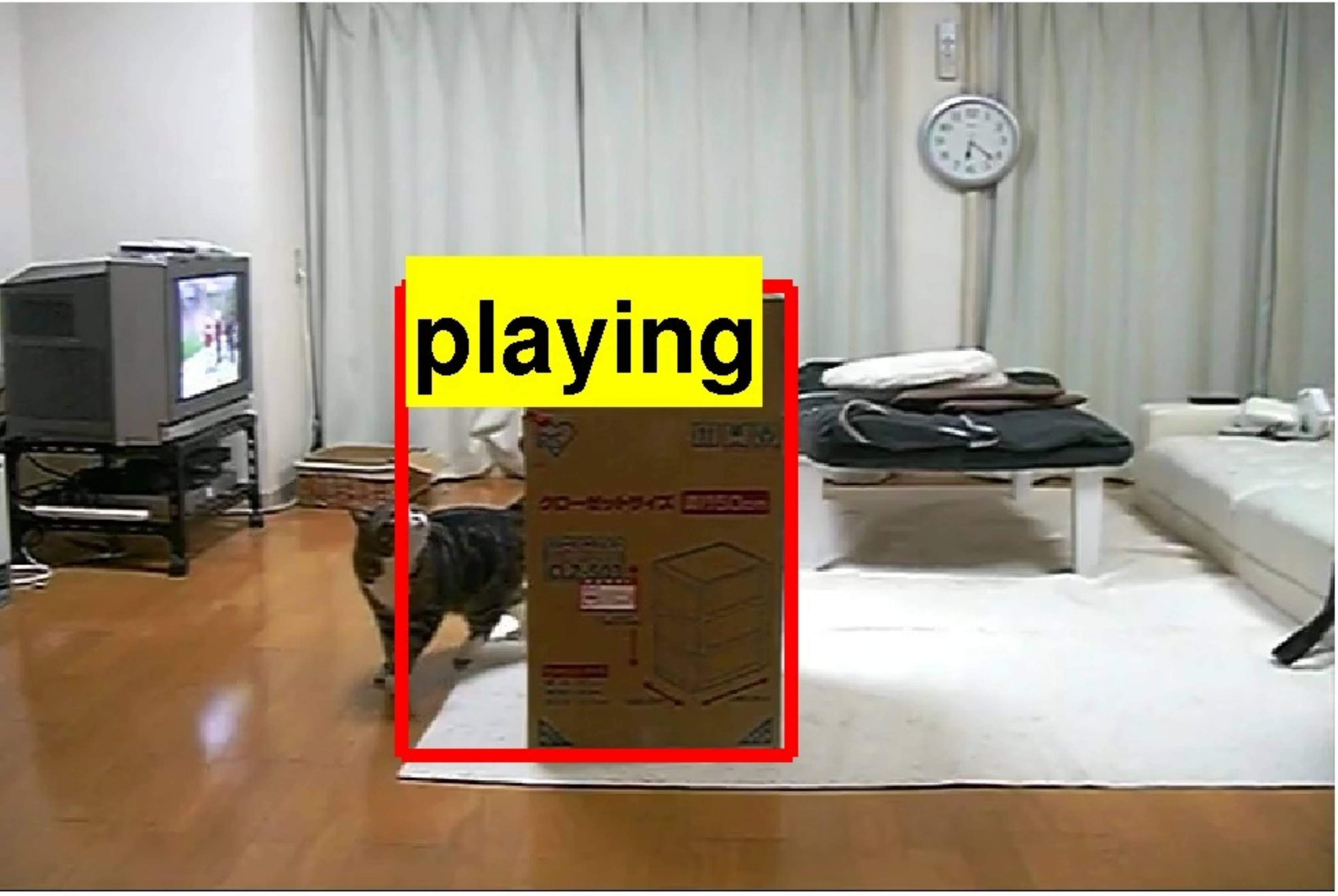} &
 \includegraphics[height=42pt]{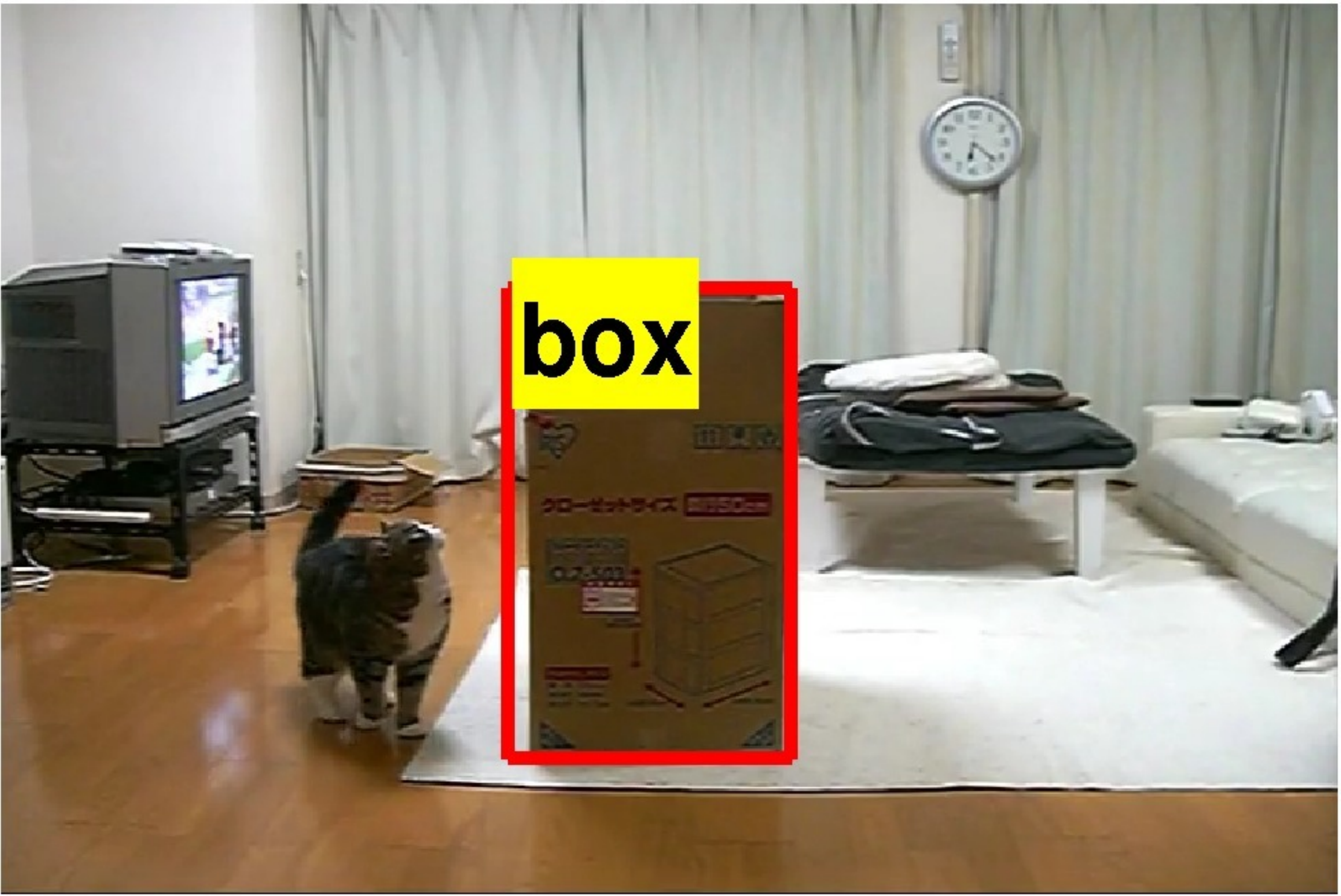} &
  \cfbox{green}{Ours: A \textbf{cat}  is \textbf{playing} with a \textbf{box}.}
  
  \hspace{0.02cm} Ref: A cat is jumping into a box. \\
  
        \includegraphics[height=47pt]{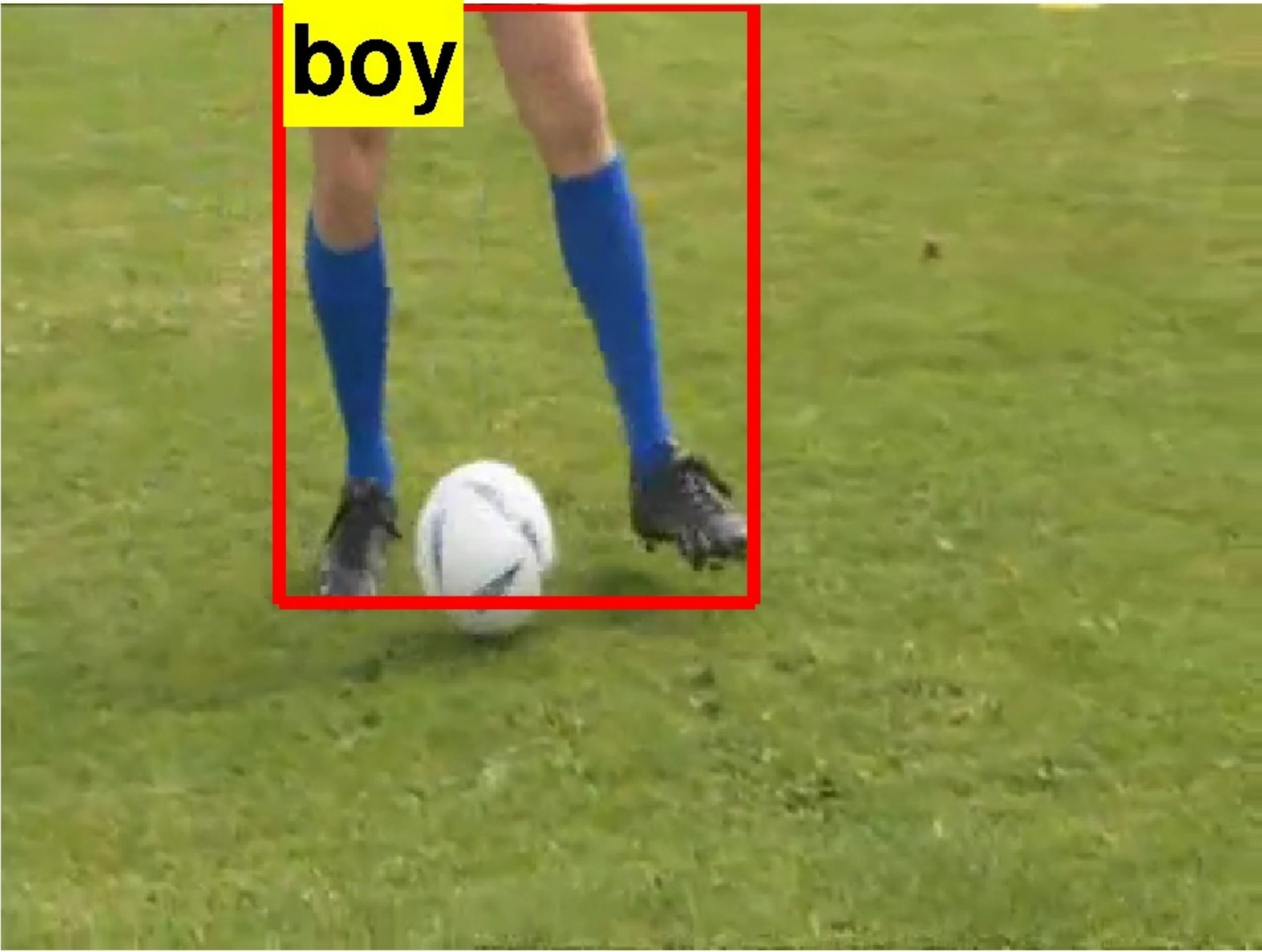}  &
 \includegraphics[height=47pt]{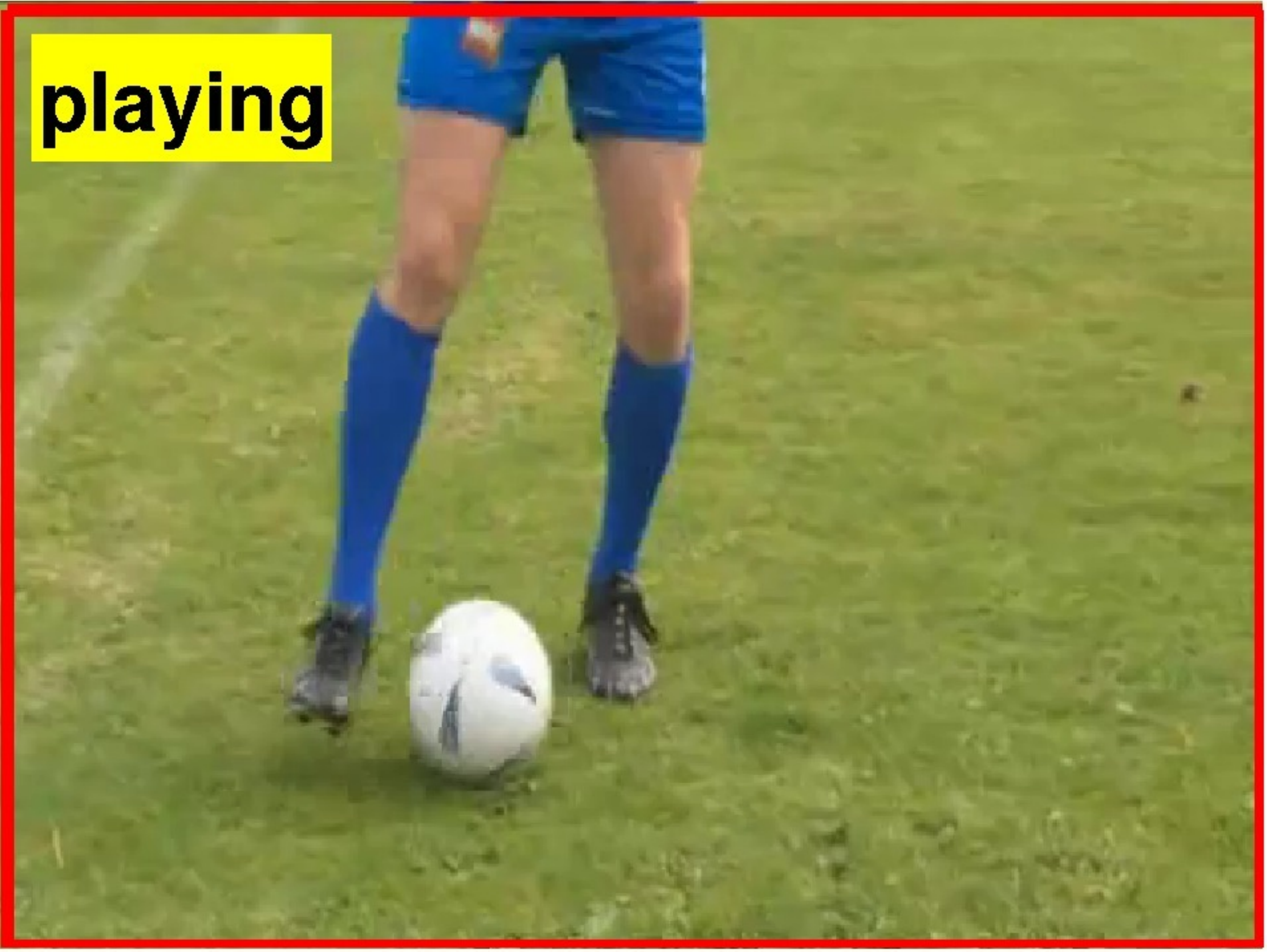} &
 \includegraphics[height=47pt]{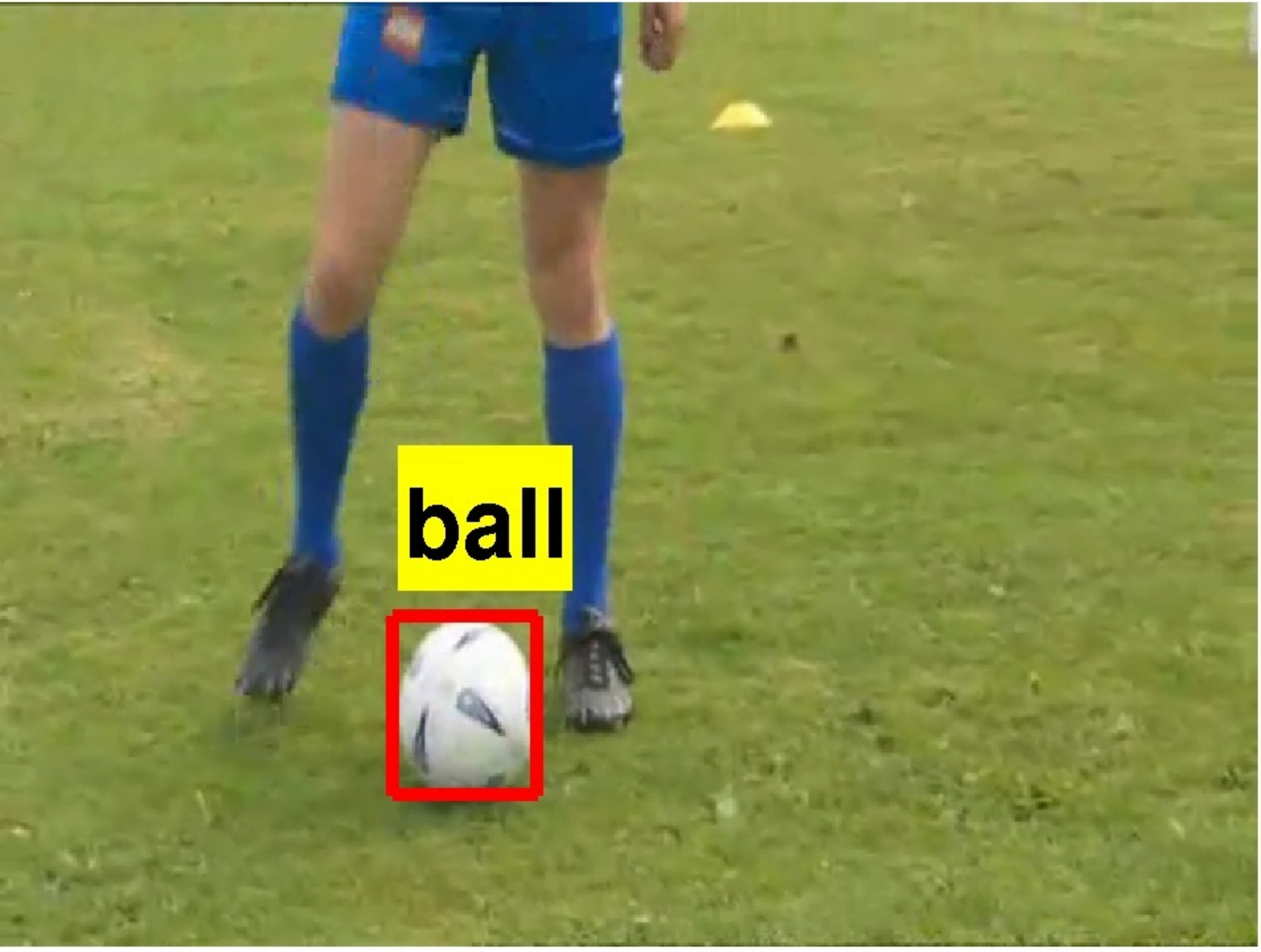} &
  \cfbox{green}{Ours: A \textbf{boy}  is \textbf{playing} a \textbf{ball}.}
  
  \hspace{0.02cm} Ref: A man playing with ball.\\

 

 \includegraphics[height=47pt]{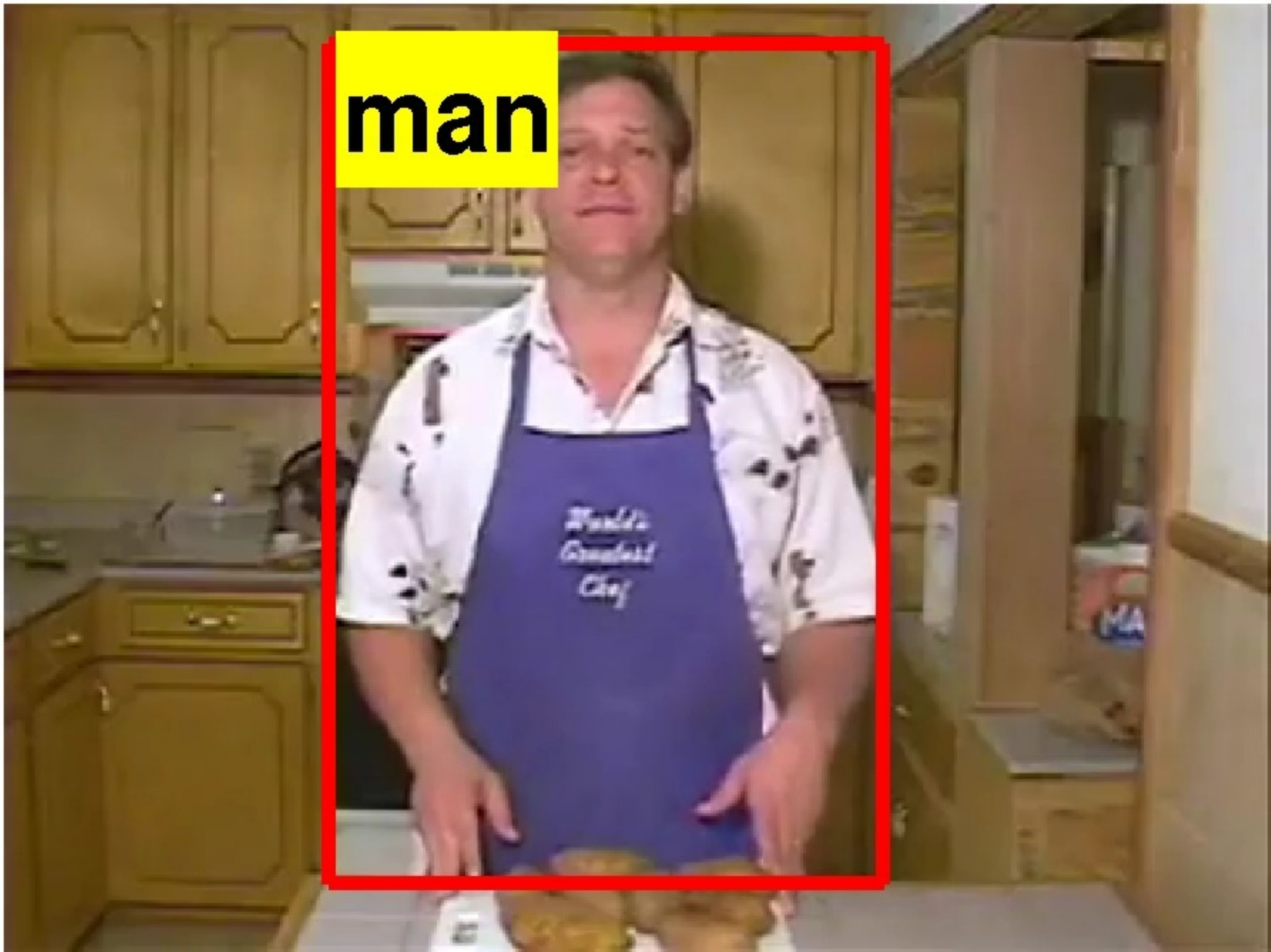}  &
 \includegraphics[height=47pt]{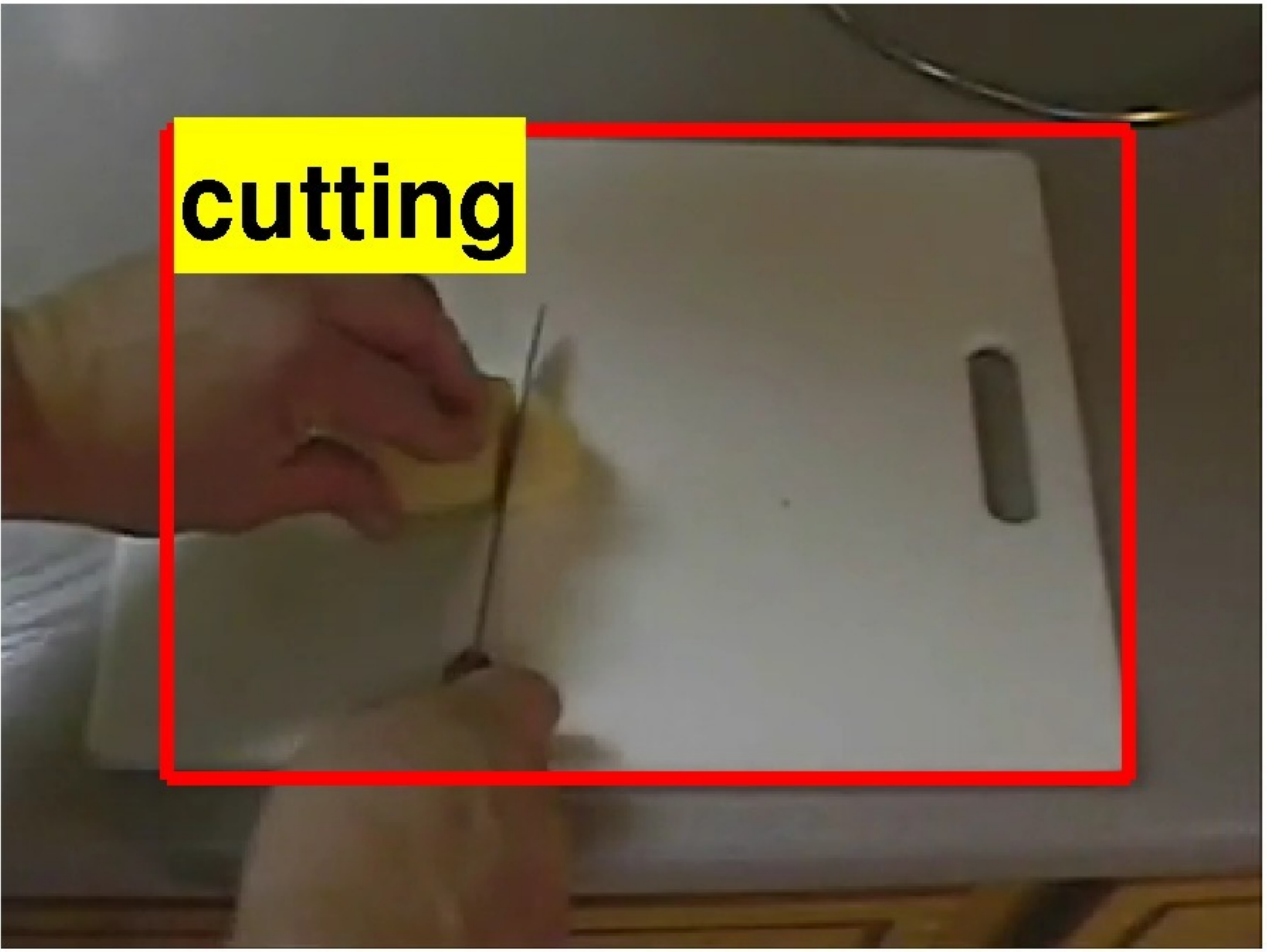} &
 \includegraphics[height=47pt]{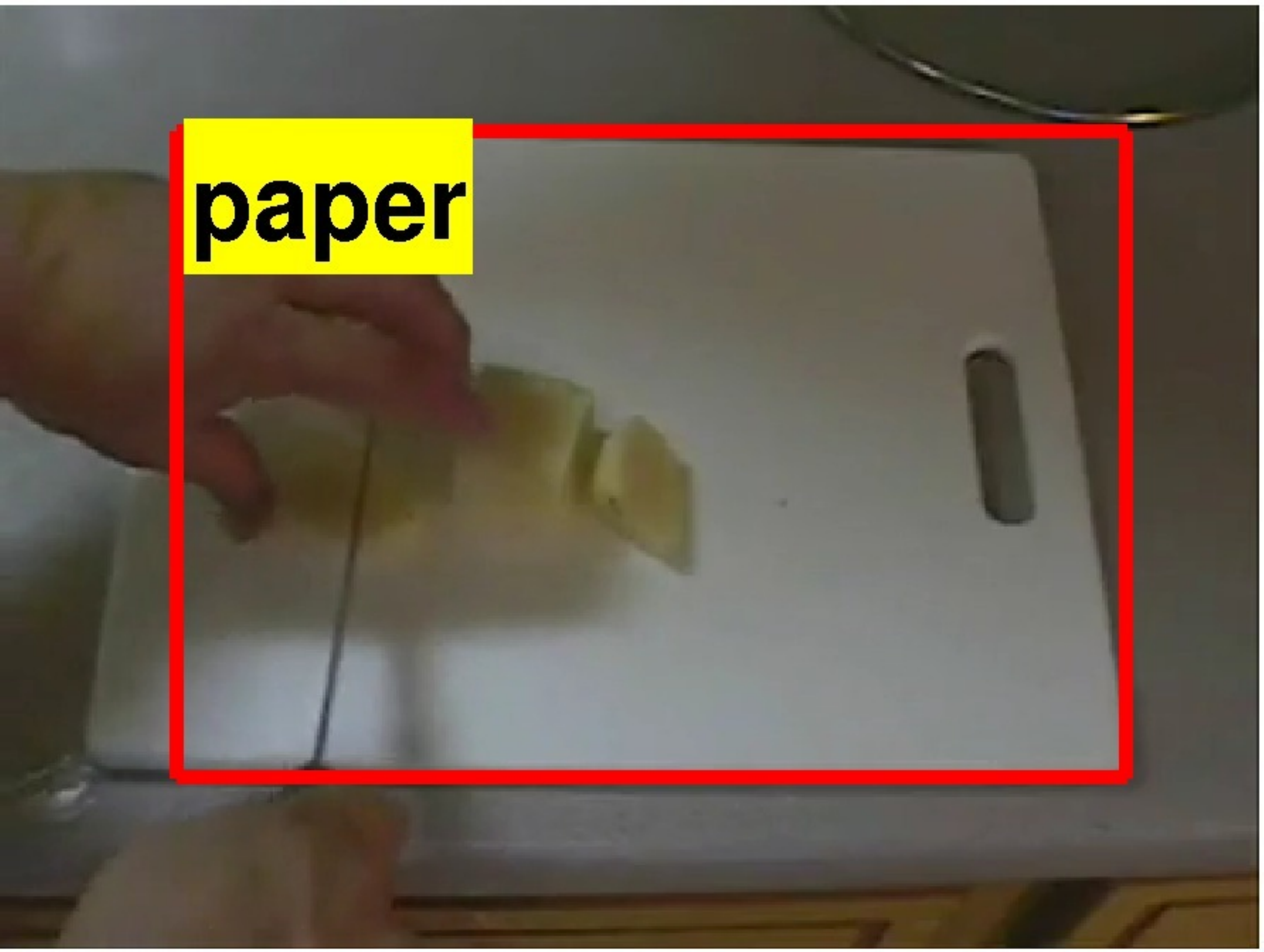} &
 \cfbox{red}{Ours: A \textbf{man} is \textbf{cutting} a piece of \textbf{paper}.}
 
 \hspace{0.02cm} Ref: A man cutting a potato.\\

 \includegraphics[height=47pt]{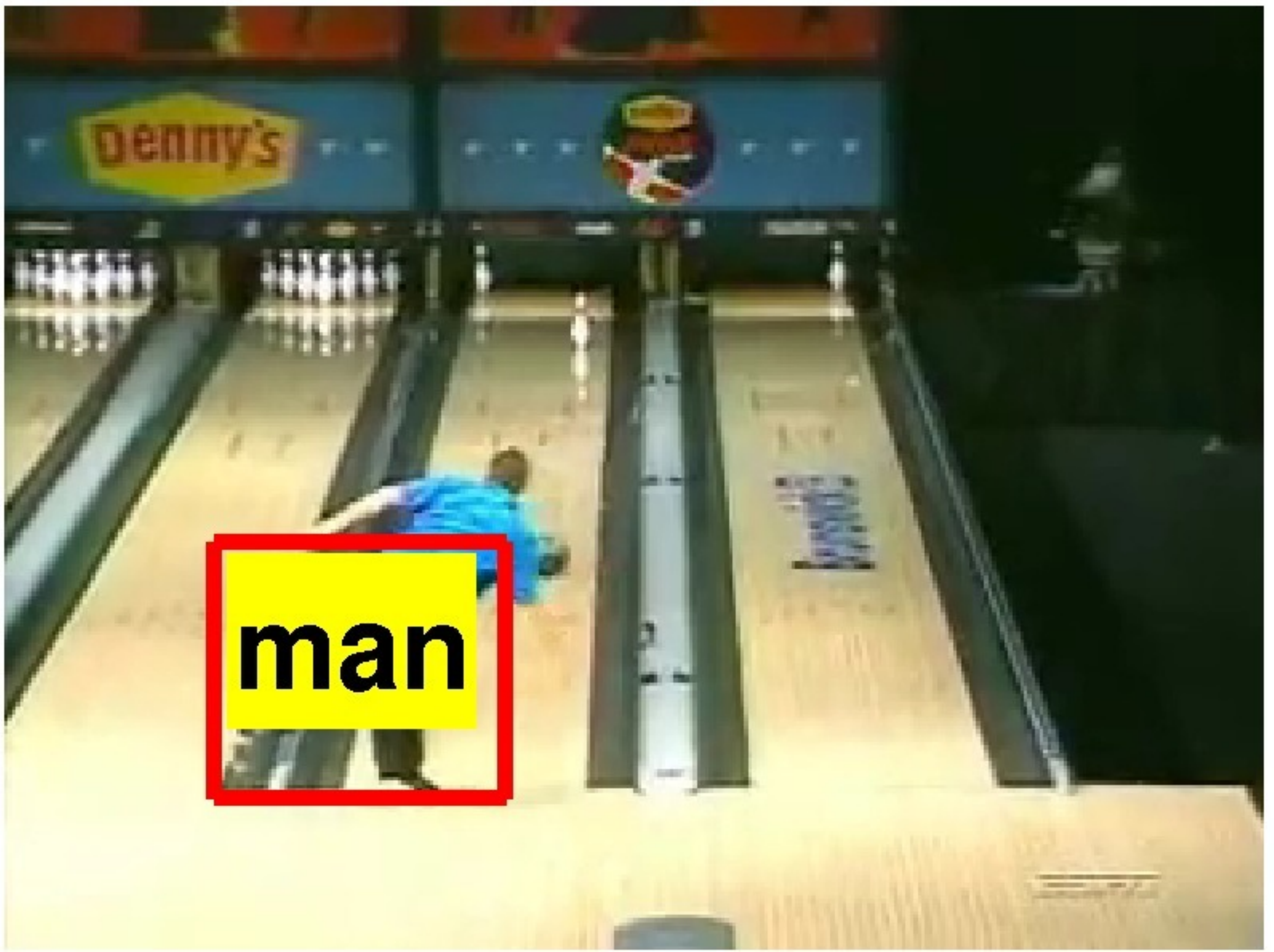}  &
 \includegraphics[height=47pt]{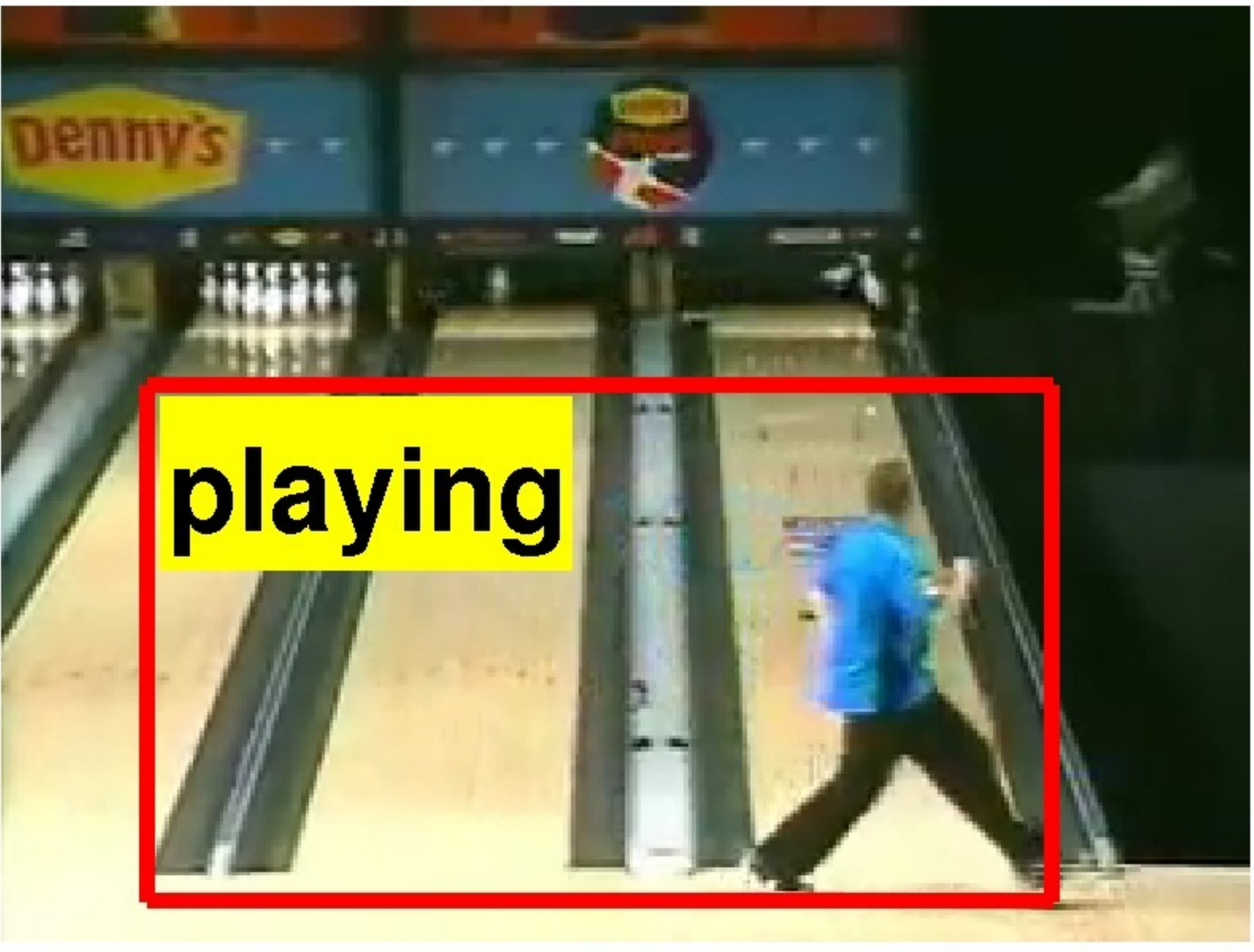} &
 \includegraphics[height=47pt]{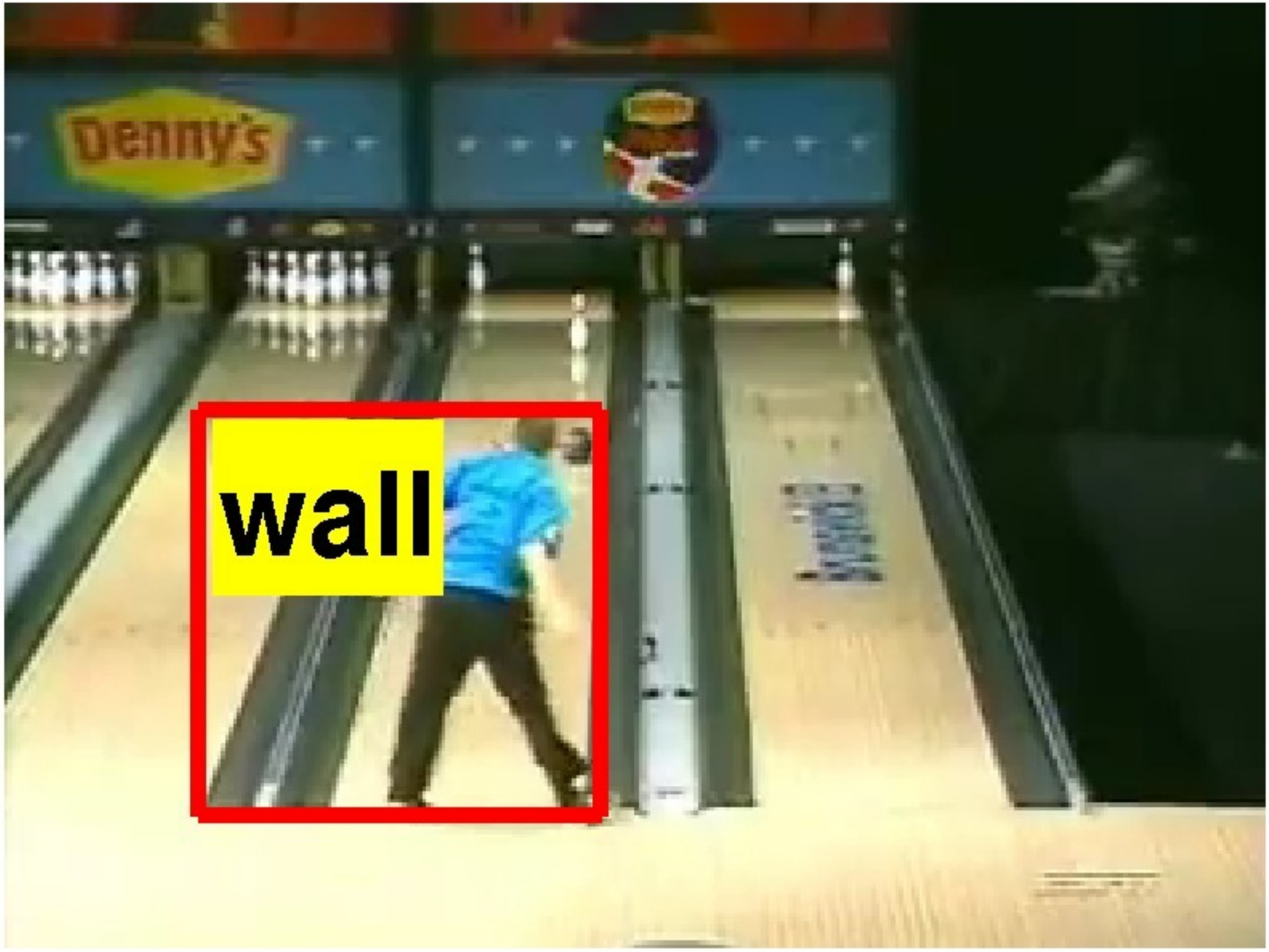} &
 \cfbox{red}{Ours: A \textbf{man} is \textbf{playing} a \textbf{wall}.}
 
 \hspace{0.02cm} Ref: A man is playing 8-balls.\\
 
 \includegraphics[height=36pt]{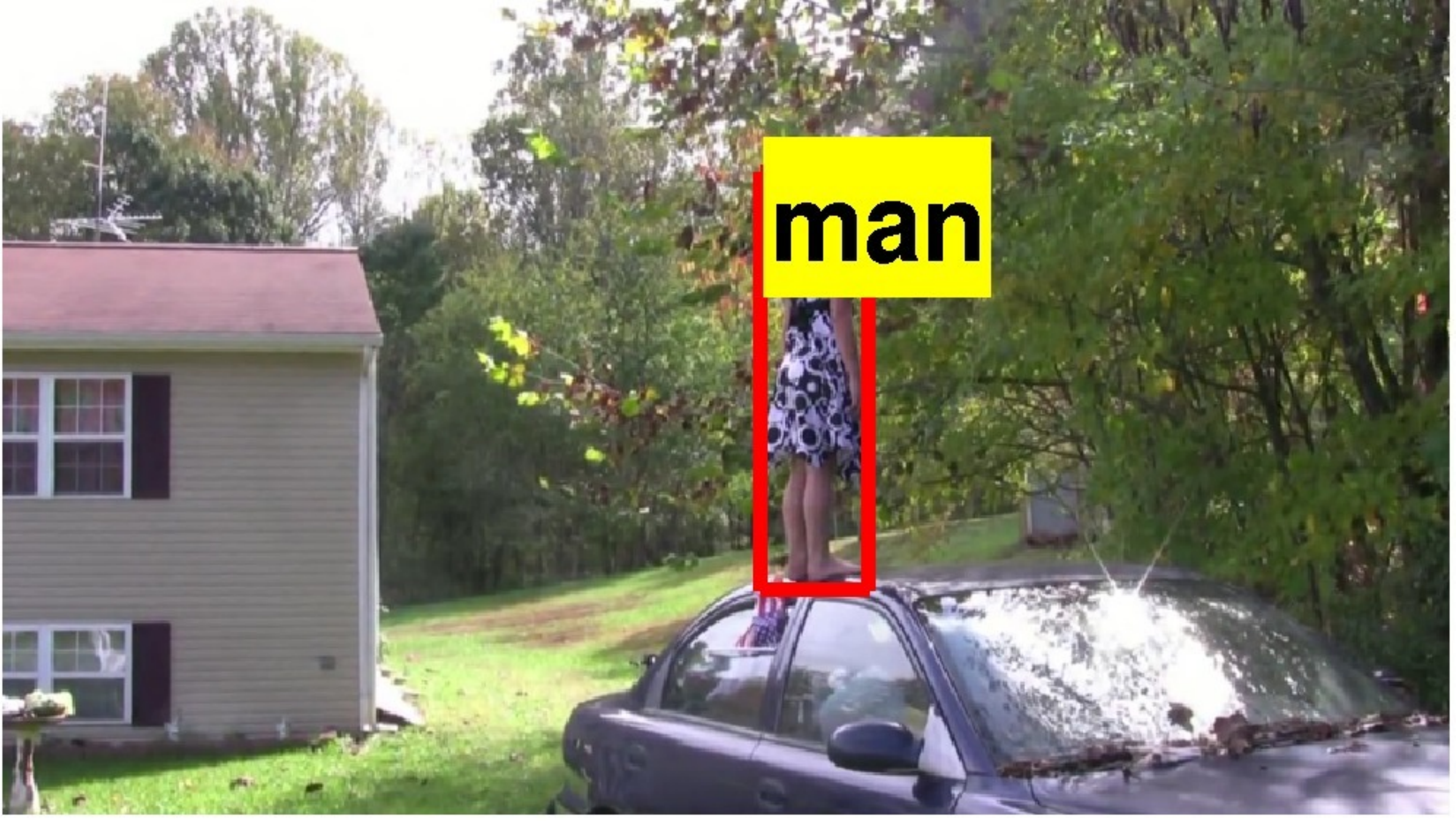}  &
 \includegraphics[height=36pt]{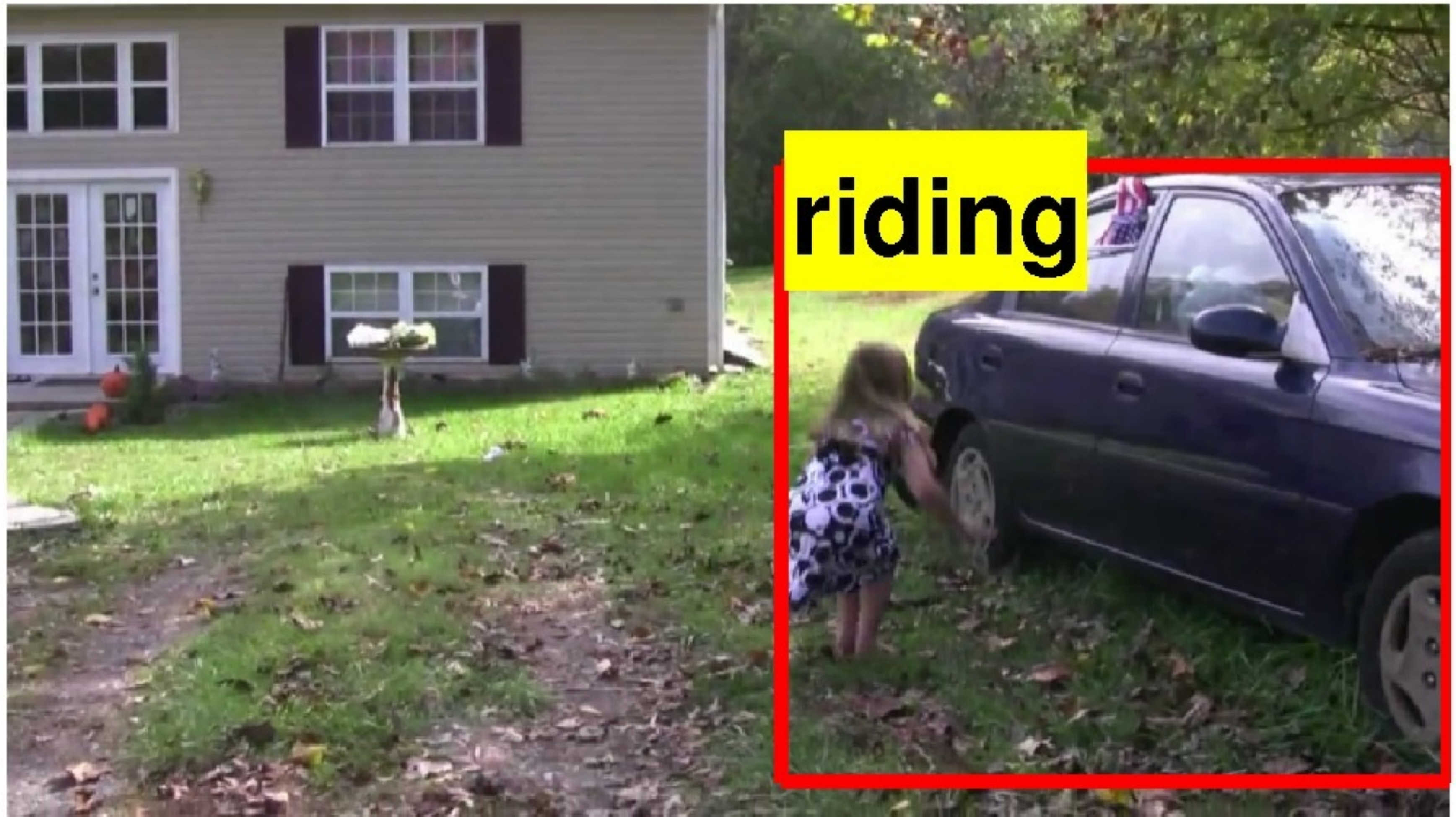} &
 \includegraphics[height=36pt]{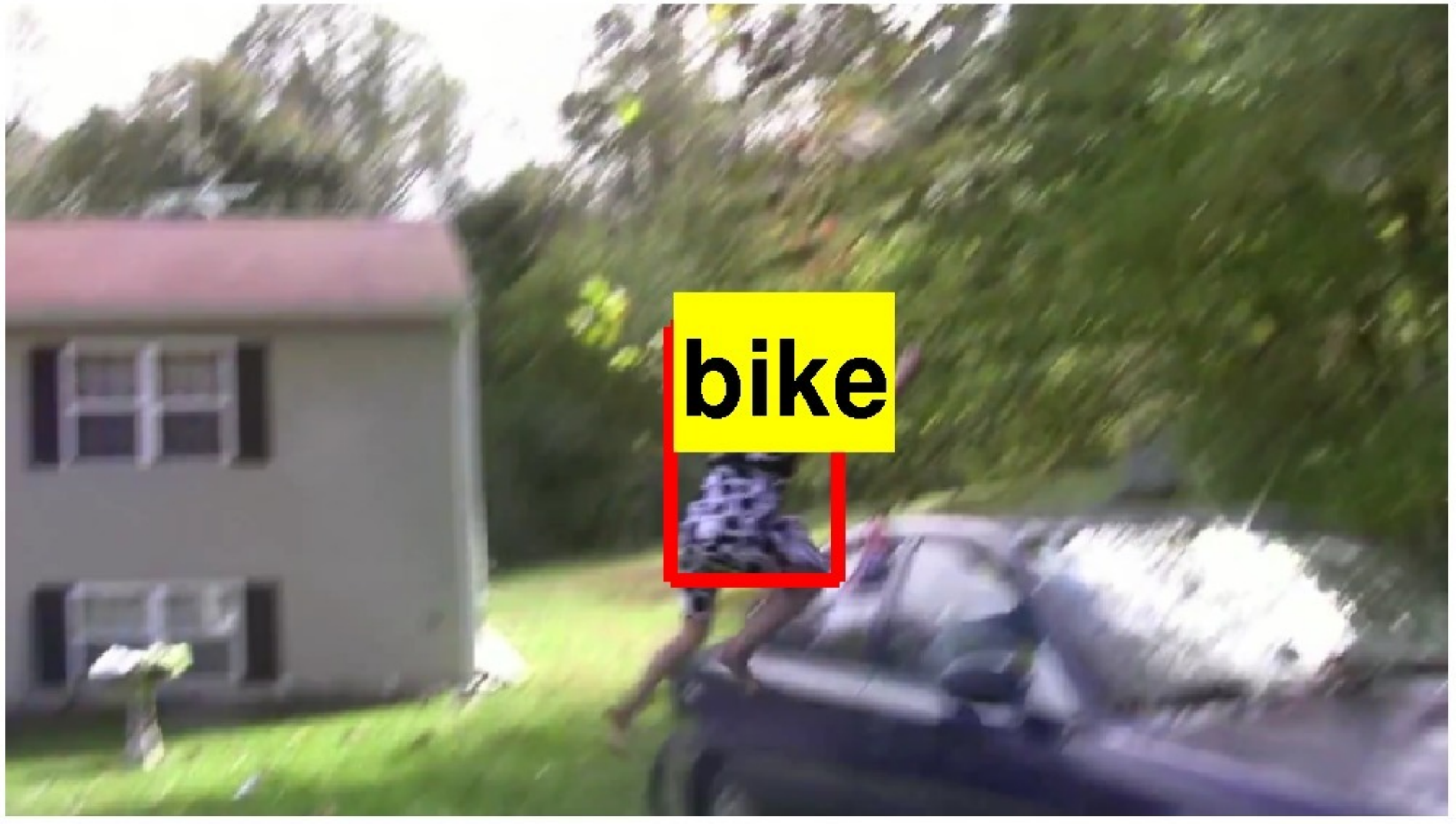} &
 \cfbox{red}{Ours: A \textbf{man} is \textbf{riding} a \textbf{bike}.}
 
 \hspace{0.02cm} Ref: A girl jumps on top of a car.\\

 \includegraphics[height=47pt]{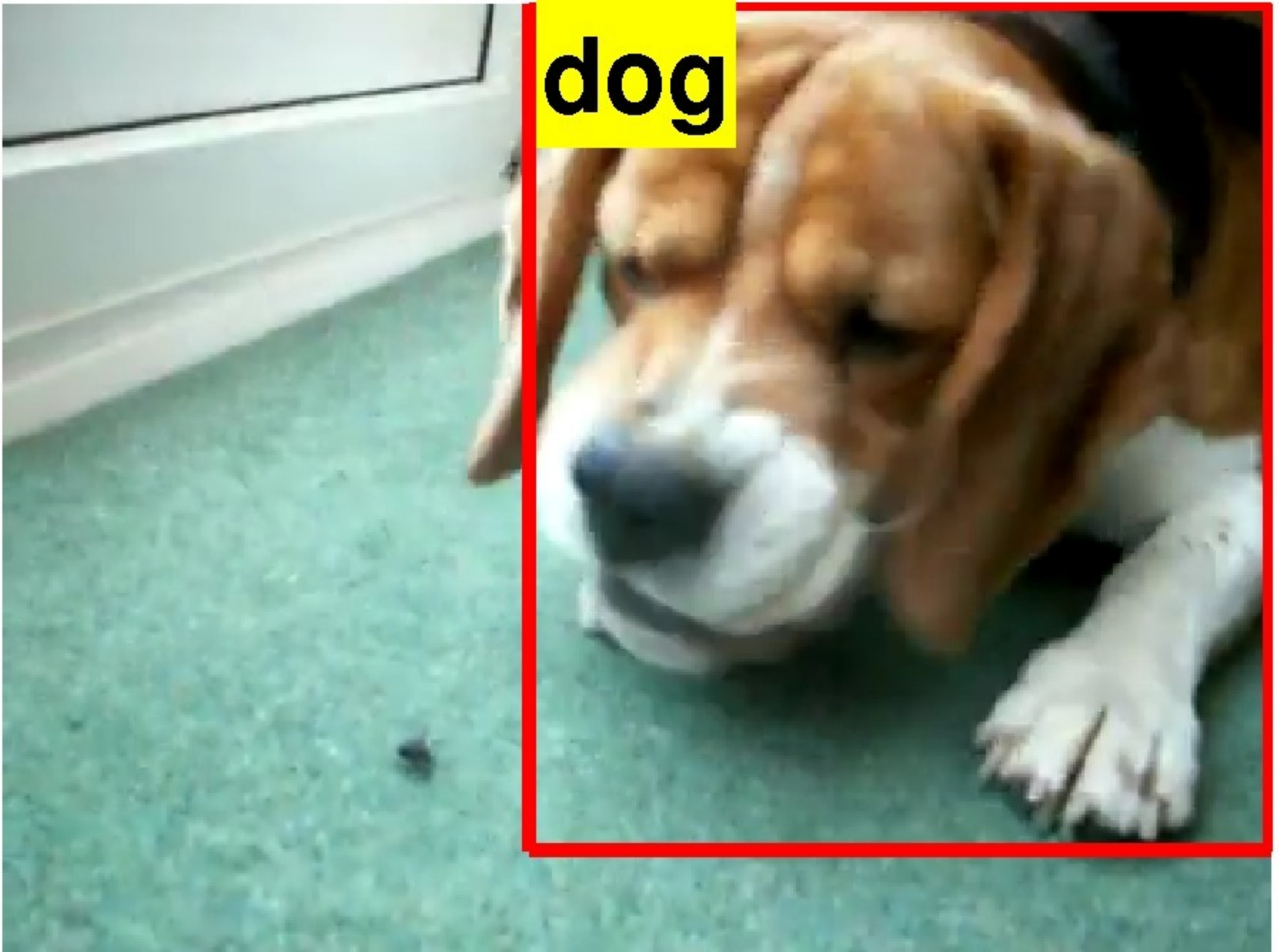}  &
 \includegraphics[height=47pt]{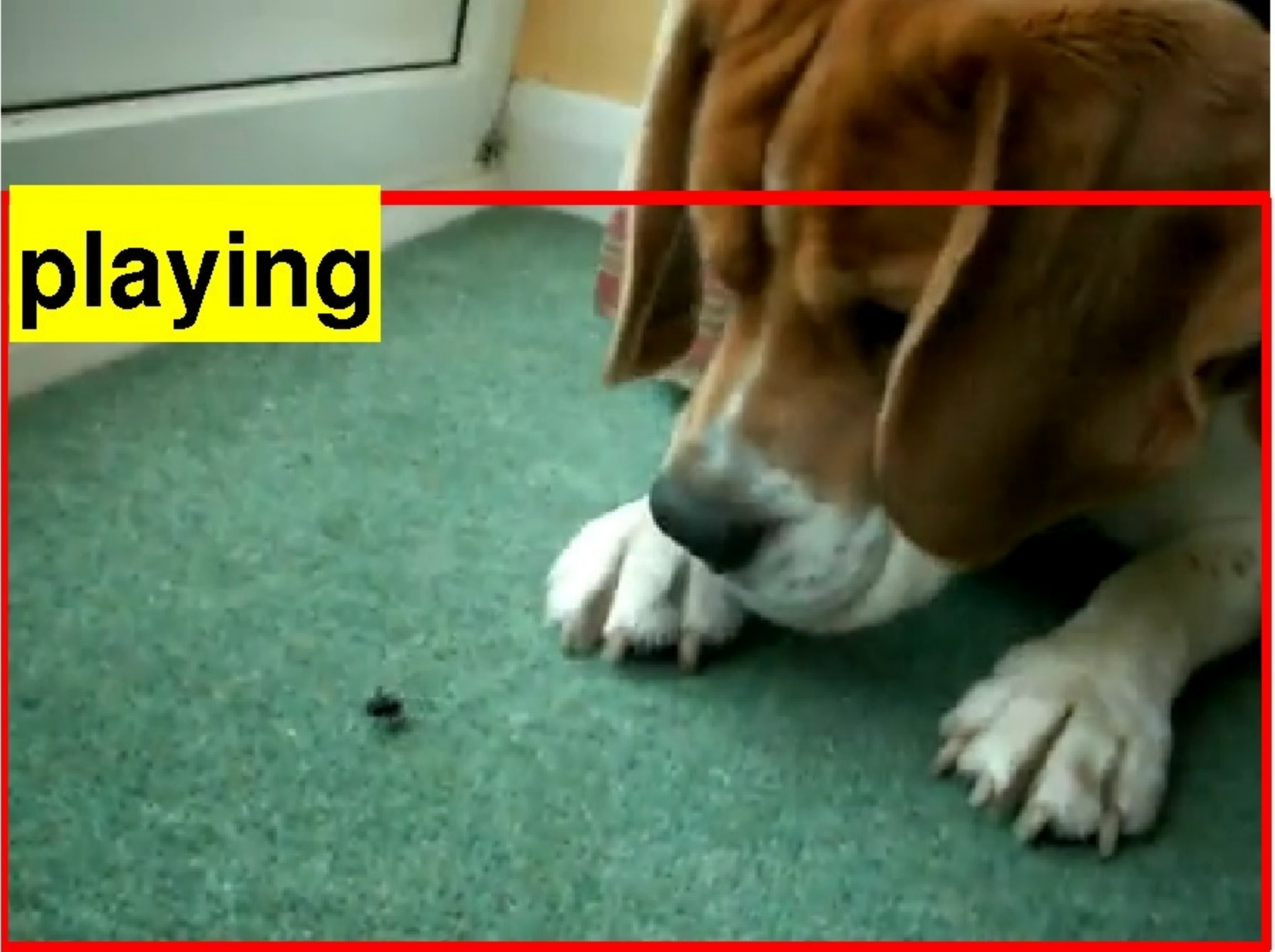} &
 \includegraphics[height=47pt]{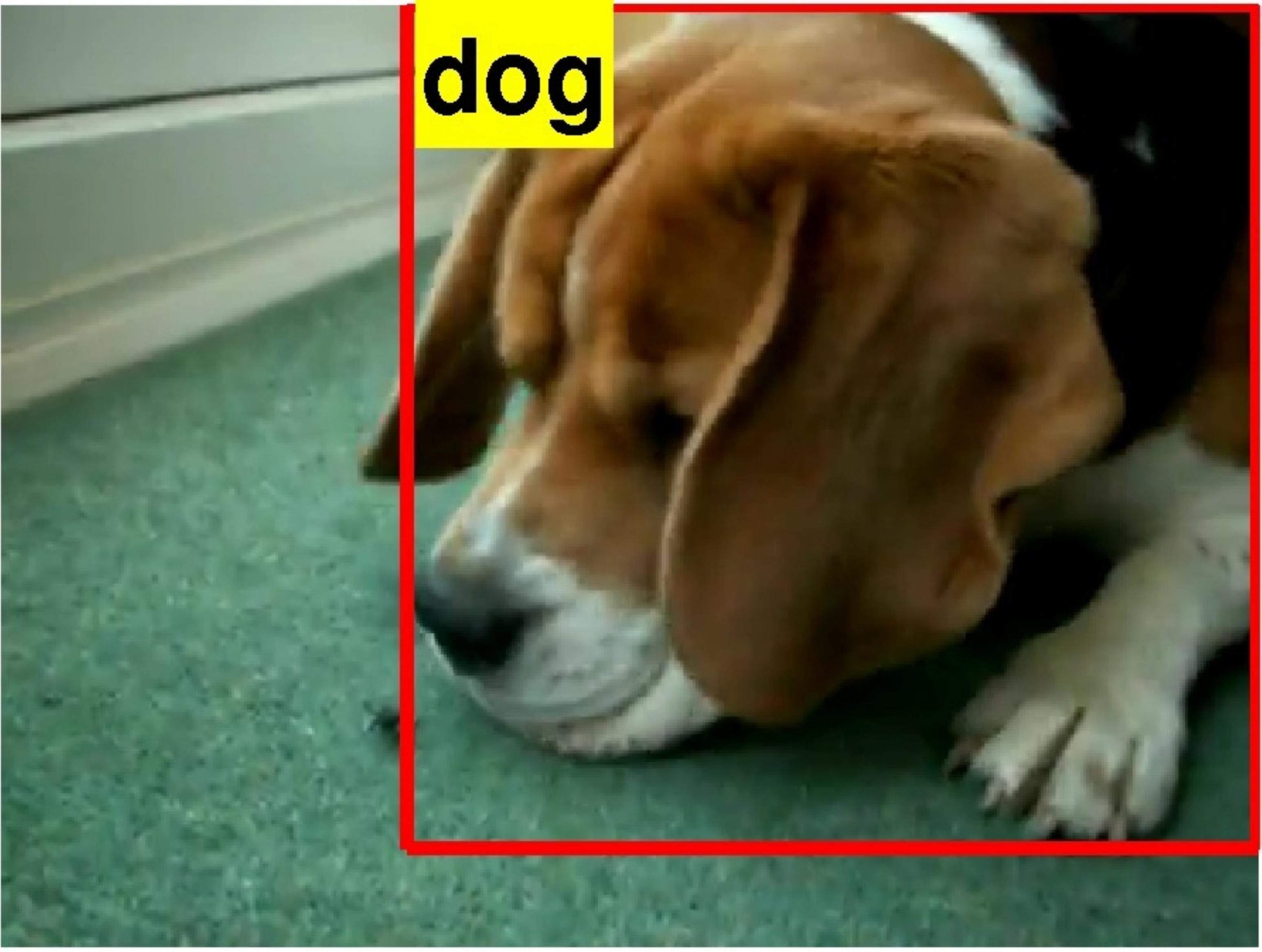} &
 \cfbox{red}{Ours: A \textbf{dog} is \textbf{playing} with a \textbf{dog}.}
 
 \hspace{0.02cm} Ref: A dog is playing with a fly.\\

\end{tabular}
}
\end{centering}
\caption{Highest scoring proposals of our model (according to $\beta$ eq. \ref{ec:weights}) for each emitted word in the sentence. We only illustrate the grounding of the main words in the sentence and ignore linking words. The complete sentence is shown in the right column together with the closest reference from the human annotations. For each proposal we show a single, randomly selected frame.}
\label{fig:visual_results}
\end{figure*}

 \mypar{Qualitative Results}. Our attention mechanism, built on top of spatio-temporal object proposals, allows for a \textit{visual explanation} of what the model ranked as the most relevant visual support for emitting a particular word. This can be done by inspecting the learned weights $\beta$ (see equation~\ref{ec:weights}) and their associated proposals. In figure~\ref{fig:visual_results} we show the proposal with the highest associated $\beta$ weight (a random frame from it) that was used in generating a particular word. For display purposes, we ignore linking words and articles that do not have a visual grounding in video. Our model correctly indicates the localization of the key video description components and the words it emitted, even for those having a very small spatial extent such as \textit{pepper}, \textit{ball}, \textit{toy}, \textit{gun}. There are cases when a single spatio-temporal proposal is chosen as the best visual explanation for multiple words, as it is with (\textit{girl, riding, horse}). We also show examples when the obtained sentences are wrong (marked with red in figure~\ref{fig:visual_results}). In some of the cases, our algorithm correctly identifies  parts of the sentences - especially subjects and verbs such as (\textit{man-cutting, dog-playing}) - but fails to find the correct object. This is due to the large variability in objects appearance and size and also depends on the quality of the spatio-temporal proposals pool.

\begin{table*}
\begin{center}
\scalebox{0.9}{
\begin{tabular}{l||c|c|c|c|c}
\hline
\textbf{ Method} & \textbf{BLEU@1} & \textbf{BLEU@2} & \textbf{BLEU@3} & \textbf{BLEU@4} & \textbf{METEOR} \\
\hline
\hline
FGM \cite{hvc-fgm} & - & - & - & 13.68 & 23.9\\
\hline
S2VT\cite{s2vt:iccv15}  & - & - & - &- &  29.8\\
\hline
MM-VDN\cite{MM-VDN} & - & - & - & 37.64 & 29.00\\
\hline

LSTM-YT-coco \cite{venugopalan:naacl15} & - & - & - & 33.29 & 29.07\\
\hline
LSTM-YT-coco+flicker \cite{venugopalan:naacl15}  & - & - & - & 33.29 &28.88\\
\hline

Temporal attention \cite{temporal_att15} & - & - & - & 41.92 & 29.60 \\
\hline
LSTM-E (VGG+C3D) \cite{pan2015}  & 78.8 & 66.0 & 55.4 & 45.3& 31.0\\
\hline
 h-RNN\cite{hierarchical_rnn15}  & 81.5 & 70.4 & 60.4 &  49.9& 32.6\\
\hline
 HRNE with attention\cite{hrne_15}  & 79.2 & 66.3 & 55.1 & 43.8 & \textbf{33.1}\\
\hline
GRU-RCNN \cite{gru_rcnn_15}  & - & - & - & 49.63 & 31.70\\
\hline
LSTM & 78.0 & 66.4 &  56.7 & 45.4  & 31.2\\
LSTM(SVO) & 80.1 & 68.1 &  57.5 & 45.8  & 31.2\\
LSTM(DET,CLS) & 81.2 & 68.9 &  57.9 & 46.2  & 31.1\\
LSTM(SVO,DET,CLS) & 80.8 & 69.3 &  59.3 & 48.3  & 30.7\\
LSTM-ATT & 80.1 & 68.9 &  59.4 & 48.7  & 31.9\\
LSTM-ATT(SVO)  & 81.0 & 70.5 & 61.2 & 50.5& 32.3\\
LSTM-ATT(DET,CLS)& 81.9 & 70.9 & 60.9 & 50.5 & 31.6\\
LSTM-ATT(SVO,DET,CLS) & 82.0 & 71.6 & 62.4 & 51.5 & 32.0\\
LSTM2-ATT(SVO)  & \textbf{82.4} & \textbf{71.8} & \textbf{62.5} & \textbf{52.0}& 32.3\\
LSTM2-ATT(DET,CLS)& 80.6 & 68.1 & 57.4 & 46.0 & 31.8\\
LSTM2-ATT(SVO,DET,CLS) & 81.5 & 70.8 & 61.5 & 50.6 & 32.4\\

\hline
\end{tabular}
}
\end{center}
\caption{Comparison with previous works on BLEU@1 - BLEU@4 and METEOR metrics. Values are reported as percentage \%. }  
\label{tab:accuracy}
\end{table*}

\section{Conclusions}

In this paper we have addressed some of the challenges in automatic video captioning by aiming to spatio-temporally ground the semantic video concepts, as an intermediate step, without grounding supervision. In contrast to most existing automatic video captioning systems that map from the raw video to the high level textual description, bypassing localization, we aim at aggregating additional, potentially valuable information, by relying on spatio-temporal video proposals and image classification responses for content localization and improved generalization, fused using deep neural network attention models, based on long short-term memory. Our resulting system produces competitive, state-of-the-art results in the standard YouTube captioning benchmark and offers the additional advantage of localizing the concepts (subjects, verbs, objects), with no grounding supervision, over space and time.

\vspace{1mm}
\noindent {\bf Acknowledgement} This work was supported in part by CNCS-UEFISCDI under PCE-2011-3-0438, JRP-RO-FR-2014-16 and NVIDIA through a GPU donation.

\bibliographystyle{splncs}

\bibliography{egbib}



\end{document}